\documentclass[11pt]{article}
\usepackage[letterpaper,margin=1in]{geometry}
\usepackage{amsmath,amssymb,amsthm}
\usepackage{newtxtext,newtxmath,microtype}
\usepackage[super,sort&compress,comma]{natbib}
\usepackage{graphicx,booktabs,longtable,tabularx,array}
\usepackage{setspace,caption,titlesec,placeins,float,etoolbox}
\usepackage{hyperref,url,xurl}
\renewcommand*{\HyperDestNameFilter}[1]{\jobname-#1}
\titleformat{\section}{\large\bfseries}{\thesection}{0.6em}{}
\titlespacing*{\section}{0pt}{1.1em plus 0.2em minus 0.1em}{0.4em}
\titleformat{\subsection}{\normalsize\bfseries}{\thesubsection}{0.6em}{}
\titlespacing*{\subsection}{0pt}{0.85em plus 0.15em minus 0.1em}{0.3em}
\hypersetup{hidelinks,pdfauthor={Tianchi Chen and Shihua Yu},
  pdfsubject={Finite-window response and tests of retained representations},
  pdfkeywords={response theory, retained state, distributed AI, representation}}
\newcommand{\externalnumber}[1]{\ifcsname publicref@#1\endcsname
  \csname publicref@#1\endcsname\else\textbf{??}\fi}
\newcommand{\siref}[1]{\externalnumber{#1}}

\newcolumntype{Y}{>{\raggedright\arraybackslash}X}
\newcounter{suppstatement}[section]

\newenvironment{nphysstatement}[1]{%
  \refstepcounter{suppstatement}%
  \par\smallskip\noindent\textbf{#1.}\ }{\par\smallskip}

\expandafter\def\csname publicref@assump:wellposed-delayed-field\endcsname{S2.2}
\expandafter\def\csname publicref@eq:coherence-bound\endcsname{S2.10}
\expandafter\def\csname publicref@eq:coherence-functional\endcsname{S2.9}
\expandafter\def\csname publicref@eq:supp-consensus-V\endcsname{S3.5}
\expandafter\def\csname publicref@eq:supp-consensus-Zvar\endcsname{S3.2}
\expandafter\def\csname publicref@eq:supp-consensus-etam\endcsname{S3.1}
\expandafter\def\csname publicref@eq:supp-consensus-fluct-eq\endcsname{S3.3}
\expandafter\def\csname publicref@eq:supp-consensus-modevar\endcsname{S3.4}
\expandafter\def\csname publicref@eq:supp-full-state-flow\endcsname{S2.3}
\expandafter\def\csname publicref@eq:supp-linear-branch-response\endcsname{S2.6}
\expandafter\def\csname publicref@eq:supp-linear-endpoint-response\endcsname{S2.8}
\expandafter\def\csname publicref@eq:supp-memory-law\endcsname{S2.2}
\expandafter\def\csname publicref@eq:supp-retained-window-response\endcsname{S2.7}
\expandafter\def\csname publicref@eq:supp-slow-linearization\endcsname{S2.5}
\expandafter\def\csname publicref@eq:supp-slow-state-law\endcsname{S2.4}
\expandafter\def\csname publicref@eq:supp-universal-interface\endcsname{S2.1}
\expandafter\def\csname publicref@fig:consensus-criteria\endcsname{S1}
\expandafter\def\csname publicref@fig:development-decisions\endcsname{S3}
\expandafter\def\csname publicref@fig:history-decision\endcsname{S2}
\expandafter\def\csname publicref@iclr:additional-methods\endcsname{S7.1}
\expandafter\def\csname publicref@prop:interface-equivalence\endcsname{S2.1}
\expandafter\def\csname publicref@prop:representation-sufficiency\endcsname{S1.6}
\expandafter\def\csname publicref@sec:minimal-realization\endcsname{S1.6}
\expandafter\def\csname publicref@sec:slow-state-derivation\endcsname{S2.2}
\expandafter\def\csname publicref@sec:vicsek-field-window\endcsname{S3.2}
\expandafter\def\csname publicref@si:boundary-susceptibility\endcsname{S2.4}
\expandafter\def\csname publicref@si:closed-loop\endcsname{S5}
\expandafter\def\csname publicref@si:comparing-retained-responses\endcsname{S1.5}
\expandafter\def\csname publicref@si:consensus-network\endcsname{S3.1}
\expandafter\def\csname publicref@si:cross-family-comparison\endcsname{S7.6}
\expandafter\def\csname publicref@si:data-code-availability\endcsname{S8}
\expandafter\def\csname publicref@si:development-decisions\endcsname{S7.4}
\expandafter\def\csname publicref@si:empirical-protocols\endcsname{S7.1}
\expandafter\def\csname publicref@si:empirical-realizations\endcsname{S7}
\expandafter\def\csname publicref@si:exploratory-persistence\endcsname{S7.6.1}
\expandafter\def\csname publicref@si:falsifiable-retained-state\endcsname{S1.3}
\expandafter\def\csname publicref@si:formal-delayed-model\endcsname{S2.1}
\expandafter\def\csname publicref@si:history-decision\endcsname{S7.3}
\expandafter\def\csname publicref@si:implementation-guide\endcsname{S8.1}
\expandafter\def\csname publicref@si:intervention-scope\endcsname{S7.6.1}
\expandafter\def\csname publicref@si:inverse-experiments\endcsname{S4}
\expandafter\def\csname publicref@si:inverse-model-methods\endcsname{S4.2}
\expandafter\def\csname publicref@si:inverse-mwe-reconstruction\endcsname{S4.1}
\expandafter\def\csname publicref@si:live-paired-source-perturbation\endcsname{S7.1.1}
\expandafter\def\csname publicref@si:local-readout-regimes\endcsname{S2.4.1}
\expandafter\def\csname publicref@si:minimal-realization\endcsname{S1.6}
\expandafter\def\csname publicref@si:notation-terminology\endcsname{S9}
\expandafter\def\csname publicref@si:observation-state\endcsname{S1.2}
\expandafter\def\csname publicref@si:offline-graph-diagnosis\endcsname{S7.2.1}
\expandafter\def\csname publicref@si:physical-field-law\endcsname{S2}
\expandafter\def\csname publicref@si:physical-realizations\endcsname{S3}
\expandafter\def\csname publicref@si:program-diagnosis\endcsname{S7.2}
\expandafter\def\csname publicref@si:prospective-diagnostic-pilot\endcsname{S7.2.2}
\expandafter\def\csname publicref@si:reconstruction-levels\endcsname{S8.1}
\expandafter\def\csname publicref@si:reduced-coordinate-comparison\endcsname{S7.6}
\expandafter\def\csname publicref@si:reduced-state-equation\endcsname{S2.2}
\expandafter\def\csname publicref@si:refinement-methods\endcsname{S4.2.1}
\expandafter\def\csname publicref@si:reported-coordinates-controls\endcsname{S7.6}
\expandafter\def\csname publicref@si:representation-sufficient\endcsname{S1.4.1}
\expandafter\def\csname publicref@si:response-theory\endcsname{S1}
\expandafter\def\csname publicref@si:scientific-evidence-methods\endcsname{S6}
\expandafter\def\csname publicref@si:scope-organization\endcsname{S1.1}
\expandafter\def\csname publicref@si:selected-path-measurements\endcsname{S7.6}
\expandafter\def\csname publicref@si:source-encoding\endcsname{S1.4}
\expandafter\def\csname publicref@si:susceptibility-comparison\endcsname{S2.5}
\expandafter\def\csname publicref@si:tab:symbols\endcsname{S7}
\expandafter\def\csname publicref@si:vicsek-flock\endcsname{S3.2}
\expandafter\def\csname publicref@si:well-posedness-coherence\endcsname{S2.3}
\expandafter\def\csname publicref@si:worked-program\endcsname{S7.2.1}
\expandafter\def\csname publicref@si:workflow-development\endcsname{S7.5}
\expandafter\def\csname publicref@tab:closed-loop-development-counts\endcsname{S3}
\expandafter\def\csname publicref@tab:crossed-counts\endcsname{S2}
\expandafter\def\csname publicref@tab:execution-roles\endcsname{S6}
\expandafter\def\csname publicref@tab:operator-dictionary\endcsname{S1}
\expandafter\def\csname publicref@tab:prospective-diagnostic-pilot\endcsname{S5}
\expandafter\def\csname publicref@tab:reconstruction-coverage\endcsname{S6}
\expandafter\def\csname publicref@tab:rpw-specialization\endcsname{S1}
\expandafter\def\csname publicref@tab:scientific-evidence-counts\endcsname{S4}

\hypersetup{pdftitle={Active Neural Matter: A Finite-Window Response Framework for Physical Systems and Distributed AI Workflows}}
\begin{document}
\begin{center}
{\Large\bfseries Active Neural Matter:\\[0.25em]
A Finite-Window Response Framework for\\[0.25em]
Physical Systems and Distributed AI Workflows\par}
\end{center}
\begin{center}
Tianchi Chen\textsuperscript{1,*}, Shihua Yu\textsuperscript{2}\\[0.5em]
\small
\textsuperscript{1}Merck \& Co., Inc., Rahway, NJ, USA\\
\textsuperscript{2}Department of Electrical and Computer Engineering,\\
Northeastern University, Boston, MA, USA\\
\textsuperscript{*}Corresponding author: \href{mailto:chentianchi@gmail.com}{chentianchi@gmail.com}
\end{center}

\begin{abstract}
Physical systems and distributed AI workflows can retain past information,
receive delayed feedback and combine conflicting signals. These features pose
a common modeling problem: determining what information about past
interactions a state must retain to account for future responses. Active
Neural Matter (ANM) addresses this problem through a finite-window response
framework. Forward analysis uses each system's dynamics to predict state and
output responses to perturbations, separating propagation from output
sensitivity. Inverse analysis tests whether a candidate state retains enough
information for these responses. With the system and future inputs or
interaction rules fixed, distinguishable responses from histories assigned the
same state identify distinctions the representation must preserve. Their
distributional separation gives a lower bound on the worst-case error of
approximating behavior from that state. Uncertainty bounds distinguish
rejection, support within a fixed tolerance on tested comparisons, and
unresolved cases. Delayed-consensus simulations check analytical predictions
of response amplitude and phase over a narrow range of delays; flocking
illustrates dependence on the chosen observable. An exact delayed-return
construction establishes a sufficient state for a specified program. The AI
studies identify distinctions in interaction history and evidence order that a
representation must preserve to reproduce observed response distributions.
Tests on new histories in a controlled feedback task provide bounded support
for a state specified by the task instructions; the instructed rule predicts
every observed action. ANM connects response prediction to measurable
information requirements, separating behavioral fidelity from task correctness
and initial-state adequacy from a closed evolution law for that state.
\end{abstract}

\section{Introduction}
\label{iclr:introduction}
A local perturbation can remain in a system after the interaction that produced
it has ended. Its later effect depends on how the system carries that change
forward and how the chosen observable responds. In a delayed physical network,
two identical instantaneous fields can have different futures because their
recent histories differ. In an AI workflow, an identical tool result can have
different meanings because it answers a different earlier request. These
examples pose a common question: what information must a state description
preserve to account for a specified response over a finite observation window?

A candidate description of information carried forward by a system is its
\emph{retained state}. It may be a physical field with its recent history, a set
of communication records, or selected coordinates of either. A \emph{readout}
is a measured observable, or a vector of observables; an endpoint readout is
evaluated at the end of the chosen window. A readout can also summarize that
window, provided the necessary observations are retained. These definitions
separate the state being proposed from the measurements used to test it.

\begin{figure}[tbp]
\centering
\includegraphics[width=\linewidth]{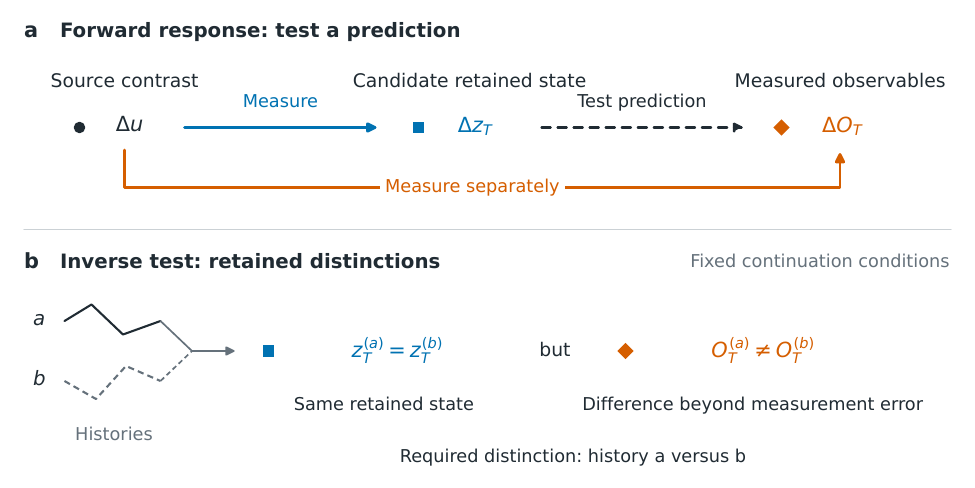}
\caption{\textbf{Forward response and inverse representation tests.}
\textbf{a}, A finite source contrast \(\Delta u\) produces separately measured
candidate-state and observable changes \(\Delta z_T,\Delta O_T\). Solid arrows denote
measurements; the dashed state-to-readout relation is a hypothesis, not an established causal path.
\textbf{b}, Histories \(a,b\) with identical states but deterministic readout differences
beyond measurement error must be distinguished by an adequate representation.
Stochastic readouts require resolved differences in the specified statistic or
conditional distribution. The continuation fixes later inputs or the policy
and environment rules generating them;
approximate matches require sensitivity and error bounds (Section~\ref{iclr:protocol}).
Panel b describes states at the end of the observation window. A test of a
preparation state instead compares initial summaries under the same
continuation conditions with the readouts obtained afterward.}
\label{fig:universal-pipeline}
\end{figure}

Active Neural Matter (ANM) treats a controlled input change as a source
perturbation and measures its effects on the state and readout separately
over a finite window (Figure~\ref{fig:universal-pipeline}).
The term \emph{neural} is inspired by neural dynamics, where past inputs can
influence later responses through persistent activity~\citep{funahashi1989mnemonic}
or activity-dependent changes in synaptic efficacy
(synaptic plasticity)~\citep{mongillo2008synaptic}.
ANM abstracts this relationship into source perturbations, retained states
and readouts. Information may be retained in field histories, internal memory
or workflow records; each realization supplies its own retention mechanism
and dynamics.

\emph{The choice of representation becomes an experimental hypothesis.}
Forward analysis predicts responses from specified states and dynamics. Inverse
analysis tests whether a candidate state retains enough information: with the
system and continuation conditions fixed, a resolved readout difference between
identical candidate states reveals an omitted distinction. Stochastic outputs
require distributional or statistical comparisons; approximate state matches
also require sensitivity and measurement-error bounds
(Section~\ref{iclr:protocol}). These tests constrain adequacy without uniquely
identifying a missing variable.

The description applies to continuous physical dynamics and discrete,
event-driven workflows, with realization-specific states, dynamics and response
functions. Here \emph{distributed AI workflow} means that producing, retaining
and using information occur in separate components. A downstream component
reads the records and produces an output; it can be a deterministic program
or a language model. The communication records permit response measurements
without access to model weights.
An externally specified graph and the request--return example in
Section~\ref{sec:worked-tool} instantiate these roles; neither requires or
identifies a state jointly formed by interacting live agents.

This organization draws on response theory
\citep{kubo1957,kubo1966,ruelle2009}, projection methods with memory
\citep{mori1965,zwanzig1961}, and state-based diagnosis
\citep{gertler1991analytical}.
Section~\ref{si:related-work} compares its measurement access and inferential
targets with these foundations and with workflow, memory and intervention methods.

The contributions comprise a common response formulation and its forward and
inverse uses:
\begin{enumerate}
\item \textbf{A common response formulation across physics and AI.}
Source perturbations, retained states and readouts define a common finite-window
response problem, with dynamics supplied by each realization. Keeping history
and readout choice explicit connects perturbation propagation to the information
needed for the resulting observations.
\item \textbf{Predict and separate responses.} Paired state and readout
measurements distinguish propagation from readout gains in a linear model,
as in standard state-space identification (Section~\ref{iclr:protocol}).
Finite-window predictions retain the history required by delays. Nested readout criteria distinguish continued operation from correct operation;
other readouts need not be nested.
\item \textbf{Test whether a representation retains sufficient information.}
Comparisons of response slopes or finite changes distinguish rejection, bounded
support within a preset tolerance, and unresolved adequacy. Resolved readout
differences at identical candidate states identify histories that an adequate
representation must distinguish. Their separation also gives a lower bound on the worst-case
behavioral approximation error of merging them
(Section~\ref{sec:compression-error}). New histories test a revision with the
program or model and its interface unchanged. In a controlled closed-loop
workflow, this procedure tests the version distinction prescribed by the
instructions, rejecting version-omitting summaries and supporting a predefined
refinement on new histories; the instructed rule predicts every observed action
(Section~\ref{sec:closed-loop}).
\end{enumerate}

The physical and graph studies first compare predicted and measured responses.
The inverse studies then ask what information must be retained when future
inputs are fixed, when actions select later observations, and when a model
judges scientific text. These comparisons lead to the information requirements
and approximation-error bound in Section~\ref{sec:information-requirements}.
The model results remain conditional on the tested histories and readouts;
Sections~\ref{si:inverse-model}--\ref{sec:routing-refinement} examine how
support changes with the choice of recipient, identifiers and request order.

\section{A common response framework for physical and AI systems}
\label{sec:cross-domain}
\label{iclr:protocol}

The common structure connects a source change to both retained information
and a measured observable. Continuous evolution and discrete transitions
supply different response maps, while the same representation question applies:
does the candidate state preserve every distinction needed for the specified
readout? Differentiable responses admit a local linear test; finite changes
are compared directly.
Each realization supplies its state space, dynamics, units and noise law.
In a computational graph, a \emph{typed source event} specifies an update time,
graph coordinate and evidence role; its numerical amplitude is the controlled
source. Table~\ref{tab:cross-domain-roles} maps the common measurement roles
to the physical and computational operations.
To compare readouts across realizations, specify a correspondence between
reached states. Readouts agree on corresponding states exactly when their
maps agree under that correspondence (SI
Section~\siref{si:comparing-retained-responses}). The reported systems share
measurement roles, without an assumed correspondence between their states.

\begin{table}[htbp]
\centering
\caption{\textbf{The same measurement roles, with different dynamics.}
Source, retained information and readout are specified separately.
The last column states the question answered by each realization; it does not
assert equality of their response functions.}
\label{tab:cross-domain-roles}
\begin{tabularx}{\linewidth}{@{}>{\raggedright\arraybackslash}p{0.17\linewidth}YYY@{}}
\toprule
\textbf{Realization} & \textbf{Controlled input} &
\textbf{Retained information and readout} & \textbf{Scientific role} \\
\midrule
Delayed consensus & Applied drive; coupling delay or noise in separate sweeps
& Field and recent history; measured amplitude, phase and spatial variance
& Calibrate the predicted driven response against an analytic solution. \\
\addlinespace[0.35em]
Polar flock & Angular-noise amplitude at fixed aligning field
& Positions and headings; polarization and angular alignment
& Compare nonlinear steady-state observables and their dependence on the chosen criterion. \\
\addlinespace[0.35em]
Typed-event graph & Amplitude of an event with specified time, site and channel
& Numerical graph state; a program or model reads specified terminal coordinates
& Measure imposed propagation and readout response separately. \\
\addlinespace[0.35em]
Delayed-tool workflow & Preparation histories or subsequent sensor returns
& Complete request--return records; analyst-selected descriptions and a fixed downstream component's output
& Test which historical distinctions a candidate representation must preserve. \\
\addlinespace[0.35em]
Adaptive test workflow & Which test belongs to an old version; new test outcome; return delivery
& Initial test records; distribution of model actions before and after a selected tool return
& Test state information and conditional predictions when actions determine later observations. \\
\addlinespace[0.35em]
Scientific-text comparison & Excerpt selection or document order
& Selected abstract text; the reader's conditional judgment distribution
& Test whether document summaries or unordered excerpts preserve the measured response distinctions. \\
\bottomrule
\end{tabularx}
\end{table}

In the graph realization, an extractor forms events from text, prescribed
updates transport their contributions, and a model reads the terminal scores
(Section~\ref{si:worked-graph}). This open-loop experiment separates propagation
from model response without readout feedback or an identification of the graph
update with the physical field equation.

The adaptive test workflow instead couples model actions to tool observations:
the chosen test determines the next request, and its delivered return affects
the next decision (Section~\ref{sec:closed-loop}). Its action policy and environment rule define discrete stochastic dynamics;
testing an initial summary does not assume autonomous reduced dynamics.

The inverse studies use records directly: the downstream component receives
complete histories or ordered excerpts, while candidate states are computed
only for analysis. Refining a candidate leaves the component and each input
unchanged. Controlled histories, excerpt selections or orderings provide the
comparisons; editing the component or its input would test a system intervention.

The full state \(s_t\) contains the histories and records required for
evolution and measurement. A candidate retained state \(z_t=G_f(s_t)\)
selects information from it; the tests ask whether this selection preserves
the specified response. Here \(f\) fixes experimental conditions, and
\(G_f:X_f\to Z_f\) maps the full-state space to the candidate state space.
A physical field \(c_t\) may be part of \(s_t\); with delay, its current
value alone need not determine the future. Sufficiency is relative to the
readout, window and continuation. Choosing a representation for analysis
leaves the system and its inputs unchanged.

A delayed return makes this distinction concrete. Two histories have identical
snapshots---no return and two pending requests---but exchange which sensor was
requested for recipients \(A\) and \(B\). Under the same later returns,
2 from sensor 1 and 7 from sensor 2, a program matching returns to requests
reports 2 for \(A\) in one history and 7 in the other. The snapshot merges
histories that the response distinguishes; retaining the association separates
them. Section~\ref{sec:worked-tool} establishes exact sufficiency for this
fixed program and then tests whether the association state also preserves a
language model's responses.

\subsection{Fixed inputs and adaptive observations}
\label{sec:adaptive-response}
A preparation-state test compares histories before the measured evolution.
Let \(h\) denote that history and \(z=G_f(h)\) its candidate summary. In an
open-loop experiment the later input sequence is specified externally. In a
closed-loop workflow an action changes the next observation, so fixing that
sequence would suppress the adaptation being tested.

A discrete description makes this distinction explicit. Let \(h_t\) be the
visible history before decision \(t\), \(a_t\) the selected action and
\(r_{t+1}\) the next tool observation. A fixed decision policy
\(\pi(a\mid h)\) gives action probabilities. A tool-response rule
\(q_u(r\mid h,a)\), with controlled external condition \(u\), gives observation
probabilities. Then
\begin{equation}
 a_t\sim\pi(\cdot\mid h_t),\qquad
 r_{t+1}\sim q_u(\cdot\mid h_t,a_t),\qquad
 h_{t+1}=\operatorname{append}(h_t,a_t,r_{t+1}).
 \label{eq:adaptive-workflow}
\end{equation}
The append operation adds the action and delivered observation to the history.
A return-withholding intervention changes the observation delivered to the next
call while preserving the request. Unrolling these decisions in time gives
forward event dependencies even though the workflow has feedback.

Fix the policy, environment rule, observation intervention and window, and
compare the resulting readout laws for histories with the same \(z\).
The sampled actions need not coincide. This supplies the same representation
question as a fixed-input comparison: does the chosen description preserve the
specified response distribution? The differentiable formulas below and finite
categorical comparisons are distinct specializations of that question.
An initial summary sufficient for one window need not predict its own next
state; autonomous reduced dynamics require additional closure conditions.

\subsection{Forward analysis: propagation and readout response}
A source history produces a trajectory and its measured observables over the
chosen window. Finite contrasts compare the retained states and readouts under
two source histories, with preparation and other inputs fixed. Let \(u_t\)
collect the controlled source coordinates and let
\(O_T(s_T)=(O_{1,T},\ldots,O_{d,T})\in\mathbb R^d\) collect \(d\ge1\)
numerical observables at time \(T\). Window summaries such as amplitude and
phase require the accumulated measurements in \(s_T\) (SI
Section~\siref{si:observation-state}). Stochastic outputs are compared through
a specified conditional statistic or output distribution.

\begin{figure}[htbp]
\centering
\includegraphics[width=\linewidth]{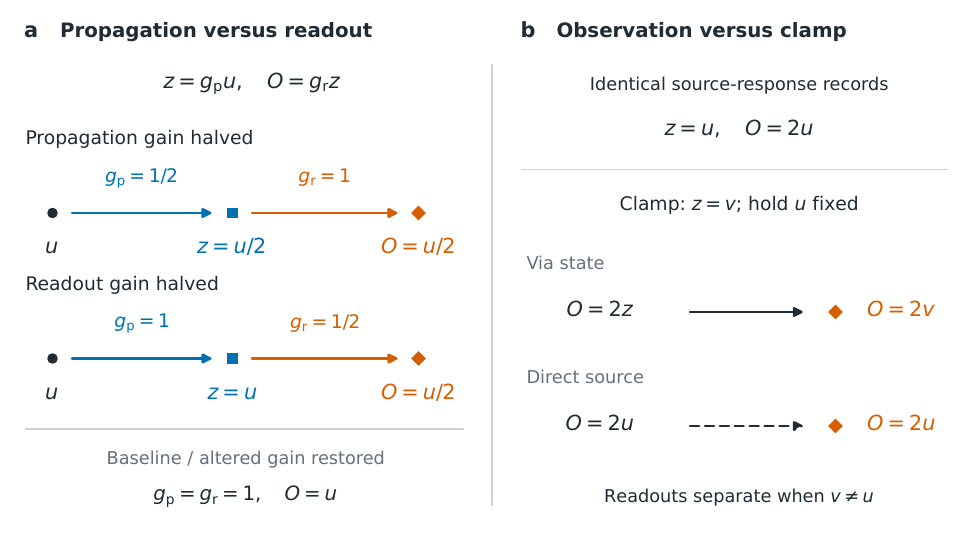}
\caption{\textbf{State measurement and causal intervention answer different questions.}
Here \(u\) is the source, \(z\) the retained state and \(O\) the endpoint readout.
\textbf{a}, In \(z=g_{\rm p}u,\ O=g_{\rm r}z\), \(g_{\rm p}\) is the propagation
gain and \(g_{\rm r}\) the readout gain. Halving either from
\(g_{\rm p}=g_{\rm r}=1\) gives the same endpoint \(O=u/2\), but different
states. Restoring the altered gain restores \(O=u\); ordinary state-space
identification gives the same decision.
\textbf{b}, The models \(O=2z\) and \(O=2u\), both with \(z=u\), have
identical source-response records. Independently setting \(z=v\) while holding
\(u\) fixed (a clamp) separates them. These are analytic models, not AI measurements
(SI Section~\siref{si:falsifiable-retained-state}).}
\label{fig:mechanism-controls}
\end{figure}

In response formulas, \(\Delta\) denotes finite contrasts and \(\delta\) tangent
perturbations. A first-order response is the differentiable specialization.
The derivative \(\mathrm D_xF\) maps a small change in \(x\) to the
first-order change in \(F\): it is a Jacobian in finite-dimensional
coordinates, or a Fr\'echet derivative on the specified function space.
On a declared time grid write the transition as
\(s_t=F_t(s_{t-1},u_t)\), with \(A_t=\mathrm D_sF_t\) and
\(B_t=\mathrm D_uF_t\) evaluated along the baseline trajectory.
Differentiability of the transition, coarse-graining and readout maps on the
perturbation space gives
\begin{align}
\delta s_t&=A_t\delta s_{t-1}+B_t\delta u_t,
\label{eq:finite-graph-linear-response}\\
\delta O_T&=\mathrm D_sO_T(s_T)
       \sum_{t=1}^{T}\Phi(T,t)B_t\delta u_t,
&\Phi(T,t)&=A_T\cdots A_{t+1}.
\label{eq:finite-window-endpoint-response}
\end{align}
The retained-state response applies \(\mathrm D_sG_f\) to the
same propagated displacement. Each term follows one perturbation: \(B_t\) injects it, successive transitions
\(A_t\) propagate it, and \(\mathrm D_sO_T\) converts the terminal displacement
into a measured change. Here \(\Phi(T,T)=I\), the identity operator. The formula fixes the initial state;
varying its history adds \((\mathrm D_sO_T)\Phi(T,0)\delta s_0\).
If the endpoint is a differentiable function \(\bar O_T\) of the retained
state, then \(O_T(s_T)=\bar O_T(G_f(s_T))\). The bar distinguishes this
function of \(z_T\) from the function of \(s_T\); it does not denote an
average. The chain rule gives
\[
\mathrm D_sO_T(s_T)=\mathrm D_z\bar O_T(z_T)\,\mathrm D_sG_f(s_T).
\]
The rightmost derivative maps full-state to retained-state perturbations;
the other maps them to the readout. This composition is the hypothesis tested
against the independently measured response. For the identity representation
\(G_f(s)=s\), its derivative is \(I\); a graph projection must separately
retain the histories and records required by the readout.
Differentiability is additional to Lipschitz well-posedness; nonsmooth points
require finite contrasts or specified one-sided responses. \emph{Step refinement}
compares slopes at two perturbation sizes to assess a first-order interpretation.
Figure~\ref{fig:mechanism-controls} shows why state measurements and selective
state interventions answer different questions.

The time at which a state is tested matters: it can be the end of the window
or its beginning. For a
terminal state \(z_T=G_f(s_T)\), the hypothesis is that
\(O_T=\bar O_T\circ G_f\). For a preparation history \(h\), let
\(z_0=\psi(h)\) be the candidate summary before a continuation \(v\):
a fixed input sequence, or fixed external response and observation rules
for the fixed adaptive policy in Section~\ref{sec:adaptive-response}. Write
\(R_{T,v}(h)\) for the
readout obtained by evolving that history through the window. A readout
function \(H_{T,v}\) depending only on the summary would satisfy
\[
R_{T,v}(h)=H_{T,v}(\psi(h)).
\]
It asks whether the preparation summary preserves the distinctions needed for
that future readout. It does not assert that the summary evolves autonomously.
For a stochastic system, replace readout equality by equality of the
specified statistic or conditional output distribution. Comparisons always
hold \(v\) and the system fixed. In the running example, the pending-request
snapshot is tested before the returns arrive, rather than against an output
already stored at the endpoint.

\subsection{Inverse analysis: is the retained information sufficient?}
Exact sufficiency requires histories with the same candidate state to give
equal specified readouts or output laws under common continuations. A resolved
difference rejects sufficiency. Finite comparisons can instead bound agreement
at a specified resolution or leave adequacy unresolved.

For differentiable responses, a first-order test asks whether a source direction
invisible to the state remains visible to the readout.
On a finite-dimensional measured source space \(\mathcal U_T\), let \(\mathcal K_{z,T}\) and
\(\mathcal K_{O,T}\) be the first-order source-to-state and source-to-endpoint
maps. For a linear map \(K\), \(\ker K\) contains inputs mapped to zero,
and \(\operatorname{Im}K\) contains its possible outputs. A linear
readout map \(\mathcal H_T^{(1)}\), with superscript \((1)\) denoting
first order, exists exactly when
\begin{equation}
\bigl[\exists\,\mathcal H_T^{(1)}:\,
\mathcal K_{O,T}=\mathcal H_T^{(1)}\mathcal K_{z,T}\bigr]
\quad\Longleftrightarrow\quad
\ker\mathcal K_{z,T}\subseteq\ker\mathcal K_{O,T}.
\label{iclr:eq-kernel}
\end{equation}
The map is unique on \(\operatorname{Im}\mathcal K_{z,T}\), the reached state
displacements. To challenge the representation with this test, the design must
include a source direction that leaves the state unchanged to first order.
Such a direction lies in \(\ker\mathcal K_{z,T}\) and requires
\(\dim\mathcal U_T>\operatorname{rank}\mathcal K_{z,T}\). At full column
rank there is no such direction, and the factorization holds automatically.

First-order agreement alone does not establish agreement for finite changes.
For continuously differentiable (\(C^1\)) maps between finite-dimensional
coordinate spaces, constant rank of the source-to-state map and kernel inclusion
throughout a neighborhood give local nonlinear factorization. Without constant
rank, \(z(u)=u^2\), \(O(u)=u^3\) obey the pointwise derivative criterion
but merge opposite inputs with different endpoints. Exact neighborhood
sufficiency instead requires equal retained states to imply equal endpoints;
stochastic readouts require repeated controlled continuations and an
appropriate distributional comparison. The proof and analytic diagnostic controls appear in
SI Section~\siref{si:falsifiable-retained-state}; the delayed-history
counterexample is given in Section~\ref{iclr:physics}.

Apply the criterion after checking response resolution and, for a first-order
claim, step refinement. The design must supply a resolved direction in the
state-response kernel or a controlled same-state comparison.
For approximately matched states and a deterministic readout, let
\(\varepsilon_z\) bound the true state difference and \(\varepsilon_O\) the
error in the measured readout contrast \(\widehat{\Delta O}\). A candidate
readout \(O=h(z)\), with independently justified Lipschitz bound \(L_h\)
(at most \(L_h\) units of output change per unit of state change),
requires
\begin{equation}
\|\widehat{\Delta O}\|\le L_h\varepsilon_z+\varepsilon_O.
\label{eq:public-matching-bound}
\end{equation}
The right-hand side is the output change allowed by imperfect state matching
and measurement error. A violation rejects this bounded readout class;
with exact matching, it rejects the representation for the compared histories.
For an approximate match, check the sensitivity assumption as well as the state.
Local support requires adequate comparison coverage, contrast intervals within
fixed equivalence bounds and a detectable positive control: a deliberately
varied input whose readout change is resolved beyond measurement uncertainty.
Stochastic readouts require corresponding statistical or
distributional bounds. Unresolved comparisons remain undecided; different
directions and readouts can receive different judgments.

Equivalence margins and positive-control thresholds are study-specific design
choices, fixed before the observations used for those decisions. The response
framework supplies no universal numerical values for them. Sample size, output
alphabet and simultaneous error allocation determine the attainable precision
of a chosen confidence bound. Even identical empirical distributions can remain
unresolved when that precision is insufficient for the chosen agreement margin.
For categorical outputs, total variation (TV) is the largest difference in
probability assigned to an output event. A matched-state pair has bounded
agreement when its TV upper bound is at most the study's tolerance; a lower
bound above that tolerance rejects it; overlapping bounds leave it unresolved.
Informative support also requires the study's coverage and positive controls.
Methods selected after observing responses provide exploratory evidence and
leave the original decision classifications unchanged.

Table~\ref{tab:response-judgments} translates these conditions into experimental
choices. Repeated measurements reduce uncertainty, but cannot supply a direction in the state-response
kernel or a matched-history pair absent from the design.

\begin{table}[!t]
\centering
\caption{\textbf{Response comparisons guide experimental design.}
Judgments concern a specified readout, perturbation class and window;
state-matching error and measurement uncertainty must be bounded. First-order
claims additionally require slopes consistent across perturbation sizes.
Judgments may coexist across directions. Stochastic readouts require bounds on
differences in the chosen statistic or conditional output distribution.
}
\label{tab:response-judgments}
\begin{tabularx}{\linewidth}{@{}>{\raggedright\arraybackslash}p{0.36\linewidth}>{\raggedright\arraybackslash}p{0.25\linewidth}Y@{}}
\toprule
\textbf{Observation} & \textbf{Supported conclusion} & \textbf{Next experiment} \\
\midrule
Adequate direction coverage; readout-difference bounds, including errors, within
the fixed equivalence margin; detectable positive control.
& Bounded support on the tested comparisons at the declared resolution.
& Extend directions, readouts or windows to test the scope of support. \\
\addlinespace[0.35em]
Exactly equal retained states give a readout difference beyond measurement error.
& Reject sufficiency for these histories under the declared conditions.
& Separate the violating histories in a revised representation; test it with
system and interface unchanged. \\
\addlinespace[0.35em]
Approximately matched states violate the independently justified bound combining
readout sensitivity and measurement error.
& Reject the bounded readout class, not the representation alone.
& Check the readout class and independent sensitivity bound; tighten state matching. \\
\addlinespace[0.35em]
Neither an informative direction in the state-response kernel nor a
matched-state comparison is resolved.
& The design leaves adequacy unresolved.
& Change source directions or compare different histories at matched states. \\
\addlinespace[0.35em]
Response is detected; slopes fail the step-refinement tolerance.
& Finite contrasts; first-order interpretation is unsupported at these steps.
& Test finer steps or another readout; recheck resolution. \\
\addlinespace[0.35em]
A task-required contrast is absent despite preserved input records.
& Test how the program or model uses the supplied records.
& Compare changes to the program or model with controls; remeasure state and readout. \\
\bottomrule
\end{tabularx}
\end{table}

\section{Forward response in physical systems}
\label{iclr:physics}
Known physical dynamics make the forward question directly testable: does the
measured response agree with the response calculated from the equations of motion?
A field with local memory and delayed sources supplies the dynamical setting;
stationary susceptibility follows when additional closure and stability
conditions hold. Simulated delayed consensus calibrates amplitude and phase
against an analytic prediction, while a polar flock compares nonlinear
steady-state readouts. These tests use standard delay-system and active-matter
models~\citep{hale1993,olfati2004,vicsek1995,toner1995,marchetti2013}.
The shared response description fixes the roles of source, retained state
and readout; each system supplies its own equations and noise law.

\subsection{History-dependent state and dynamics}
Neural-population and neural-field models describe collective activity through
local interactions and spatial coupling
\citep{wilson1972excitatory,amari1977dynamics,bressloff2012spatiotemporal}.
To include local memory and delayed sources, consider the controlled
semilinear field
\begin{equation}
\partial_t c_t=Lc_t+\sum_{i=1}^{N}J_i(c_t,m_i(t))
             +\sum_{\ell=1}^{N_\tau}E_\ell y_\ell(t-\tau_\ell).
\label{eq:anm-universal-interface}
\end{equation}
Here \(c_t\in L^2(X,\mathbb R^q)\) has \(q\) channels on a spatial or graph
domain \(X\). The operator \(L\) describes how the field spreads or relaxes
between source contributions; a diffusion--decay operator is one example.
The \(N\) local sources \(J_i\) depend on the field and local memories
\(m_i\). Each of the \(N_\tau\) external channels injects a prescribed
signal \(y_\ell\) through \(E_\ell\) after delay \(\tau_\ell\).

The local source can itself be resolved into physical operations. Let \(R_i\)
read the field and memory, \(\pi_i\) compute a local update, \(W_i\) convert
that update to a source, and \(G_i\) place it into the field's sites and
channels. Then
\begin{equation}
J_i(c,m_i)=G_iW_i\!\left(\pi_i(R_i(c,m_i))\right).
\label{eq:public-local-source}
\end{equation}
This separates injection from propagation. A physical component senses and
drives the medium; a workflow component reads a record and writes an update.
Each realization specifies these maps. The coupled response includes changes
in field-dependent sources and memories as well as propagation generated by
\(L\).

Delay determines what must be included in the state. In consensus with coupling
delay \(\tau\), retain the history segment
\(\xi^t(\theta)=c(t+\theta)\), \(-\tau\le\theta\le0\), together with any
local memories. Two preparations can agree at \(c(t)\) while their past
fields produce different forcing later in the observation window. This internal
feedback belongs to the modeled dynamics, with its history suppressed in the
displayed arguments of \(J_i\). The external-input sum instead describes
signals whose generators lie outside the modeled system. In a delayed-tool
workflow, earlier requests similarly supply context for later returns;
whether a particular summary preserves that context is a representation
question.

The inverse criterion exposes what a present-value description can lose.
For the delay equation
\[
\dot x(t)=-a x(t)-k x(t-\tau)+u(t),\qquad k\ne0,\quad\tau>0,
\]
let the candidate state retain only \(x(t)\). Here \(a\) is the local
relaxation coefficient, \(k\) the delayed coupling, and \(u\) the applied
input. Two allowed history preparations with the same current value but
different \(x(t-\tau)\), under the same subsequent input, have different
current derivatives. In tangent form,
\(\delta x(t)=0\) and \(\delta x(t-\tau)\ne0\) give
\(\delta\dot x(t)=-k\delta x(t-\tau)\ne0\). The readout
\(O=x(t+\Delta)\) therefore separates the preparations for sufficiently
small \(\Delta>0\). The present value omits a response-relevant history distinction.
In the delayed-tool example of Section~\ref{sec:worked-tool}, the corresponding
comparison holds a snapshot fixed while changing which earlier request a
later return answers. Both examples test history compression using the
inverse criterion of Section~\ref{iclr:protocol}.

SI Section~\siref{si:formal-delayed-model} specifies the source and memory
maps; SI Sections~\siref{si:reduced-state-equation}
and~\siref{si:well-posedness-coherence} give the history dynamics and
existence conditions. Finite-window measurements follow this augmented state
directly. Autonomous reduction and steady susceptibilities add the closure and
stability assumptions stated next.

\subsection{From finite-window response to stationary susceptibility}
The finite-window response also connects to the familiar displacement of a
stable operating state under a persistent source. Suppose the retained
coordinates admit a closed, finite-dimensional equation
\(\dot z=f(z,u)\), with \(f\) continuously differentiable, and let
\(f(z_\ast,u_\ast)=0\). Controlled closure means that these coordinates determine their evolution
for the allowed inputs on a common invariant domain, with compatible
initialization. Invariance at each fixed input alone does not supply this
property; explaining one endpoint readout is also a separate condition. Define the drift
Jacobian \(A_z=\mathrm D_z f(z_\ast,u_\ast)\) and source Jacobian
\(B_z=\mathrm D_u f(z_\ast,u_\ast)\). If all eigenvalues of \(A_z\) have
negative real part, a constant tangent source \(\delta u\), starting from
zero initial displacement, gives
\begin{equation}
\delta z(t)=\int_0^t e^{A_z(t-s)}B_z\delta u\,\mathrm ds
\ \longrightarrow\ \chi_\ast\delta u,
\qquad \chi_\ast=-A_z^{-1}B_z.
\label{eq:public-stationary-susceptibility}
\end{equation}
Each contribution is injected by \(B_z\) and propagated by the linearized
dynamics; stability makes their accumulated effect converge. For a
differentiably factored readout \(O=\bar O(z)\), its stationary response is
\(\delta O=(\mathrm D_z\bar O)\chi_\ast\delta u\).
A time-independent discrete transition with Jacobians \(A,B\) and spectral
radius \(\rho(A)<1\) gives the corresponding sum
\(\chi_\ast^G=\sum_{k\ge0}A^kB=(I-A)^{-1}B\).
With exact discretization under an input held constant during each time step,
the two susceptibilities agree (SI Section~\siref{si:reduced-state-equation}).
Thus finite-window propagation and stationary susceptibility belong to the
same response description under their respective assumptions. For an AI
realization, the stationary interpretation requires the same closure and
stability conditions; finite contrasts remain applicable without that
interpretation.

\subsection{Operating boundaries depend on the readout}
\label{iclr:conditions}
An operating criterion specifies acceptable values of the scalar or vector
readout \(O_T\) defined in Section~\ref{iclr:protocol}: a threshold for one
observable, or a region for several, such as amplitude, phase and spatial
variance. The requirement defines the boundary; dynamics determine whether
loading carries the readout across it. Readout dimension and units follow the
measurements and need not match retained-state dimension. Neither two components
nor nesting is required.

\label{si:observables-directions}
\label{si:nested-observables}
To distinguish continued operation from a stricter endpoint criterion,
fix an experimental family \(f\): its instances, interface, stopping rule,
loading coordinate and readouts. The loading coordinate parametrizes the
conditions under which the readouts are compared. When the two readouts factor
through a candidate state \(z\in Z_f\), write
\[
P_f,Q_f:Z_f\to[0,1],\qquad 0\le Q_f(z)\le P_f(z)\le1.
\]
Here \(P_f\) represents the outer criterion for continued operation and \(Q_f\) a
stricter condition contained within it. Nesting concerns the meanings of the
conditions; numerical ordering alone does not establish it. Their reduced
coordinates are
\begin{equation}
w_f:=P_f,\qquad r_f:=Q_f/P_f\quad\text{where }P_f>0.
\label{iclr:eq-nested}
\end{equation}
Thus \(w_f\) measures remaining workability and \(r_f\) the inner readout
relative to the outer one. The ratio is a conditional correctness fraction
only for nested measures on the same population. Criteria that do not factor
through the candidate state remain readouts of the full measured system.

Write \(w=w_f\), \(r=r_f\) for one family.
The four regimes are exact conditional on workability (\(w>0,\ r=1\)),
partially exact (\(w>0,\ 0<r<1\)), workable but inexact (\(w>0,\ r=0\)),
and failed (\(w=0\), with \(r\) undefined).
At \(r=1\) the two readouts are equal; \(w\) separately records workability.
Measured boundary assignments require stated uncertainty and tolerances.
Readout level sets define boundaries; dynamics determine how a loading path
approaches them. Encountering all four regimes in order requires additional
regularity, monotonicity and separated-crossing assumptions. Here monotone
loading excludes re-entry into a departed regime; the path and loading
coordinate are specified before inspecting the crossing order. Coincident
thresholds can skip regimes, and nonmonotone recovery remains compatible
with nested readouts (SI Section~\siref{si:local-readout-regimes}).

\begin{figure}[!htb]
\centering
\includegraphics[width=\linewidth]{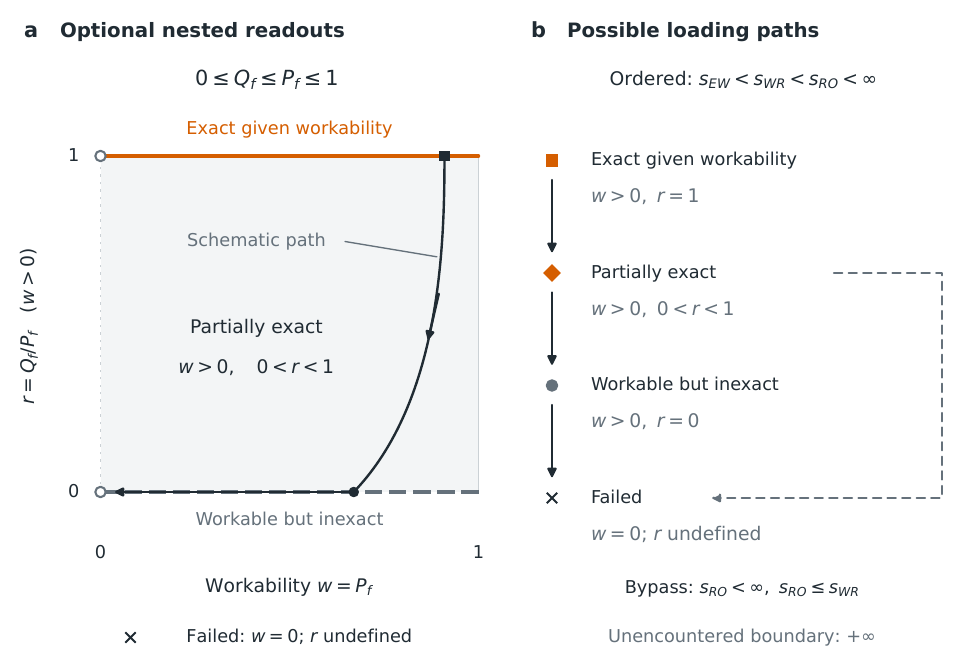}
\caption{\textbf{Readout-regime map and boundary order under monotone loading.}
\textbf{a}, The coordinates are workability \(w=P_f\) and the readout ratio
\(r=Q_f/P_f\) for \(P_f>0\). The four local readout regimes comprise
one open partially exact region, the line \(r=1\) (exact given workability),
the line \(r=0\) (workable but inexact), and the failed case \(w=0\). Hollow endpoints exclude
\(w=0\) from the ratio chart: failure is specified without a value of \(r\).
The plotted path is schematic. SI Section~\siref{si:reported-coordinates-controls}
discusses retrospective coordinate examples, which are not encoded by this path. \textbf{b}, Along a specified monotone loading path whose
boundary encounters cross between adjacent readout regimes, the
four-step sequence holds when all three encounters occur with
\(s_{EW}<s_{WR}<s_{RO}\). Here \(s_{EW}\) is the first encounter with
the exact/partially exact boundary, \(s_{WR}\) with the partially
exact/workable-but-inexact boundary, and \(s_{RO}\) with the workable/failed
boundary; an unencountered boundary has parameter \(+\infty\).
If \(s_{RO}<\infty\) and \(s_{RO}\leq s_{WR}\), the path can pass directly
from the partially exact region to the failed boundary. Nested readouts fix the
allowed boundary cases but do not determine their encounter order.}
\label{fig:reduced-frontier}
\end{figure}

Source perturbations can change either readout; their responses are measured
separately. The readout pair determines neither state dimension nor
sufficiency, and response magnitude alone does not measure intervention cost.
Figure~\ref{fig:reduced-frontier} separates the allowed regimes from the
additional assumptions needed for an ordered sequence of boundary crossings.

Nested readouts can be same-population fractions or single-trial indicators.
A binary indicator is not differentiated: response measurements use finite
contrasts or differentiable observables, such as amplitude and phase, with
thresholds calibrated to the binary criteria. SI
Section~\siref{si:local-readout-regimes} gives the boundary definitions and
conditional crossing results.

In the consensus experiment below, \(P_f\) indicates continued admissibility
and \(Q_f\) its stricter endpoint criterion. In the flock, the same
family-local labels denote polarization and angular alignment: these are
distinct, nonnested readouts. The distinction allows
the same response framework to compare both, without assigning the nested
regime labels to the flock.

\subsection{Calibrating predictions in delayed consensus}
A sinusoidal drive supplies an analytic reference for measured amplitude and
phase. The experiment checks this response over a narrow range of coupling
delays near one operating point.
This weak-delay calibration tests the driven response. Establishing history
necessity from measurements would additionally require preparations with equal
current values and different histories.

Let \(c_i(t)\) be the scalar activity at site \(i\) of a periodic,
degree-four ring with \(N=64\) sites. Each site relaxes at rate \(a\),
receives its neighbors' earlier activity with coupling strength \(K\) and
delay \(\tau\), and is driven by a uniform sinusoidal source of amplitude
\(F\) and angular frequency \(\Omega\), with site-noise amplitude \(\sigma\):
\begin{equation}
\dot c_i(t)
= -a\,c_i(t)
+ \frac{K}{|N_i|}\sum_{j\in N_i}c_j(t-\tau)
+ F\cos(\Omega t)
+ \sigma\xi_i(t).
\label{eq:supp-driven-consensus}
\end{equation}
Here \(N_i\) is the set of four neighbors and \(\xi_i\) are independent
unit-intensity white noises. The delay calibration uses \(a=1.0\),
\(K=0.40\), \(F=0.25\), \(\Omega=0.35\) and \(\sigma=0.015\),
with \(\Omega\tau=0.1505\)--\(0.1610\).

The ring averaging matrix is symmetric with unit row and column sums.
Consequently the spatial mean \(m=N^{-1}\sum_i c_i\) closes exactly,
including the projected noise:
\begin{equation}
\dot m(t)=-a\,m(t)+K\,m(t-\tau)+F\cos(\Omega t)+\eta_m(t),
\qquad
\eta_m(t)\equiv\frac{\sigma}{N}\sum_{i=1}^{N}\xi_i(t).
\label{eq:supp-consensus-mean-eq}
\end{equation}
This closure removes the transverse spatial modes from the mean equation,
but retains its delay history. For the stated parameters, \(a>|K|\)
ensures delay-independent stability of the normalized network modes.
After transients, the deterministic component or ensemble mean has the form
\(m_{\rm det}(t)=\operatorname{Re}\{F\chi(\Omega,\tau)e^{i\Omega t}\}\),
where harmonic balance gives
\begin{equation}
\chi(\Omega,\tau)=\bigl(a-Ke^{-i\Omega\tau}+i\Omega\bigr)^{-1}.
\label{eq:consensus-harmonic-susceptibility}
\end{equation}
The factor \(e^{-i\Omega\tau}\) is the phase shift acquired by the
delayed coupling. It changes both the amplitude and phase of the driven
response. This frequency-dependent susceptibility complements the
constant-source, stable-branch limit in
Eq.~\eqref{eq:public-stationary-susceptibility}.

To compare this prediction with a finite-window measurement, a lock-in
estimator extracts the part of the mean oscillating at the drive
frequency over a window of duration \(T_{\rm win}\). The readout combines
its amplitude and phase lag with the transverse spatial variance:
\begin{equation}
\begin{gathered}
\hat Z=\frac{2}{T_{\rm win}}\int_{\rm window}m(t)e^{-i\Omega t}\,dt,
\qquad \boldsymbol y_{\rm cons}=(\mathcal A,\phi,V),\\
\mathcal A=|\hat Z|,\qquad \phi=|\arg\hat Z|,\qquad
V=\frac{1}{NT_{\rm win}}\int_{\rm window}
\sum_i\bigl(c_i(t)-m(t)\bigr)^2\,dt.
\end{gathered}
\label{eq:consensus-window-readout}
\end{equation}
Amplitude and phase measure the coherent response; \(V\) records spatial
fluctuations that the mean discards. These three observables summarize a
trajectory, rather than forming an autonomous three-dimensional state.
The harmonic targets are
\begin{equation}
\begin{aligned}
\mathcal A_{\rm th}(\tau)
&=F\left[(a-K\cos\Omega\tau)^2+(\Omega+K\sin\Omega\tau)^2\right]^{-1/2},\\
\phi_{\rm th}(\tau)
&=\operatorname{atan2}\bigl(\Omega+K\sin\Omega\tau,\;a-K\cos\Omega\tau\bigr).
\end{aligned}
\label{eq:consensus-harmonic-targets}
\end{equation}
For a local calibration over the narrow interval \(\tau\in[0.43,0.46]\),
100 noise realizations at each of 13 delays supply
the mean amplitude and phase compared with these targets
(Table~\ref{tab:phys-anm-compact}). Most of the phase discrepancy comes from the conjugate-frequency contribution
within the finite rectangular window. SI Section~\siref{si:consensus-network}
separates this bias from the numerical correction and gives the update, initial
history, seeds and provenance limits. Spatial concentration compares squared
peak amplitude with transverse variance; it does not determine state dimension.

\begin{table}[htbp]
\centering
\caption{\textbf{Consensus response errors and criterion crossings.}
Response errors compare 100-seed means with harmonic targets (maximum over
13 delays in \([0.43,0.46]\)), including finite-window estimator bias;
relative amplitude error uses the harmonic amplitude as denominator.
\(\mathcal A\) is peak amplitude and \(V\) transverse variance.
Drive \(F\) and noise \(\sigma\) use model units. Subscripts \(P,Q\) denote
admissibility and its stricter amplitude/phase criterion, respectively.
Crossing values interpolate where the fraction of the same 50 seeds satisfying
each criterion crosses one half; they are sample statistics, not identified
population boundaries or critical points.}
\label{tab:phys-anm-compact}
\begin{tabularx}{\linewidth}{@{}p{0.27\linewidth}Y@{}}
\toprule
\textbf{Comparison} & \textbf{Recorded result} \\
\midrule
Coherent response & Relative amplitude error \(\le2.51\times10^{-3}\);
absolute phase error \(\le0.74^\circ\). \\
\addlinespace[0.25em]
Spatial concentration & \(\mathcal A^2/(\mathcal A^2+V)\ge0.999\);
transverse variance is small relative to squared peak amplitude, without
determining state dimension. \\
\addlinespace[0.25em]
Decreasing drive & \(F_Q\approx0.163\), \(F_P\approx0.118\). \\
\addlinespace[0.25em]
Increasing noise & Sample crossings \(\sigma_Q\approx0.138\),
\(\sigma_P\approx0.310\); the early \(Q_f\) crossing is consistent
with finite-sample phase fluctuations. \\
\bottomrule
\end{tabularx}
\end{table}

Predicting a response and judging acceptable operation are separate questions.
Retrospective sweeps compare continued admissibility with a stricter
amplitude/phase requirement as drive or noise changes.
At \(\tau=0.43\), both sweeps hold the original criteria fixed:
\begin{equation}
\begin{aligned}
P_f&=\mathbf 1\!\left[
  \text{stable},\ \mathcal A\ge0.16,\ V\le0.05\right],\\
Q_f&=P_f\,\mathbf 1\!\left[
  \mathcal A\ge0.22,\ \phi\le35^\circ\right].
\end{aligned}
\label{eq:consensus-operating-criteria}
\end{equation}
Here ``stable'' denotes finite-valued numerical integration through the
horizon with a nonempty observation window; it is distinct from the analytic
stability condition \(a>|K|\). The stricter criterion requires both a larger
coherent response and a small phase lag. The sweep denominator comprises
26 control values with 50 matched seeds each, separately from the delay
calibration. All 1,300 recorded sweep trajectories satisfy the numerical
condition, while 442/1,300 remain admissible but fail the stricter endpoint
criterion (\(Q_f=0<P_f\)). This counts trajectories between the two operating criteria.

The noise sweep illustrates why a sample crossing must be distinguished from
a population boundary. All 179 trajectories with \(P_f=1,Q_f=0\) fail the
phase criterion. Yet at \(\sigma=0.15\), where only \(24/50\) seeds pass
that criterion, the linear Gaussian dynamics and numerical estimator give a
population phase-pass probability of approximately \(0.583\). It remains above
one half throughout the recorded noise range. Reusing the same noise
realizations strongly correlates neighboring measurements, so their early
crossing \(\sigma_Q\approx0.138\) is consistent with finite-sample phase
fluctuations. The complete \(Q_f\) criterion can still cross one half when
its variance constraint becomes active. SI Section~\siref{si:consensus-network}
gives the complete counts, pointwise intervals and conditional bootstrap
analysis; these quantify sampling variation without identifying a population
phase boundary.

\begin{figure}[!htbp]
\centering
\includegraphics[width=\linewidth]{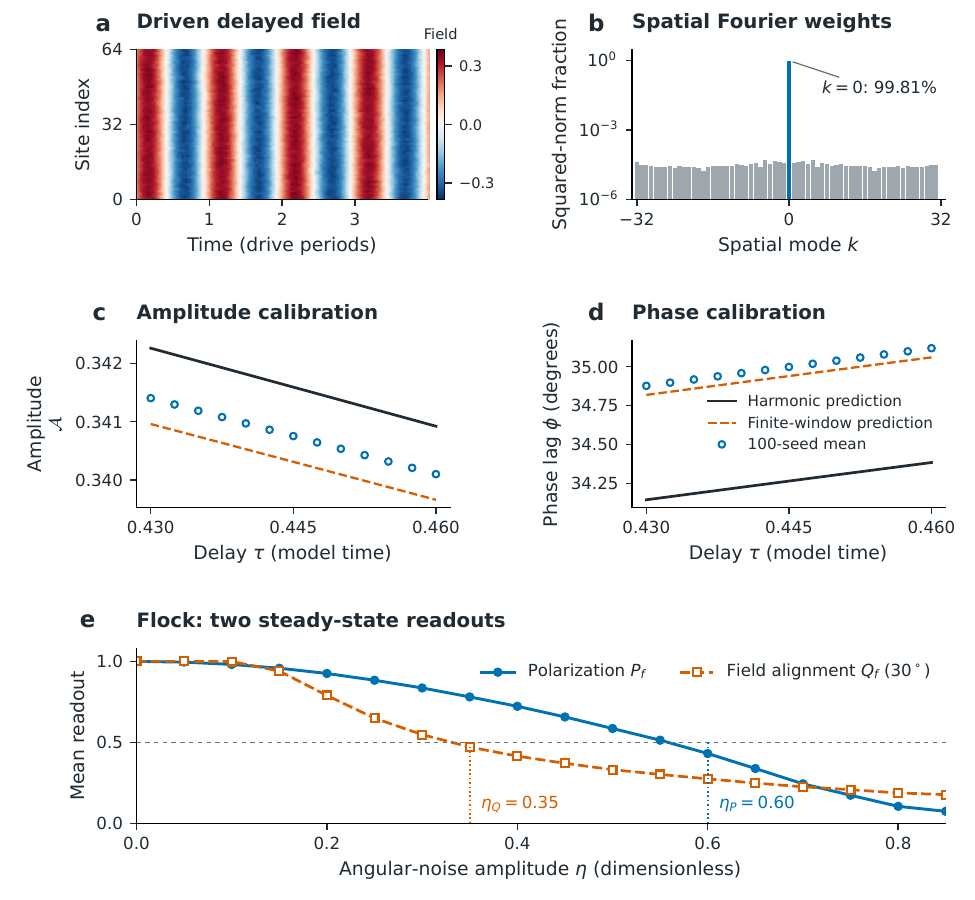}
\caption{\textbf{Physical response calibration and complementary steady-state readouts.}
\textbf{a}, Time-centered field on a 64-site delayed ring over four drive
periods; color is in model units. \textbf{b}, Fractions of its squared norm
in spatial Fourier modes, with \(99.81\%\) in the uniform mode \(k=0\).
Panels a,b use one seed at \(\tau=0.46\); these weights differ from the
peak-amplitude and quadrature weights in SI Section~\siref{si:consensus-network}.
\textbf{c,d}, The separate 13-delay calibration compares 100-seed means
(circles), harmonic predictions (solid lines) and the finite-window estimator
applied to the harmonic signal (dashed lines). Relative to the harmonic target,
amplitude and phase have
maximum relative and absolute differences \(2.51\times10^{-3}\) and
\(0.74^\circ\), respectively. The displayed differences include finite-window
estimator bias, decomposed in SI Section~\siref{si:consensus-network};
no uncertainty interval is inferred from the plotted means.
\textbf{e}, Four-seed mean polarization \(P_f\) and field alignment within
\(30^\circ\), \(Q_f\), in the driven Vicsek flock. Guides mark one half
and the first sampled noise levels below it, \(\eta_Q=0.35\) and
\(\eta_P=0.60\). The readouts are nonnested, despite numerical ordering
on \([0.15,0.70]\); the sweep is not a boundary trajectory. Hollow squares
expose coincident values without offsets. Shading spans the minimum and maximum
of the four seed means; it is not a confidence interval and excludes
within-run fluctuations.}
\label{fig:physical-realizations}
\end{figure}

\subsection{Comparing readouts in a polar flock}
The consensus criteria are nested by definition. A driven polar flock permits
comparison of distinct criteria without that restriction. Increasing angular noise tests how coherent motion and field alignment differ
in a nonlinear steady-state response. Polarization measures agreement among
particle headings; field alignment measures agreement with an external direction.

In this realization, \(N=256\) particles move at constant speed
\(v_0=0.3\) in a periodic square of side \(L=8\). Their positions determine
the neighbors \(N_i\) within radius \(r_{\rm int}=1\), including the
focal particle. Headings \(\theta_i(t)\) turn toward the local vector sum
and a fixed field of strength \(h=0.6\) and direction
\(\theta_{\hat e}=0\), subject to angular noise of amplitude \(\eta\):
\begin{equation}
\theta_i(t+1)=\operatorname{Arg}\!\Big(\sum_{j\in N_i}e^{\mathrm{i}\theta_j(t)}
+ h\,e^{\mathrm{i}\theta_{\hat e}}\Big)+\eta\,\xi_i(t),
\qquad \xi_i\sim\mathcal U(-\pi,\pi).
\label{eq:vicsek-field}
\end{equation}
Positions advance with the updated headings after each unit time step.
The angular noises are independent across particles and steps. Their
amplitude \(\eta\) is varied while the field stays fixed. Thus positions
and headings supply the evolving state, local interactions propagate
alignment, and angular noise competes with collective order. This model
uses instantaneous neighbor headings; unlike consensus, its update has no
explicit transport delay.

Let \(\delta_i(t)=\operatorname{Arg}e^{i(\theta_i(t)-\theta_{\hat e})}
\in(-\pi,\pi]\) be the wrapped angle from the field and let
\(\mathrm{tol}=30^\circ=\pi/6\) be the alignment tolerance. Instantaneous
polarization and aligned population fraction are
\begin{equation}
p(t)=\left|\frac{1}{N}\sum_{i=1}^N e^{\mathrm{i}\theta_i(t)}\right|,
\qquad
q(t)=\frac{1}{N}\sum_{i=1}^N
\mathbf 1\!\left[|\delta_i(t)|\le\mathrm{tol}\right].
\label{eq:flock-instantaneous-readouts}
\end{equation}
The reported readouts average each quantity over time and then four seeds:
\begin{equation}
P_f=\left\langle\langle p(t)\rangle_{\rm time}\right\rangle_{\rm seed},
\qquad
Q_f=\left\langle\langle q(t)\rangle_{\rm time}\right\rangle_{\rm seed}.
\label{eq:flock-averaged-readouts}
\end{equation}
The modulus in \(p(t)\) precedes averaging. A coherent flock can turn away
from the field while keeping polarization near one; \(Q_f\) instead asks
how many particles point within the chosen tolerance. The readouts are
therefore nonnested. Their numerical order on \(\eta\in[0.15,0.70]\)
fails outside that interval, and \(Q_f/P_f\) is not a conditional fraction.
The sampled half-values in Figure~\ref{fig:physical-realizations}e quantify
threshold-dependent separation, not a critical point, boundary trajectory or
slow mode. Averaged values do not distinguish heading broadening from
collective-direction fluctuations. SI Section~\siref{si:vicsek-flock} gives
sampling, tolerance comparisons and the angular-distribution approximation.
Together, the studies calibrate driven consensus near one operating point and
compare flock readouts; history necessity follows from the analytical delay
example.

\FloatBarrier
\section{Forward response in an AI event graph}
\label{iclr:ai-evidence}
\label{si:worked-graph}
The physical studies predict responses from known dynamics. In an AI workflow,
propagation through a prescribed graph can likewise be calculated, while the
language model's response to that graph must be measured. This forward test
separates graph propagation from readout response and tests a first-order
description at the measured perturbation sizes.

Here a language model extracts events from visible records, and fixed rules
map the checked events to a numerical graph retaining support and opposing
evidence for candidate values. Separate model calls read a table derived from
its state. Both model roles use Qwen2.5-14B-Instruct~\citep{qwen2024qwen25}
with MLX 4-bit quantization.
Measurement fixes the extracted graph, perturbs a source amplitude and compares
the propagated state and model support for candidate values over a fixed window.
The response is conditional on this graph; changing source text and repeating
extraction is a different experiment.

\subsection{From evidence records to graph dynamics}
To make the source encoding explicit, consider a constructed telescope task:
choose an instrument, filter and observation block for a calibration plan.
Instrument labels are alpha, beta and sigma; filters are broad, hydrogen-alpha
and infrared; block labels are A, B and C. Three evidence statements specify beta's stable
lamp, the hydrogen-alpha filter and block C's matching exposure. A fourth
repeats infrared and block B while explicitly excluding them. These are
authored constraints, not astronomical observations.

The extractor receives the task, allowed values and evidence statements.
Each event specifies its source
record, update time \(t_d\in\{0,1,2,3\}\), type, candidate targets and a
confidence score. Checks enforce valid record references, times, types,
targets and confidence values; they do not verify semantic accuracy.
The event type determines the outgoing edges. A support
event feeds a candidate's support coordinate; contradiction and inhibition
events feed its opposing-evidence coordinate; a coupling event divides its
contribution equally between two candidates' support coordinates.
Contradiction and inhibition therefore share the same numerical action in
this implementation. The graph builder, rather than the language model,
sets edge weights from these rules. Confidence is recorded but does not set
the imposed source amplitude. SI Section~\siref{si:source-encoding} defines
this general map from records to numerical sources.

The telescope extraction illustrates why format checks and faithful evidence
interpretation are separate. It assigns support to beta, hydrogen-alpha and C
at times 0, 1 and 2, but encodes the explicitly excluded infrared/B pair as a
coupling at time 3. This recorded extraction remains fixed during the response
experiment.

The event assignments fix transport paths. The general transition \(F_t\)
permits heterogeneous, time-dependent or nonlinear updates, with required
memory retained (Section~\ref{iclr:protocol}). This experiment uses a fixed
graph with uniform retention and transport gains.
For \(N_e\) extracted events, the numerical state \(z_t\in\mathbb R^{N_e+18}\)
contains one coordinate per event and two per candidate: support and opposing
evidence for each of the nine allowed field--value pairs. The telescope
instance has \(N_e=4\) and hence 22 coordinates. Let \(W\) be the fixed
transport matrix: \(W_{ij}\) is the weight from source coordinate \(j\) to
target coordinate \(i\). An event's outgoing weights sum to one; candidate
coordinates have no outgoing edges. Let \(e_d\) select event \(d\)'s coordinate,
\(u_d\) be its imposed amplitude, and \(b_t\) the total deposit at update \(t\).
Let \(\alpha\in(0,1)\) be the retention factor, \(\beta\ge0\) the transport
gain, and \(I\) the identity. Starting from \(z_0=0\), the updates are
\begin{equation}
 b_t=\sum_{d:t_d=t}u_d e_d,\qquad
 z_{t+1}=\mathcal C\!\left[(\alpha I+\beta W)(z_t+b_t)\right].
 \label{eq:ai-graph-update}
\end{equation}
The dimensionless gains specify retention at each coordinate and transfer
along edges; they are not complementary probabilities. The numerical
range-control map \(\mathcal C\) divides by the
largest coordinate when it exceeds one, then clips coordinates to \([0,1]\).
The reported experiment fixes \(\alpha=0.72\), \(\beta=0.45\), and four
updates \(t=0,1,2,3\). Neither range-control operation activates on its measured
perturbation trajectories. Every baseline amplitude is \(u_d=0.5\), independently of extractor confidence;
these amplitudes are experimental inputs, not truth probabilities. Thus propagation along these trajectories is linear with
\(A=\alpha I+\beta W\). An amplitude perturbation enters through
\(Ae_d\), because deposition precedes the update, separating source
injection from propagation (Section~\ref{iclr:protocol}). SI Section~\siref{si:live-paired-source-perturbation}
gives its parameter-dependent state response. The measured model-readout
results below concern the fixed coefficient pair; parameter robustness
requires comparisons across coefficient settings.

Equation~\eqref{eq:ai-graph-update} instantiates the finite-window response
construction: the source amplitude enters a known transition, and its terminal
effect is measured separately from the reader's response. Its update index
counts computational steps. The dynamics are prescribed and open-loop,
without readout feedback or a claimed discretization of the continuous field
equation. The representation tests in Section~\ref{sec:worked-tool} use original
request--return records instead of this graph apparatus.

\subsection{Measuring propagation and model readout response}
To separate graph motion from model response, the reader receives the task,
one candidate and a table extracted from \(z_4\),
containing each candidate's support, opposing evidence and their difference.
It receives neither the event-source coordinates nor the original evidence
text. A fixed input transformation, chosen on separate development instances,
divides candidate contrasts by 16 before the reader is queried.
Let \(V_j\) be the three allowed values in field \(j\). Each displayed
quantity \(r_{jv}\), for value \(v\in V_j\), is mixed toward its
within-field mean \(\bar r_j\):
\begin{equation}
 r_{jv}^{(\lambda)}=(1-\lambda)r_{jv}+\lambda\bar r_j,\qquad
 \bar r_j=\frac{1}{3}\sum_{v\in V_j}r_{jv},\qquad \lambda=0.9375,
 \label{eq:ai-reader-input}
\end{equation}
The measured response is conditional on this setting.

A separate query asks whether each candidate is best supported. Write
\(p^Y_{jv}\) for the model's Y probability conditional on the two answer tokens
Y and N, and let \(v_j^\star\) be field \(j\)'s reference value, used only
in post-call scoring. At the terminal update \(T=4\), the scalar readout is
\begin{equation}
 O_T=\frac{1}{3}\sum_{j=1}^{3}
       \frac{p^Y_{jv_j^\star}}{\sum_{v\in V_j}p^Y_{jv}},\qquad T=4,
 \label{eq:ai-candidate-readout}
\end{equation}
Within each field, the denominator sums support across its three candidates;
the ratio is the share assigned to the reference value. The readout averages
these shares across fields; it is neither complete-plan accuracy nor a
calibrated joint-choice probability. Extraction
and graph construction are recorded before these reference values are used.

Step refinement tests whether the model readout has a consistent local slope.
Three development instances determine the reader-input transformation and
response-detection thresholds; four different instances supply the response
comparisons. For each measured graph, a fixed ordering selects three events
without using endpoint labels. After extraction, perturb one selected amplitude
at a time from 0.5 to \(0.5\pm\epsilon\),
first with \(\epsilon=0.1\), then with \(\epsilon/2=0.05\). Graph topology,
event timing, other amplitudes and paired reader settings stay fixed. Record
both the displacement of the complete graph state and the independently measured
readout slope; subscripts \(d,+\) and \(d,-\) indicate the increased and
decreased amplitude of event \(d\):
\begin{equation}
 \Delta z_d=z_{d,+}(4)-z_{d,-}(4),\qquad
 \widehat g_d(\epsilon)=\frac{O_{d,+}-O_{d,-}}{2\epsilon}.
 \label{eq:ai-paired-measurement}
\end{equation}
Paired candidate queries use matched seed settings. Repeated-input controls
check repeatability; routing controls check delivery of the numerical input.
The latter restore the baseline graph state or disable edge transport;
they check the numerical input path, not causal mediation.
The telescope's beta event illustrates the slope comparison. Amplitudes 0.4
and 0.6 give aggregate readouts
\(O_{d,-}\simeq0.8298\) and \(O_{d,+}\simeq0.9276\), as defined in
Eq.~\eqref{eq:ai-candidate-readout}. Their difference gives a slope of 0.489 per
unit source amplitude; the halved step gives 0.639. Dividing the absolute
slope difference by the larger slope magnitude gives 23.5\% disagreement,
within the fixed 25\% tolerance. This is one direction's readout
response, not an improvement in task accuracy.

The graph state responds in every tested direction, but the model readout
satisfies the first-order test in only some directions. Across four tasks, all twelve Euclidean displacements of the complete terminal
graph state exceed 0.05; nine readout slopes exceed 0.01 in magnitude, and two
pass the step-refinement criterion. Four of the nine above-threshold directions
reverse sign when the step is halved. In 355 of 432 perturbation calls
(82.2\%), the conditional probability of Y or N is at least 0.99.
These descriptive saturation counts and finite-step comparisons cannot
separate saturation from other nonlinear readout effects.

Within each three-direction design, no nonzero linear combination of tested
source contrasts preserves the complete graph state. The reader receives
evidence only through the candidate table. These measurements characterize
readout response after prescribed propagation, without a same-state comparison
that could expose omitted response-relevant information.
SI Section~\siref{si:live-paired-source-perturbation} gives sampling settings,
the exact refinement criterion, denominators, controls and saturation diagnostics.

\section{Inverse tests under fixed future inputs}
\label{sec:worked-tool}
The forward studies predict or measure responses for a chosen state. The
inverse question tests whether that state retains enough information: can
histories with the same candidate state give different responses to the same
future inputs? A request--return program first establishes exact sufficiency
for a sensor--recipient association state. Language-model comparisons then
test that state, its invariance to identifier and order changes, and a more
detailed candidate. Each comparison keeps its program or model fixed while
the analyst's description changes.

\subsection{Request associations distinguish identical snapshots}
\label{si:inverse-mwe}
The checkpoint test asks whether request associations can be omitted while
preserving later readouts.
Constructed histories and returns test a known message-processing program.
Each execution combines one history and continuation; matched-state histories
are compared under the same continuation. Selection and new executions test
refinement within a supplied candidate set, rather than model performance on
unknown tasks.

At the observation cut \(t=0\), two requests have been issued to logical sensor
channels \(S_1,S_2\). Each request associates its sensor with one recipient,
\(A\) or \(B\), and the association is bijective. A preparation history
\(H=(\sigma,\nu)\) consists of the association orientation \(\sigma\) and
the remaining request information \(\nu\). For \(\sigma=+1\), \(S_1\)
is associated with \(A\) and \(S_2\) with \(B\); for \(\sigma=-1\),
these associations are reversed. The other records include request identifiers,
issue times before the cut, and a text note unused by the program. The common checkpoint
snapshot records a current value of zero and two pending requests.

The future continuation \(C\) specifies sensor return values
\((u_1,u_2)\), a query \(q\in\{A,B\}\), and an arrival schedule.
Both returns arrive after the cut and by the fixed horizon \(T=3\), with
returns at \(T\) processed before the final readout. The return value for
each logical sensor is held fixed when comparing histories. Identifiers in
return records match the original requests in each history; different identifier
strings therefore do not change the logical continuation. The program joins
returns to requests, assigns each value to its recipient, and reads the queried
recipient's quantity \(v\). Let \(\mathbf1[\cdot]\) equal one when its
condition holds and zero otherwise. The vector observable is
\begin{equation}
O_T(H;C)=\bigl(v,\mathbf 1[v\ge\theta]\bigr),\qquad \theta=6.
\label{si:eq-inverse-readout}
\end{equation}
The threshold component is a function of the quantity, not an independent
measurement channel. The example uses finite comparisons; the binary component
does not require a derivative.
Times specify the constructed event schedule in arbitrary units, not measured
network latency.

The program retains and reads the full original records in every comparison.
Only the analyst's candidate representation changes: \(z_P\) is the snapshot,
\(z_D\) adds the two issue times, \(z_J\) adds the sensor--recipient
association, and \(z_F\) retains the full preparation records. Extraction
uses records available at the cut, before future values or the query are
supplied. Sufficiency here requires a readout function \(h_T\) of the
candidate state and continuation:
\begin{equation}
O_T(H;C)=h_T\bigl(z(H),C\bigr)
\quad\text{for every declared legal continuation }C.
\label{si:eq-inverse-continuation}
\end{equation}
Keeping the future input explicit distinguishes a sufficient checkpoint
description from autonomous dynamics. No candidate representation or prediction
is supplied to the program, so refinement does not change its answers.

For query \(A\) and sensor values \((2,7)\), the two orientations have the
same \(z_P\) but outputs \((2,0)\) and \((7,1)\). This rejects
\(z_P\). Equal issue times in the two histories also reject \(z_D\).
The comparison identifies a distinction that must be retained; selecting its
encoding remains a separate step.

\subsection{Exact sufficiency of the association state}
\label{sec:routing-sufficiency}
The sensor--recipient association is sufficient because the program sends
each sensor's return to its associated recipient. Histories with the same
association therefore give the same quantity and threshold indicator for
every allowed continuation, regardless of request identifiers, pre-cut issue
times, notes or arrival order. Both returns arrive by the observation horizon.

Conversely, exchanging the associations changes the quantity reported for
recipient \(A\) whenever \(u_1\ne u_2\). Such values are allowed, so a
sufficient representation must distinguish the two associations. The state
\(z_J\) does exactly this: for the specified program, common snapshot and
continuation class, it is the coarsest sufficient representation, with two
states up to relabeling. This argument covers finite real returns under the
specified schedule; the executions below check its implementation.

The execution check asks whether a fixed selection rule finds the association
state without retaining every history. Among the supplied candidates, it
excludes same-state response violations, selects the fewest distinct states
on the discovery histories and breaks ties by a fixed order. Selection uses
those outputs; predictions use the analytical readout rule.

Four discovery histories vary association orientation and unrelated notes.
Snapshot descriptions merge histories with different responses; full history
leaves no matched pair. The selection rule chooses the two-state association
description \(z_J\), then fixes it before new executions. SI
Section~\siref{si:inverse-mwe-reconstruction} gives the inputs and checks.

New executions test the selected state across changed notes, issue times and
future inputs. Sixteen histories are evaluated under twenty-four continuations
varying return values, arrival order and queried recipient, including equal
returns that do not distinguish associations. These give 384 base evaluations.
Each association state still merges eight distinct histories, so the test
compares different histories rather than assigning each input its own state
(Table~\ref{si:tab-inverse-mwe}).

\begin{table}[htbp]
\centering
\caption{\textbf{Exact comparisons after representation selection.}
The constructed data comprise sixteen histories evaluated under twenty-four
continuations, giving 384 base executions. State counts give the number of
distinct candidate states. Each comparison is an unordered pair of histories with
the same candidate state under one common continuation; a violation means
their quantity/threshold readout vectors differ. These are dependent
comparisons reusing executions, not independent tasks.}
\label{si:tab-inverse-mwe}
\begin{tabularx}{\linewidth}{@{}Yrrr@{}}
\toprule
\textbf{Candidate representation} & \textbf{States} & \textbf{Comparisons} & \textbf{Violations} \\
\midrule
Snapshot \(z_P\) & 1 & 2,880 & 1,024 \\
Snapshot and issue times \(z_D\) & 2 & 1,344 & 512 \\
Snapshot and sensor--recipient association \(z_J\) & 2 & 1,344 & 0 \\
Full preparation history \(z_F\) & 16 & 0 & --- \\
\bottomrule
\end{tabularx}

\smallskip
\raggedright\small
The dash means no matched-state comparison is available, not successful
validation. For \(z_J\), \(2\binom{8}{2}\times24=1,344\).
\end{table}

Predictions agree with all 384 base executions. Fixing the association to
\(S_1\!\to A\) instead makes 128 errors. Eight continuations have equal
sensor returns (128 executions): their agreement cannot support snapshot
sufficiency because the inputs do not distinguish the associations. The largest
matched-snapshot quantity difference is 12.

The controls test response to the queried sensor and invariance to the other.
Starting from returns \((3,11)\), with \(S_1\) arriving at time \(1\)
and \(S_2\) at time \(2\), each history and query receives a unit increment
to either sensor. The quantity increases by one in all 32 queried-sensor
controls and stays unchanged in all 32 unqueried-sensor controls. Predictions
agree in all 64. Together with four discovery and 384 base evaluations, these
are 452 deterministic program executions, not model calls or independent samples.

The analytical argument establishes sufficiency for the specified program
and continuations; new executions check its implementation after selection.
The construction establishes an information requirement, not an algorithmic
advantage over direct execution of the known program. Language-model adequacy and intervention value require separate measurements.

\subsection{Testing the association state in a language model}
\label{si:inverse-model}
Associations suffice for the program in Section~\ref{si:inverse-mwe}; whether
they preserve a language model's responses requires measurement. We replace
the program with Qwen2.5-14B-Instruct (MLX 4-bit), which reads complete
request--return records and reports the queried recipient's quantity.
Its checkpoint, instructions and sampling settings stay fixed. Candidate
states and reference answers never enter its input.

The restricted comparison tests reports for recipient \(A\), chosen after
development involving both recipients. It uses the checkpoint and candidates
\(z_P,z_D,z_J,z_F\) of Section~\ref{sec:worked-tool}. Returns carry request
identifiers, quantities and arrival times; requests precede the cut and both
returns arrive by \(T=3\). These are constructed record times, not measured
network delays.

Model inference samples an integer quantity in a single-field JSON object;
other completed responses enter \(\mathrm{INVALID}\). All incorrect and
malformed responses remain in the analysis. Joining returns to requests gives
an independent task reference for scoring correctness.

A preliminary screen finds no candidate that passes across both recipients.
Four histories and two continuations receive four calls per input condition.
The screen requires matched pairs with empirical total variation (TV) at most
\(0.15\) on each, then minimizes state count with fixed ties. It supplies no
equivalence confidence statement. Because no candidate passes, the planned
768 confirmation calls remain unrun.

The failure is concentrated in queries about recipient \(B\). All 16
responses querying \(A\) are correct; 5/16 querying \(B\) give the other
recipient's value, and an association-matched pair has empirical TV \(0.75\).
Identifiers and notes vary together, while query, values and arrival order
covary across continuations; the cause is not isolated. Shared issue times make
\(z_D\) identical to \(z_P\). SI Section~\siref{si:inverse-model-methods}
retains the inputs, all 32 development responses and 16 unscored sampling checks.

A separate follow-up tests recipient \(A\), a restriction chosen after
development. Selection reuses its 16 responses to choose \(z_J\); these are
not independent confirmation. The representation and requests are fixed before
new responses. Histories, values, arrival orders and controls retain the
unexecuted broader design; only the second query changes from \(B\) to
\(A\), with corresponding reference quantities
(SI Section~\siref{si:inverse-model-methods}).

The follow-up tests whether opposite associations give different answers at
the same snapshot, while matching associations give agreement despite other
request changes. The prompt instructs matching returns to original request
identifiers independently of arrival order.

The follow-up compares four new histories combining both association
orientations with two identifier/time/note preparations. Two continuation
combinations query \(A\), giving eight base conditions and four matched
association pairs. They vary sensor values and arrival order together, not as
independent factors. Four controls increment the queried or unqueried sensor
for one preparation in each orientation. The design fixes 64 calls per
condition (768 total), the margin and the stopping rule before responses, with
no replacement histories or additional draws. SI
Section~\siref{si:inverse-model-methods} specifies the inputs.

For matched histories, reference-answer frequencies bound the distance between
full output distributions. Each condition fixes a history and continuation.
Let \(\Pi_i\) be the parsed-output law in condition \(i\), including \(\mathrm{INVALID}\),
and \(a_i\) its independently
specified reference quantity. The count \(k_i\) of responses equal to \(a_i\),
among \(n=64\) draws, gives an exact one-sided binomial lower bound \(L_i\) on
\(\Pi_i(a_i)\)\cite{ClopperPearson1934}. Allocating error \(0.05/12\)
per condition gives simultaneous coverage at least 95\%, conditional on independent
draws from stationary output laws within each condition. Reused bounds require no
independence between pair comparisons. Requested seeds and identical answers
do not establish these sampling assumptions or their server implementation.

On this simultaneous event, overlap at a common reference answer gives
\begin{equation}
a_i=a_j\;\Longrightarrow\;
d_{\mathrm{TV}}(\Pi_i,\Pi_j)\le 1-\min(L_i,L_j).
\label{si:eq-model-tv-upper}
\end{equation}
If the reference quantities differ, \(\Pi_j(a_i)\le1-\Pi_j(a_j)\), hence
\begin{equation}
a_i\ne a_j\;\Longrightarrow\;
d_{\mathrm{TV}}(\Pi_i,\Pi_j)\ge\max\{0,L_i+L_j-1\}.
\label{si:eq-model-tv-lower}
\end{equation}
Responses unequal to the reference, including invalid outputs, remain in the denominator.
Comparisons using these bounds can remain unresolved when two distributions
concentrate on the same incorrect answer;
task accuracy is therefore not equivalent to representation adequacy.
At this fixed sample size, the bounds favor concentration at a common reference
answer; identical dispersed empirical distributions may remain unresolved.
Some non-reference responses are compatible with bounded agreement, so these
bounds do not require complete empirical concentration.

Bounded support requires every matched comparison to have an upper bound at
most \(0.15\). A lower bound above \(0.15\) rejects the candidate on that
comparison; otherwise the result is unresolved. For equal-reference pairs this
rule based on reference-answer probabilities can support or leave unresolved, but cannot reject. The fixed
snapshot comparison uses opposite sensor--recipient assignments with the
first note and issue-time preparation, under returns \((3,11)\) with \(S_2\) arriving first.
That comparison and both queried-sensor increments additionally require lower
bounds above \(0.50\). Both unqueried-sensor controls require upper bounds at
most \(0.15\). Full history supplies no matched pair and is uninformative.
The model retains access to information omitted by the snapshot, so the
interface does not impose agreement at matched candidate states.

The restricted comparison supports the association state: all 768 responses
are valid and equal their references, and all queried- and unqueried-sensor
controls pass. Opposite associations give quantities 3 and 11 under one
continuation, and 13 and 1 under the other, while association-matched histories
agree. Each condition has 64 reference responses, so
\(L_i=(0.05/12)^{1/64}\). Equations~\eqref{si:eq-model-tv-upper}
and~\eqref{si:eq-model-tv-lower} give conservative TV upper and lower bounds
\(0.0821\) and \(0.8358\), meeting the agreement and detection thresholds.
These are repeated measurements of four histories, not 768 independent tasks.

\begin{figure}[!htbp]
\centering
\includegraphics[width=\linewidth]{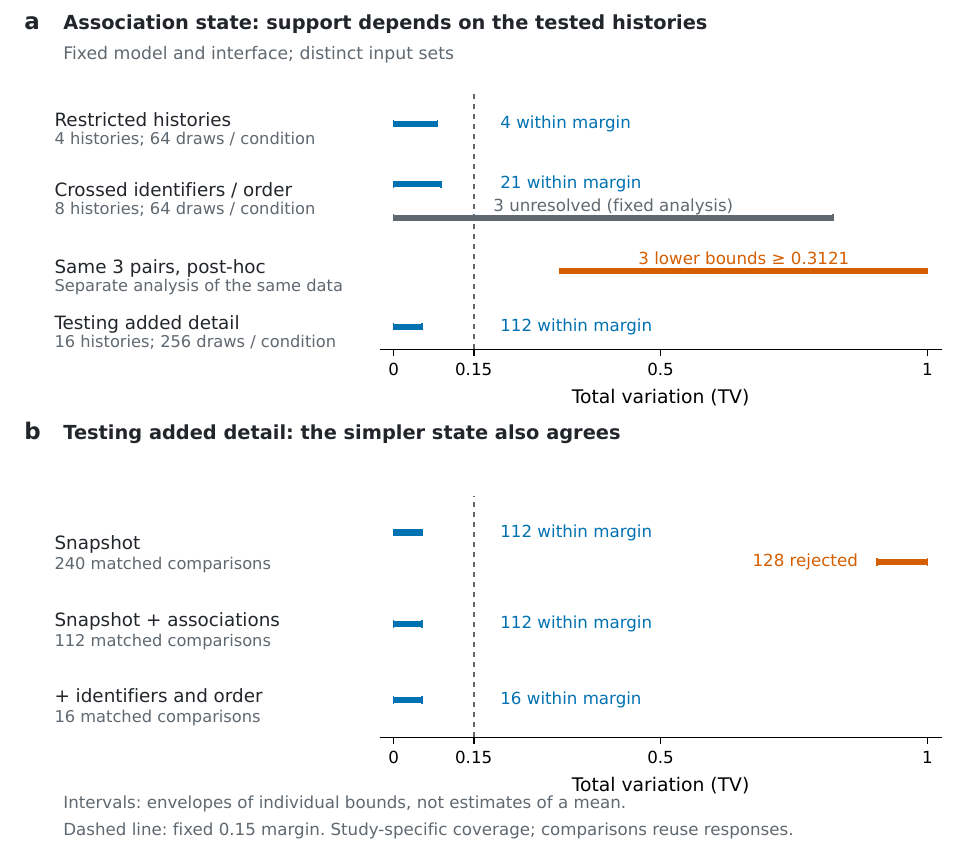}
\caption{\textbf{Bounded agreement and its limits for one fixed language model.}
Candidate descriptions are computed by the analyst; the model always reads
complete request and return records. The snapshot records the current value
and pending-request count; the association state also retains the
sensor--recipient mapping. \textbf{a}, Association-state comparisons
in three distinct input sets. The restricted comparison supports four matched
pairs. The crossed design has 21 within-margin pairs and three unresolved under
its fixed reference-probability bounds; a separate post-hoc categorical analysis gives those
three pairs a total-variation (TV) lower bound of \(0.3121\). The comparison testing
added identifiers and order supports all
112 pairs. Bars summarize individual bounds by classification; labels give comparison counts.
\textbf{b}, That comparison rejects the snapshot but supports both the association
state and its revision retaining identifiers and order. Displayed ranges span
the individual simultaneous bounds, not confidence intervals for an average.
The dashed line marks the fixed TV margin \(0.15\). Nominal 95\% coverage is
separate for each study and analysis, assuming independent, stationary replicates
within each study; conditions within a replicate may be dependent.
All studies query recipient \(A\), a restriction chosen after development.
Agreement in this comparison does not remove the exploratory evidence of
crossed-study separation or demonstrate an advantage for the revision.}
\label{fig:inverse-validation}
\end{figure}

Adequacy depends on which histories a candidate merges. Reusing these responses,
the snapshot has eight rejected and four within-margin pairs; adding issue
times leaves four rejected pairs. All four association-matched pairs lie within
the margin (Figure~\ref{fig:inverse-validation}a), whereas full history
provides no matched pair and remains uninformative.

This bounded support concerns four histories, two continuation combinations
and recipient \(A\). Identifier families covary with association orientation,
and sensors appear in a fixed request order. The comparison therefore cannot
distinguish the association partition from correlated identifier or position
descriptions; this does not establish which information the model uses.
Selection within a supplied library does not identify a unique hidden
variable. The next comparison independently varies identifiers and request
order to test extension beyond this restricted support.

\subsection{Testing identifier and request-order invariance}
\label{sec:crossed-routing}
If associations suffice, changing request identifiers or list order while
preserving those associations should leave the output distribution unchanged.
A fixed comparison tests this requirement on eight new histories crossing two
identifier schemes, two association orientations and two request orders. The identifier schemes are reused in both
orientations, and the analyst records the association as a sensor-sorted map;
changing request order therefore leaves this candidate state unchanged.
Neither operation changes the model instruction or its access to the
complete records.

Both future-input combinations query \(A\), with return references renamed
consistently to preserve the sensor-level continuation. The fixed design gives
64 calls in each of 16 base conditions and four controls, forming 24
association-matched pairs. It allocates error \(0.05/20\) per condition and
retains the earlier agreement, detection and control margins. SI
Section~\siref{si:inverse-model-methods} gives the input values, crossed factors
and sampling procedure.

The fixed bounds leave association-state agreement unresolved for some identifier
and request-order changes: 21 comparisons lie within the equivalence margin
and three remain unresolved. All 1,280 responses are valid. The unresolved comparisons share one
input: with the first identifier scheme, the original request order and
\(S_2\) assigned to \(A\), the first continuation gives the reference quantity
\(17\) in 21/64 calls and the other sensor's quantity \(5\) in 43/64 calls.
Changing only the identifier scheme or only the request order gives \(17\) in
64/64 calls. Each of these empirical distribution differences is \(43/64\).
The specified reference-probability bounds give an upper limit of \(0.8234\) and a
lower limit of zero for these equal-reference comparisons; they therefore
leave the comparisons unresolved, rather than formally rejecting the state.

The crossed design fails the combined support criterion. Seven of the eight
single-factor orientation comparisons resolve differences,
rejecting both the snapshot and snapshot-with-times candidates. Queried and
unqueried controls pass in one orientation and remain unresolved in the other,
so support does not extend to independent identifier and order changes under
this criterion. These new histories test extension beyond the restricted
comparison, rather than retesting its original inputs.
These behavioral observations identify neither internal computation nor
invariance under arbitrary renaming and permutations.

An exploratory analysis chosen after observing responses compares output
categories without reference answers across all 20 conditions, using simultaneous
component-probability bounds (SI Section~\siref{si:inverse-model-methods}).
Each of the three unresolved association pairs has a conservatively rounded
TV lower bound of \(0.3121\), above \(0.15\). An association-state-only
approximation assigns both histories the same output law; the triangle
inequality makes its error on at least one input at least half their separation.
This gives worst-case behavioral approximation error at least \(0.1560\)
on one input per pair, under this post-hoc family's nominal 95\% simultaneous
coverage and the stated sampling assumptions
(Section~\ref{sec:compression-error}).
The exploratory separation leaves the original classifications unchanged.
Both analyses reuse the same observations, providing neither independent
replication nor evidence of internal mechanism.

\subsection{Testing an extended representation on new histories}
\label{sec:routing-refinement}
The crossed comparison motivates testing whether retaining identifiers and
request order, in addition to associations, improves agreement with model
responses. A new experiment tests that revision against the
simpler association state, with all candidates, histories, comparisons and
analysis rules fixed before new responses. The model, instruction and input
schema remain unchanged; the recipient-\(A\) restriction is inherited from the
broader development comparison. The fixed refinement criterion requires both
agreement for the revised description and a violation of the simpler description
on the new inputs. The combined criterion additionally requires the specified
controls.

Sixteen new histories cross identifier scheme, association orientation,
request order and a joint timing/note change. The revised state retains the
snapshot and ordered request-ID/sensor/recipient triples. It merges the two
timing/note histories at each of eight states, giving 16 comparisons under two
new continuations. Each revised-state pair has identical raw future returns;
cross-identifier comparisons for the simpler state consistently rename return
references. SI Section~\siref{si:refinement-methods} specifies all inputs.

The design specifies 256 calls in each of 32 base conditions and four controls.
Simultaneous bounds compare the complete integer/invalid-output laws through
marginal event probabilities and paired-output disagreements; reference answers
only score correctness. The pairing requests one common seed across conditions
in each replicate. Its validity assumes independent stationary replicate vectors
with the stated marginal distributions, without requiring independence between
conditions. SI Section~\siref{si:refinement-methods} gives the error allocation,
controls and the treatment of other integers.

Both the association state and its revision pass on these new inputs; the
fixed refinement criterion therefore does not establish a benefit from retaining
identifiers and order. All 9,216 calls complete. The snapshot has 128 resolved violations among 240
comparisons; 64 occur with identical raw future returns. In contrast,
all 112 association-state comparisons and all 16 revised-state comparisons
have TV upper bounds at most \(0.0531\), below the fixed \(0.15\) margin
(Figure~\ref{fig:inverse-validation}b). Both declared queried-sensor controls
exceed the \(0.50\) detection threshold, and both unqueried-sensor controls
lie within \(0.15\); these checks cover their specified layouts only.

Task accuracy is assessed separately: 8,190/8,192 base responses and
1,024/1,024 control responses are correct, with no invalid outputs.
Repeated calls are not additional tasks.

These comparisons support omission of the specified timing/note differences.
They do not resample the earlier input with 21/64 reference answers: its
exploratory separation and the three originally unresolved comparisons remain
unchanged. Model support concerns the tested histories; exact sufficiency
concerns the specified request--return program.

\FloatBarrier
\section{Inverse tests under tool feedback}
\label{sec:closed-loop}

\begin{figure}[tbp]
\centering
\includegraphics[width=\linewidth]{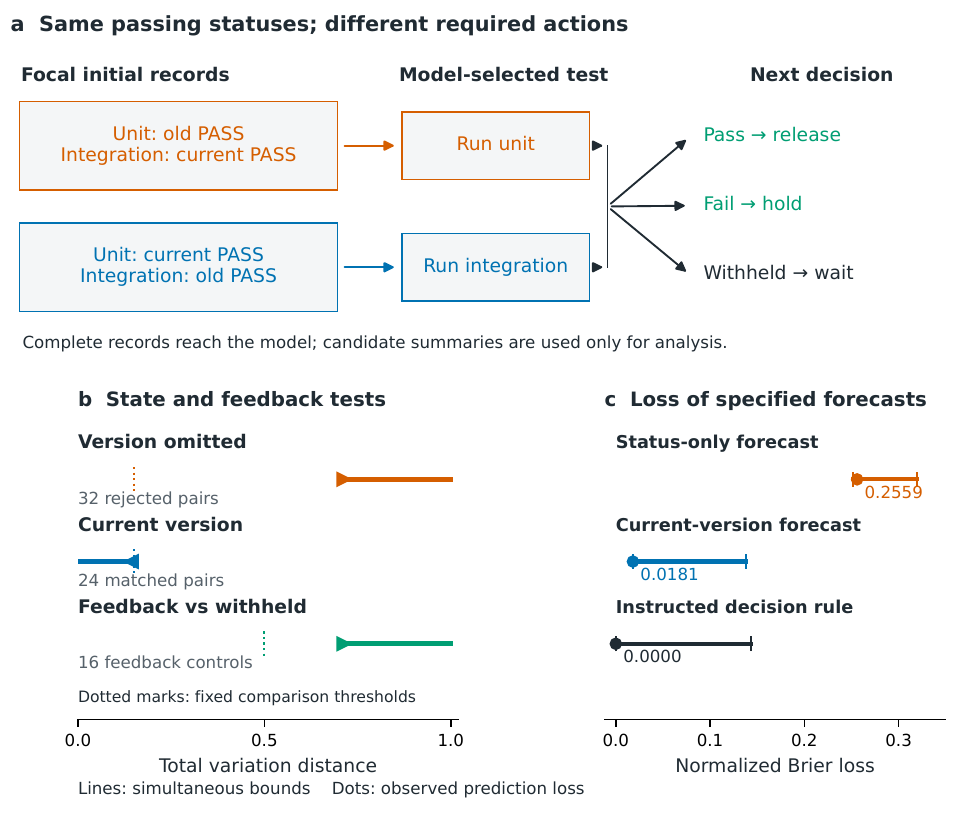}
\caption{\textbf{Prescribed version information tested through adaptive actions and tool feedback.}
The model is instructed to assess current-version statuses and then follow a
complete decision rule; record joining is performed before the model call.
\textbf{a}, The two focal passing results concern different software
versions. The schematic omits a redundant predecessor result present in all
confirmation inputs. The model selects the missing current-version test; its return changes
the next decision. Withholding that return preserves the executed request.
\textbf{b}, Intervals show conservative total variation (TV) bounds: the 32 violations for a
version-omitting state (the second such state gives identical results), all 24
comparisons for the current-version state, and all 16 feedback controls.
Each row shows their common bound; the fixed thresholds are 0.15 for state
agreement and 0.50 for feedback controls.
\textbf{c}, Dots show observed normalized Brier loss; lines bound expected loss,
averaged over the 16 feedback conditions. Forecasts were fitted to development responses and fixed
before new observations; the third predictor is the instructed decision rule.
All predictions condition on the imposed pass/fail outcome and observation
branch; they predict the model's actions.
Panels b and c share one nominal 95\% confidence event, conditional on
independent stationary repetitions within each condition--branch distribution.
On that event, the expected loss difference between fitted forecasts is at least
\(0.1180\), conditional on the fits.
The 1,024 episodes and 3,072 calls are repeated measurements of eight
constructed histories.}
\label{fig:closed-loop}
\end{figure}

The request--return tests fix later sensor inputs. We now test state adequacy
when an action determines the next observation: the model selects a software
test, receives its result and acts again. Release requires both unit and
integration tests to pass for the current version; an earlier version's pass
leaves a current test outstanding. Fresh calls to Qwen2.5-14B-Instruct
(MLX 4-bit), with simulated tools and releases, test this prescribed distinction
over two decisions. The instructions give the complete rule and require
current-version status assessment before action. This is a controlled test of
specified information requirements, rather than discovery of an unknown
requirement or a software benchmark (Figure~\ref{fig:closed-loop}).

\subsection{Versioned records and action-dependent observations}
Two histories contain the same passing statuses but differ in which test
belongs to the old version: one requires a unit test, the other an integration
test. Before inference, the experiment program joins returns to requests by
identifier and supplies completed records and pending requests. The model
reports current-version statuses and chooses a test, wait, release or hold;
the output format permits every status--action combination. The instructions
also declare identifier spelling and record order irrelevant to version.
Thus the measured invariances, like the version distinction, are prescribed.
SI Section~\siref{si:closed-loop} gives the exact instructions.

The chosen test returns an externally imposed pass or fail, hidden from the
first call. Two fresh second calls share that first action: one receives the
request and return, the other only the request. Policy, prompt and model stay
fixed. The expected decisions are release after a pass, hold after a fail,
and wait when the return is withheld. The readout is the action sequence, with invalid outputs retained; branches
sharing the first action are dependent observations.

The candidate states test whether version information is needed beyond passing
statuses. Three analyst-only initial summaries retain the displayed statuses
and pending count (status-only), the latest status per test regardless of
version, or each current-version result including absence. Only the last
distinguishes the missing tests; it omits identifiers and order.
The model receives complete records throughout. After interface development,
these task-derived candidates, the model, prompt, comparison rules and forecasts
were fixed before confirmation. Eight histories cross the missing test with
two identifier schemes and two request orders; version names stay fixed.
Each imposed pass/fail condition has 64 episodes per history. SI
Section~\siref{si:closed-loop} specifies construction and sampling.

\subsection{Testing version information through action sequences}
Reproducing the observed action sequences requires distinguishing the missing
current-version test; the two passing statuses alone are insufficient. Every call produced
a valid action: request the missing test, then release after a pass, hold after
a fail and wait when its return is withheld.

To test whether each summary preserves action-sequence distributions, we use
the total-variation rules in Section~\ref{iclr:protocol}: agreement tolerance
0.15, control threshold 0.50 and 16 possible parsed sequences, including
invalid outcomes. One nominal 95\% confidence event covers
all 32 distributions, crossing eight histories with pass/fail test outcomes
and delivered/withheld returns. Repeats are assumed independent and
stationary within each distribution; branches may be dependent. At this sample size,
bounded agreement requires both samples to concentrate on the same sequence;
this does not establish a deterministic law.

For either version-omitting summary, the same 32 of 56 matched pairs are rejected:
one set of missing-test contrasts, not independent evidence for two omissions.
The current-version state has bounded agreement on all 24 matches, with maximum
TV upper bound \(0.1437\). The rejected pairs and all controls---eight
pass-versus-fail and 16 return-withholding comparisons---have TV lower bound
at least \(0.7126\) (Figure~\ref{fig:closed-loop}b). These results support the
predefined refinement and resolve a tool-return effect on the next action.
The same separation implies worst-case behavioral approximation error
\(\ge 0.3563\) for either rejected summary on at least one merged history,
on the same confidence event (Eq.~\eqref{eq:compression-diameter}).

Forecasts quantify the predictive cost of omitting version information.
Both are fitted to the development counts in SI
Table~\siref{tab:closed-loop-development-counts}, with identical estimation
and smoothing, then fixed before confirmation. They condition on imposed
pass/fail outcome and observation branch, without future sampled actions.
Across 16 new feedback conditions, mean normalized Brier loss~\citep{brier1950verification}
(squared probability error scaled to zero--one) is \(0.2559\) for status-only
and \(0.0181\) for current-version prediction (Figure~\ref{fig:closed-loop}c).
The instructed rule has zero observed loss. These scores reuse the representation
observations; they establish neither optimal compression cost nor an advantage
over that rule.

Readout choice changes the information requirement: both histories may end
in release after different tests. An analysis prompted by interim observations
retains only the final action, projecting the original confidence set. All
56 status-only matches then have TV upper bound \(0.1437\), while the
32 full-sequence rejections remain unchanged.

The supported summary concerns the initial history and omits pending-request
information needed later; the comparison tests initial-state adequacy rather
than autonomous dynamics.
First actions were constant within each condition, so feedback was tested after
that observed action. The scope is one model, a developed interface and two
decisions. An earlier action-only interface returned RELEASE on all 32 initial
calls. The successful interface jointly changes instructions, joins records
before inference and requests status assessment; their separate effects were
not tested. SI Section~\siref{si:closed-loop} reports the interface failures
and other feedback pilots.

\section{Inverse tests with scientific text}
\label{sec:scientific-evidence}
The instructed tasks specify which history distinctions should matter. We now
ask which content and ordering distinctions affect a fixed model's judgments
of scientific text.
A workflow retrieves abstracts, selects excerpts and passes an ordered packet
to a fixed reader. Comparing summaries of that packet tests whether document
identity, excerpt content and document order account for its judgments,
separately from scientific correctness.

\subsection{Text inputs and candidate representations}
We vary excerpt content and document order separately to test what the
representation must preserve. SciFact supplies claims, abstracts and annotated
evidence~\citep{wadden2020fact}; eight training-split claims follow an existing
development list without response screening or substitutions
(SI Section~\siref{si:development-decisions}). For each claim, BM25 keyword
ranking retrieves six abstracts. Deterministic selection takes the first three
or last three sentences of each, producing 18-sentence packets in retrieved
or reversed document order. Sentence order within each document stays fixed.
The four conditions therefore vary excerpt content, document order or both.
This uses real abstract text and new calls, but is a controlled comparison,
not a held-out benchmark. Selection never uses annotations or answer labels.
Annotations only track whether complete annotated rationales remain; shortened packets receive no
correctness score because removal need not preserve the original label.

The coarse state retains the claim, document identities and excerpt counts,
merging all four conditions. The refined state adds exact excerpts and
provenance but omits document order, giving two matched pairs per claim.
Both are analyst descriptions; the reader always receives the ordered packet.
Document identity cannot recover which sentences were passed, and matching
unordered excerpts does not make the reader's inputs identical.

Qwen2.5-14B-Instruct (MLX 4-bit) receives the claim, titles and excerpts with
document and sentence identifiers. Its task and instructions remain fixed.
Outputs are support, refutation, insufficient evidence or INVALID for any
response outside the one-letter format. Sixty-four fresh calls per condition
estimate 32 four-category distributions: eight tasks and 2,048 repeated calls.
The source is excerpt selection or ordering; the readout is the judgment
distribution after that operation. These finite contrasts require neither a
derivative nor a graph or feedback model.

The total-variation comparisons of Section~\ref{iclr:protocol} use tolerance
\(0.15\) and simultaneous 95\% bounds across all four categories and 32
conditions, assuming fixed sample sizes and independent stationary repeats
within each. Informative support additionally requires an INVALID-probability
upper bound at most \(0.15\) and a same-claim excerpt-policy contrast with TV
lower bound above \(0.5\). This positive control tests sensitivity to changed
content, not to order. All comparisons are reported, including those without
these controls. With 64 calls, the bound resolves agreement only for samples
concentrated on the same category; dispersed samples require greater precision.
SI Section~\siref{si:scientific-evidence-methods} gives the intervals and
complete observations.

\begin{figure}[tbp]
\centering
\includegraphics[width=\linewidth]{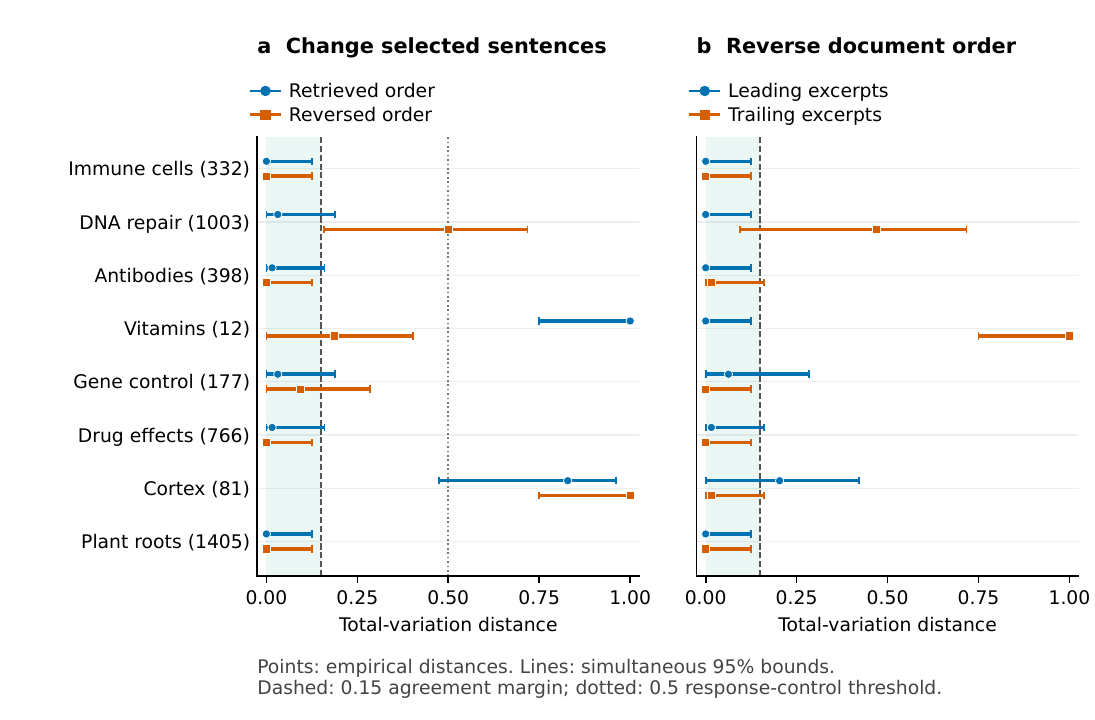}
\caption{\textbf{Excerpt content and document order make different demands on a state.}
Rows identify the eight SciFact claims by topic and original dataset identifier.
\textbf{a}, Replace the first three sentences of each abstract with its last
three, keeping document order fixed. The candidate description retains the
claim, document identities and excerpt counts. \textbf{b}, Reverse document
order while keeping all selected sentences fixed; the candidate description
also retains the exact excerpts and their provenance, but omits document order.
Leading and trailing excerpts are shown separately. Points are empirical
total-variation distances; lines give
simultaneous 95\% bounds across the 32 conditions, each measured with 64 calls,
assuming independent, stationary repeats within each condition. The dashed line marks the \(0.15\)
agreement margin; an interval wholly to its left gives bounded agreement,
wholly to its right rejects that margin, and an overlapping interval is
unresolved. The dotted line in \textbf{a} marks the positive-control threshold:
an excerpt-policy contrast must have a distance lower bound above \(0.5\).
Only the leading-excerpt comparison for the vitamin claim (12)
combines bounded agreement with a detectable positive control for the same claim.
The distance measures changes in model judgments, not task accuracy.}
\label{fig:scientific-evidence}
\end{figure}

\subsection{Response differences from content and order}
The tested responses require distinctions in both excerpt content and document
order (Figure~\ref{fig:scientific-evidence}). Neither candidate preserves all
of these distinctions. The document-and-count state is rejected
for three of eight claims, including claim 1003 whose lower bound of about
\(0.1578\) lies close to the \(0.15\) tolerance. Among 16 order comparisons
at fixed excerpts, nine show bounded agreement, one rejects the tolerance,
and six remain unresolved. Only one of the nine also passes the positive
control; the other eight lack it. All 2,048 responses satisfy the format,
which is separate from representation adequacy.

The SciFact claim about folic acid, vitamin B12 and kidney-disease progression
illustrates both requirements: changing excerpt content changes the judgments,
and reordering the same excerpts changes them again. For the leading excerpts, all 64 replies in each order
report insufficient evidence (TV upper bound \(\le 0.1250\)). Trailing excerpts
in retrieved order instead produce 64 support judgments, resolving the content
control (lower bound \(0.7501>0.5\)). Reversing those same trailing excerpts
produces 12 refutation and 52 insufficient-evidence judgments, rejecting the
unordered state (lower bound \(0.7501>0.15\)). Thus the leading-excerpt pair
has bounded local support with a detectable control, while another pair for
the same claim requires an ordering distinction.

The immune-cell depletion claim (332) distinguishes agreement from correctness:
all 256 replies report support although every packet retains complete
source-annotated contradicting evidence. The matched pairs have bounded
agreement but no detectable positive control, so do not meet the informative
support criterion. The annotation discrepancy is not an independently
adjudicated error rate for these shortened packets.

A revision must distinguish the rejected content and order pairs. Neither
tested description is adequate across the eight claims. A proposed revision
requires new informative comparisons; assigning each input a unique state
would remove the matched histories needed to test it.

\FloatBarrier
\section{Information required to preserve responses}
\label{sec:information-requirements}

The response comparisons raise two general questions: what must a representation
preserve across several systems, and what approximation error is unavoidable
when it merges distinguishable histories? The results below constrain retained
information for deterministic observables and stochastic output laws, without
selecting a unique encoding.

\subsection{Sufficiency for one or several systems}
\label{si:consumer-family}
A representation adequate for several systems must preserve every distinction
required by any member. The systems may still produce different output laws.
Let \(\mathscr H\) be a nonempty finite set of preparation histories with
the discrete sigma-algebra, and \(\mathscr M\) a nonempty declared set of
systems whose dynamics and observation rules remain fixed. Let \(\mathscr C\)
be a nonempty measurable set of continuation conditions admissible for every
system--history combination. These specify subsequent input sequences, or
external tool-response and observation rules for an adaptive policy included
in the fixed system \(m\) (Section~\ref{sec:adaptive-response}).
Each system produces the chosen observable from a history and such a
continuation; endogenous actions are integrated into its output law. Fix each system's interface, readout, window and
future noise law. For \(m\in\mathscr M\), denote its conditional output
distribution by \(\mathsf P_m(\cdot\mid h,v)\), where
\(h\in\mathscr H\) and \(v\in\mathscr C\).
For each system, assume these distributions are probability kernels
measurable in \((h,v)\), on a common declared measurable output space.
Deterministic outputs are included as point masses.

Two histories are response-equivalent for this family when
\begin{equation}
h\sim_{\mathscr M}h'
\quad\Longleftrightarrow\quad
\mathsf P_m(\cdot\mid h,v)=\mathsf P_m(\cdot\mid h',v)
\quad\text{for every }m\in\mathscr M,\ v\in\mathscr C.
\label{eq:public-consumer-equivalence}
\end{equation}

A representation \(\psi:\mathscr H\to Z\), with its image equipped with the
discrete sigma-algebra, is sufficient for every declared system exactly when
\(\psi(h)=\psi(h')\) implies \(h\sim_{\mathscr M}h'\). Necessity follows
because a readout depending only on \((\psi(h),v)\) gives equal laws at equal
states. For sufficiency, choose any history assigned to a state and use its
output law: the implication makes the choice immaterial. Finite
\(\mathscr H\), the discrete state image and the assumed kernel measurability
in \(v\) ensure that these state-based laws are measurable.

Equality of output laws partitions histories into response-equivalent classes.
Every sufficient representation must distinguish these classes, which therefore
give the coarsest sufficient partition, up to relabeling. Adding systems or
allowed continuations adds equality requirements and can only refine it.

The criterion makes the information
preserved by a shared representation an explicit experimental target and
specifies the comparisons needed to test it. Its mathematical basis is
the established predictive equivalence principle
\citep{shalizi2001computational,littman2001psr}, applied here to declared
response experiments.

If all deterministic systems are allowed and two output values can be
distinguished by a measurable event, adequacy requires an injective history
encoding: a system can use any distinction that a compressed description
discards. A useful compressed representation therefore has a specified
readout and set of systems.

\subsection{Finite evidence and transfer between systems}
Finite comparisons can reject a candidate or bound agreement on tested
histories and continuations. An exact same-state violation identifies histories
that a revised representation must separate; bounds within the declared
tolerance establish agreement only on those comparisons.
Informative local support additionally requires the comparison coverage and
controls in Section~\ref{iclr:protocol}. Tolerance-based agreement need not
be transitive and does not define the exact equivalence classes above.
A full-history description can give every preparation a different state,
leaving no informative matched-history comparison.

Agreement for one system can support agreement for another if their output
laws are uniformly close. This requires an independently justified bound on
the change of system, in addition to the measured within-state agreement.
Using total variation distance \(d_{\rm TV}\) as in
Section~\ref{iclr:protocol}, define the worst within-state
discrepancy over the declared histories and continuations by
\begin{equation}
d_\psi(m)=
\sup_{\substack{h,h'\in\mathscr H,\ v\in\mathscr C\\
                 \psi(h)=\psi(h')}}
d_{\rm TV}\!\left(\mathsf P_m(\cdot\mid h,v),
                  \mathsf P_m(\cdot\mid h',v)\right).
\label{eq:public-state-discrepancy}
\end{equation}
If \(d_\psi(m_0)\le\varepsilon\) and an independently justified uniform
bound on the change in output law between systems satisfies
\[
\sup_{h,v}d_{\rm TV}\!\left(\mathsf P_m(\cdot\mid h,v),
                           \mathsf P_{m_0}(\cdot\mid h,v)\right)\le\rho,
\]
then the triangle inequality gives
\begin{equation}
d_\psi(m)\le\min\{1,\varepsilon+2\rho\}.
\label{eq:public-representation-transfer}
\end{equation}
Each history in a matched comparison contributes one term accounting for
the change of system.
The independently controlled discrepancy \(\rho\) is an additional
assumption; within-state agreement in one model does not estimate it.
The bound is sufficient, rather than necessary: systems with different
output distributions can still preserve the same history distinctions.

AI memory summaries, checkpoints and records passed between components thus
have two separate experimental targets. A within-model test asks whether the
summary preserves the required distinctions. A cross-model test fixes that
summary and evaluates each declared model separately. Section~\ref{si:inverse-model}
tests one model on a finite set of histories; the family criterion specifies
the comparisons needed for several models.

\subsection{A lower bound on error from merging histories}
\label{sec:compression-error}
If a state merges histories with separated output laws, every state-only
approximation incurs error on at least one history. Fix a system \(m\), continuation \(v\) and state value \(z\).
Let \(F_z=\{h\in\mathscr H:\psi(h)=z\}\) be its nonempty set of matched
histories, and abbreviate the output law as \(P_h=\mathsf P_m(\cdot\mid h,v)\).
Any approximation using only \((z,v)\) must assign the same probability law
\(R\) to every history in \(F_z\). Define its optimal worst-case error and
the diameter of that set of output laws by
\begin{align}
 r_\psi(z,v)&=\inf_R\sup_{h\in F_z}d_{\rm TV}(P_h,R),\\
 D_\psi(z,v)&=\sup_{h,h'\in F_z}d_{\rm TV}(P_h,P_{h'}).
\end{align}
The infimum is over probability laws on the declared output space.

For any common approximating law \(R\), the triangle inequality gives
\[
d_{\rm TV}(P_h,P_{h'})
\le d_{\rm TV}(P_h,R)+d_{\rm TV}(P_{h'},R)
\le 2\sup_{g\in F_z}d_{\rm TV}(P_g,R).
\]
Taking the supremum over pairs and then the infimum over \(R\) yields
\begin{equation}
r_\psi(z,v)\ge\tfrac12D_\psi(z,v).
\label{eq:compression-diameter}
\end{equation}
Thus simultaneous lower bounds \(\ell_{hh'}\) for measured pairs in this
state class and continuation give
\(r_\psi(z,v)\ge\tfrac12\max_{h,h'}\ell_{hh'}\), with the same coverage.
This standard metric bound quantifies the cost of merging distinguishable
histories. For two output laws, their equal mixture attains half the diameter;
for larger sets the optimal error can be higher. Three point masses on distinct
categories, for example, have diameter one and optimal worst-case error \(2/3\).

The bound turns the experimental rejections into quantitative limits on any
predictor using the rejected summary. For the SciFact order contrast
(Section~\ref{sec:scientific-evidence}),
TV separation \(\ge0.7501\) implies approximation error \(\ge0.3750\)
for an unordered exact-excerpt state. For the request--return snapshot comparison
(Section~\ref{si:inverse-model}), separation \(\ge0.8358\) gives error
\(\ge0.4179\). For the version-omitting initial states in the closed-loop
workflow (Section~\ref{sec:closed-loop}), separation \(\ge0.7126\) gives
error \(\ge0.3563\). Each downward-rounded bound applies to at least one merged
history under its fixed system and continuation, inheriting that study's
simultaneous 95\% coverage and sampling assumptions, without joint coverage
across studies. These behavioral approximation limits are not independent
evidence or task error rates. A task-oriented workflow may change the program or model or standardize
presentation, then test the new target.

\section{Relation to response theory and workflow analysis}
\label{si:related-work}

ANM connects response prediction with tests of representation sufficiency.
Response theory supplies dynamical predictions; state and realization theories
specify which distinctions a representation must preserve. Workflow analysis
adds access to records and interventions. The comparison below concerns what
these methods measure and what they infer from those measurements.

Kubo's response theory relates susceptibilities to dynamical correlations
\citep{kubo1957,kubo1966}. Ruelle's review examines the response of
nonequilibrium steady-state averages to perturbations of differentiable
dynamics, with linear-response results under hyperbolicity and regularity
assumptions \citep{ruelle2009}.
The finite-window tangent response used here conditions on a prepared
trajectory and factors perturbation injection, propagation and readout.
It requires neither an equilibrium correlation representation nor a
steady-state limit.
Mori--Zwanzig projection derives reduced equations with memory from specified
microscopic dynamics \citep{mori1965,zwanzig1961}. ANM's inverse experiments
test whether a specified projection retains distinctions needed for the
measured response; they do not infer a memory kernel or establish a closed
evolution law for that projection.

State-space methods clarify what a measured response identifies. Functional
observability reconstructs chosen state functions from known inputs, dynamics
and outputs \citep{montanari2022functional,montanari2022nonlinear}.
ANM applies the standard linear factorization criterion to separately measured
state and readout responses: a source direction with zero state response and
nonzero readout response rejects the candidate. Full column rank leaves no such
direction (Section~\ref{iclr:protocol}; SI
Section~\siref{si:falsifiable-retained-state}). Data informativity asks which
properties or controllers are shared by systems consistent with observations,
without necessarily identifying one system \citep{vanwaarde2020informativity}.
Realization theory, including the Ho--Kalman construction
\citep{ho1966realization}, instead constructs dynamical states from input--output
behavior; a finite-window response partition need not evolve recursively.

Gertler's analytical-redundancy framework constructs residuals that measure
inconsistency between observed inputs and outputs and a system model, using
parity equations, diagnostic observers or Kalman filters for fault detection
and isolation \citep{gertler1991analytical}. ANM's representation tests ask
whether a proposed state description can account for separately measured
readout responses. A resolved violation constrains that description; it does
not by itself locate a faulty component or identify a causal mechanism.

State-abstraction theory classifies which model, value or policy information
an aggregation preserves \citep{li2006stateabstraction}. Bisimulation metrics
relate state similarity to optimal values \citep{ferns2004bisimulation}, while
approximate abstractions admit bounds on decision loss
\citep{abel2016approximate}. The bound in
Section~\ref{sec:compression-error} instead concerns approximation of a fixed
system's conditional output law. It follows directly from the triangle
inequality; value or policy guarantees require their own dynamical and reward
assumptions.
Approximate information states retain reward and predictive information
sufficient for approximate dynamic programming, with bounds on policy loss
\citep{subramanian2022information}. By contrast, agreement for one preparation
and readout does not by itself establish a recursive state update or a
planning guarantee.

For collective dynamics, El et al.~\citep{el2026physicsagents} fit an
Ising-inspired logistic law to opinions on signed networks with refreshed
inboxes. The model predicts trajectories and regimes, tests unseen network
families and includes a continuous-time mean-field extension. ANM instead
specifies measurements of source response and tests their factorization
through a candidate state.

The inverse question also connects to theories that define states through future
behavior. In automata theory, Nerode's construction groups input histories
that cannot be distinguished by subsequent inputs \citep{nerode1958automaton}.
Testing finitely many continuations supplies constraints on such a partition,
rather than a minimal-state reconstruction for arbitrary future inputs.
Stochastic bisimulation preserves rewards and block-transition probabilities
for every action \citep{givan2003}. Predictive-state representations recursively
summarize action--observation histories by future-test predictions
\citep{littman2001psr}; causal states group histories by conditional future laws
\citep{shalizi2001computational}. Causal-State Splitting Reconstruction
(CSSR) tests conditional next-symbol distributions and splits states to infer
recursive predictors from discrete sequences \citep{shalizi2004cssr}.
Input--output computational mechanics groups joint input--output histories
by their future output laws conditional on future inputs, yielding the
\(\epsilon\)-transducer \citep{barnett2015transducer}. Conditional prediction
and state splitting are thus established antecedents of ANM's representation
tests. ANM tests a proposed representation under specified perturbations,
readouts and finite observation windows; inference of a minimal recursive
predictor is a separate objective. For the constructed program, Section~\ref{sec:routing-sufficiency} derives
the coarsest sufficient partition for a declared continuation class; the other
tests do not establish predictive minimality.

Causal abstraction additionally requires agreement of interventional
solutions or laws under maps between states and permitted interventions
\citep{rubenstein2017causalconsistency,geiger2025abstraction}, a stronger
condition than local endpoint factorization. Structural causal models
likewise distinguish observational and intermediate-intervention agreement
\citep{pearl2009causality}.

In AI workflows, access to records determines which of these tests is possible.
Tool-using and conversational architectures expose intermediate records
\citep{yao2022react,schick2023toolformer,wu2023autogen}; GAIA supplies endpoint
tasks \citep{mialon2023gaia}. Field memory and cognitive substrates have been proposed for
retention and coordination \citep{mitra2026fieldmemory,coeva2026substrate},
while traces support error analysis
\citep{barrak2025traceability,wang2026agenttrace}.

Prompt compression already includes distributional fidelity objectives:
Wingate et al.~\citep{wingate2022compression} train soft prompts to approximate
the original prompt's conditional output distribution using KL divergence.
ANM's matched-history test evaluates a proposed state description without
substituting it into the model's input or training a compressed prompt.
Prompt compression such as LLMLingua measures token reduction against
downstream performance \citep{jiang2023llmlingua}. LoCoMo benchmarks long-term conversational memory
\citep{maharana2024locomo}. ANM's matched-history
comparisons test fidelity to a specified system's response distribution,
which is distinct from task performance.

CheckList applies behavioral and metamorphic testing through minimum-function,
invariance and directional tests \citep{ribeiro2020checklist}. Thus perturbing
an input and checking invariant behavior is established practice. The
representation tests here group comparisons by a specified candidate state,
use simultaneous uncertainty bounds, and translate resolved separations into
constraints on any readout based only on that state.
GRADE distinguishes execution order from observed, declared or inferred
dependencies and tests prediction beyond run size. Its within-corpus
degree-matched analysis establishes no edge-attachment advantage, with results
dependent on probe and grouping
\citep{zhao2026grade}. AgentGraph links relations to trace spans and combines
temporal replay, reconstructed-prompt perturbations and component attribution
\citep{wu2026agentgraph}.

Workflow diagnosis distinguishes attribution from the effect of a repair.
Who\&When localizes agents and steps from annotated logs
\citep{zhang2025failureattribution}; AIG combines step-grounded inheritance
graphs with original history \citep{bakish2026aig}. Both score attribution
against annotations. EDGE combines observational dependencies with edges
validated by repairing source spans and continuing with fixed tool outputs,
conditional on successful source repair \citep{hou2026edge}. DoVer restores
checkpoints, edits messages/plans while preserving preceding history, and
scores resumed runs by success, milestones and intervention fulfillment, permitting
multiple useful repairs \citep{ma2025dover}.

Changing a retained record addresses a different question from comparing two
unchanged histories assigned the same candidate state.
Sun et al.~\citep{sun2026constraintweakening} restrict execution to transformed
records passed between components under a fixed constraint, separately testing preservation, forbidden
actions, field restoration and verification at a fixed artifact.
Xiao et al.~\citep{xiao2026errorsmemories} cross clean/error inputs to memory
updating and question generation in four counterfactual trajectories,
measuring pathway effects and interactions; likelihood probes add
probability-level access, and subsequent histories evolve separately. These
interventions separate state and endpoint effects without establishing
arbitrary compressed-state sufficiency.

Activation patching replaces an internal activation, often through an
interchange intervention using a donor value,
\(\operatorname{do}(z=z_{\rm donor})\) \citep{zhang2024patching}; a clamp
\(\operatorname{do}(z=v)\) allows any admissible \(v\). Attribution patching
approximates replacement effects using gradients \citep{syed2024attribution}.
Mediation analysis separates direct and indirect effects through input and
mediator interventions \citep{vig2020mediation}; a patch effect alone does not
supply that decomposition. Activation interventions can propagate contextual
information across positions and layers; ANM's workflow experiments use
accessible communication records. Differences concern access, representation,
intervention and response resolution.

Input-order studies provide a closer comparison for the scientific-evidence
experiment.
SciFact separates evidence selection from claim classification
\citep{wadden2020fact}. Input-position effects \citep{liu2024lostmiddle} and
few-shot example-order effects \citep{lu2022promptorder} concern distinct
manipulations. Stable-RAG directly studies retrieved-document permutations,
using hidden-state clustering and preference training to improve answer
accuracy and consistency \citep{zhang2026stablerag}.
ANM organizes these established criteria around a common finite-window response
question. Candidate states group histories while the downstream component's
inputs stay unchanged; resolved differences constrain any readout based on those
states. The experiments test this information requirement under fixed inputs,
adaptive feedback and scientific-text judgments. Common empirical laws and
diagnostic or repair benefits require separate matched comparisons.

\section{Discussion}
\label{iclr:discussion}
A representation is adequate relative to a response question: which observable,
under which future inputs or interaction rules? ANM connects this question to
physical response prediction through common inference rules and
realization-specific dynamics. Paired measurements distinguish propagation from
readout sensitivity; inverse comparisons test whether the retained information
is sufficient, by resolving omitted distinctions or bounding agreement on tested
histories. Physical calibration tests predictions from known delayed dynamics.
The AI studies test information requirements for measured behavior, including
adaptive actions that select later observations.

Fidelity to a fixed downstream component includes its errors and presentation sensitivity;
it is distinct from the information needed for correct task performance.
In SciFact, preserving the measured response can require distinguishing document orderings,
whereas a task-oriented design may standardize presentation or change the
reader. Agreement can also coexist with judgments contrary to source
annotations. These comparisons characterize observable behavior. Inference about internal
representations or reasoning requires additional evidence; a changed target
requires a new comparison.

Prompt compression selects information for later inference~\citep{jiang2023llmlingua};
long-conversation benchmarks evaluate whether a model retains information needed
for later tasks~\citep{maharana2024locomo}. Checkpointing, caching and passing
records between components likewise involve choosing which information remains
available.
Matched-history tests assess a proposed summary while the model still receives
complete inputs. They identify distinctions the summary must preserve; new
comparisons then test a revision. Evaluating actual input compression additionally
requires running the model on the compressed inputs. A cache reproducing the reader's output distribution, for
example, cannot merge the rejected SciFact orderings at the tested tolerance.
Any distribution using only their unordered excerpt description incurs TV
error at least \(0.3750\) on one input, under the study's simultaneous
confidence guarantee (Eq.~\eqref{eq:compression-diameter}). This constrains
behavioral approximation, without prescribing a unique encoding or task error rate.

Support on one set of histories does not automatically extend to another. The association state is supported on four histories
for recipient \(A\), a restriction chosen after broader development.
Independent identifier/order changes leave three pairs unresolved under the
fixed bounds; their categorical separations are exploratory. A later comparison
uses new inputs without resampling that anomaly and supports both the simple
state and its revision, so added detail has no demonstrated benefit
(Sections~\ref{sec:crossed-routing}--\ref{sec:routing-refinement}).

The two-decision workflow tests a prescribed requirement: instructions give
current-version assessment, the decision rule and identifier/order invariance,
and the experiment program joins the records. The two coarse summaries are
constant, so their rejections express one missing-test distinction. The instructed
rule predicts every action; fitted forecasts quantify the cost of that omission
for those predictors, without method advantage or optimal-compression claims.
A supplementary final-action comparison shows why readout choice matters:
a summary may preserve release decisions while omitting which tests precede them.

Unresolved adequacy can reflect the comparison design rather than precision
(Table~\ref{tab:response-judgments}). The graph experiment's full-column-rank
state-contrast matrix leaves no nonzero source contrast with zero state change;
more repeats cannot create one. New source directions or matched-state histories are needed.
Tests across systems likewise evaluate each system separately: a shared
representation permits distinct output laws, without pooling them into an
artificial distribution (Section~\ref{sec:information-requirements}).

\paragraph{Representation information and intervention benefit.}
Changing a workflow asks a different question from testing a description of its
behavior. In an exploratory inventory task, explicitly joining requests to
returns restores a required quantity/allocation contrast on one known case.
A direct check of request--return associations selects the same operation,
and fixed joining costs fewer calls. These exploratory comparisons followed
failed calibration; scheduled validation was not performed (SI Section~\siref{si:history-decision}). In a prospective
four-program pilot, replacing one integral corrects that output while both
faulty programs retain incorrect target times or coordinates
(SI Section~\siref{si:prospective-diagnostic-pilot}).

A SciFact development comparison likewise separates response measurement from
benefit. On 89 tasks, the response-guided policy and direct reading both reach
74.72\% correctness, at 7.56 and 1.00 calls per task; ordinary features give
75.28\% at 2.55 calls without response measurements. Initial evidence selection
is invalid or unavailable on 45 tasks, limiting measurement coverage without
explaining the absent gain (SI Section~\siref{si:development-decisions}).
Establishing intervention benefit requires independent tasks, fixed policies,
comparators with the same information, and measurement and execution costs.
These development estimates establish no policy superiority.

A separate constructed software-build workflow uses the same Qwen model to
test versioned job and dependency records. Wrong-component, obsolete-revision and obsolete-build
histories produced READY on all 24 calls; across all eight faulty histories,
none of 32 calls produced a valid HOLD. Overall correctness is 32/64, equal
to always answering READY. Its preliminary representation comparison
selected no candidate, so the planned confirmation was not run.
These observations delimit that task and interface. They are not an ablation
of the closed-loop interface: task, presentation and output requirements also
differ, so the cause of the performance difference is not isolated
(SI Section~\siref{si:workflow-development}).

\paragraph{Scope.}
The AI studies concern accessible workflow records and analyst-selected states.
Closed-loop support concerns one model, fixed version names during confirmation,
and two decisions from a specified preparation; the tested summary omits pending-request information
needed at later times. It establishes neither autonomous reduced dynamics nor
a numerical discretization of the continuous field law. Simulated tool checks
also delimit transfer to live software systems. Interface development failures
and feedback pilots without task benefit remain separately reported
(SI Section~\siref{si:closed-loop}).
Tests of revised representations require fixed downstream components and interfaces,
informative comparisons and independent error controls.
Arbitrary-history sufficiency, causal mediation, autonomous reduction,
cross-model transfer, jointly formed multi-agent states, emergent transitions
and common empirical response laws require separate evidence.

\section{Measurement and reconstruction conventions}
\label{sec:public-methods}

Each comparison fixes its source coordinates, candidate representation, readout,
observation window and external conditions. Fixed-input studies hold the
subsequent input sequence constant. Adaptive studies hold the action policy,
tool-response rule and observation intervention constant, allowing actions
to select later observations. Preparation is fixed except for the
specified source or history contrast. A representation computed only for
analysis does not alter the downstream component's input; a presentation change is instead
a system intervention. These access distinctions are stated for each experiment
in Table~\ref{tab:cross-domain-roles} and the corresponding experimental
sections. Additional intervention protocols appear in SI
Sections~\siref{si:history-decision} and~\siref{si:development-decisions}.

Experimental units are simulation seeds, source directions, program cases or
scientific claims, as specified in each result. Repeated calls and comparisons
that reuse endpoints are not independent tasks. Missing or malformed responses
remain in the accounting, with dependent measurements marked unavailable rather
than replaced by zero. The SI describes the available sampling, parsing and
scoring rules, uncertainty calculations and controls, and identifies gaps in
historical specifications.

Numerical reanalysis recomputes statistics from recorded observations; reexecution
reruns the program or graph. Replication requires new model calls, and
authentication of historical execution requires separate provenance evidence. SI
Table~\siref{tab:reconstruction-coverage} identifies each study's evidential
status and limits. The described methods permit independent implementation;
the absence of public raw records and executable code prevents independent
reconstruction of all reported numerical results from the publication alone.

\FloatBarrier
\paragraph{Acknowledgements.}
\label{iclr:ai-statement}
We thank Herbert Levine, Eduardo Sontag, Jing Wang and Antong Chen for helpful
discussions.
Generative AI tools assisted research design, coding, analysis, literature
review, figure preparation and manuscript editing. The authors take full
responsibility for the work.

\paragraph{Data availability.}
\label{iclr:reproducibility}
The study uses simulated physical data, constructed AI tasks and recorded model
outputs. Public sources comprise SciFact training claims and scientific abstracts;
historical comparisons use author-constructed variants of GAIA tasks. The
manuscript and SI specify the data sources, task construction, measurement
rules and reported results. Raw experimental records have not been publicly
deposited; SI Section~\siref{si:data-code-availability} states the resulting
reconstruction limits.
\paragraph{Code availability.}
The experiment, analysis and plotting implementations are not publicly
released. The mathematical definitions, task rules and statistical methods
are described in the manuscript and SI.
\paragraph{Author contributions.}
T.C. conceived the theory and designed the experiments. S.Y. performed benchmark testing. Both authors analyzed the results and wrote the manuscript.
\paragraph{Funding.} This work received no funding.
\paragraph{Competing interests.} The authors declare no competing interests.

\FloatBarrier
\label{iclr:main-end}
\label{iclr:pre-references-end}
\phantomsection
\label{iclr:references-start}
\bibliographystyle{naturemag}
\bibliography{refs}
\end{document}


\begin{center}
{\Large\bfseries Supplementary Information\\[0.6em]
Active Neural Matter:\\[0.25em]
A Finite-Window Response Framework for\\[0.25em]
Physical Systems and Distributed AI Workflows\par}
\end{center}
\begin{center}
Tianchi Chen\textsuperscript{1,*}, Shihua Yu\textsuperscript{2}\\[0.5em]
\small
\textsuperscript{1}Merck \& Co., Inc., Rahway, NJ, USA\\
\textsuperscript{2}Department of Electrical and Computer Engineering,\\
Northeastern University, Boston, MA, USA\\
\textsuperscript{*}Corresponding author: \href{mailto:chentianchi@gmail.com}{chentianchi@gmail.com}
\end{center}

This Supplementary Information provides derivations, experimental methods and
supporting results. Equations, figures and tables use an S prefix; references
to the main manuscript are identified explicitly.
\setcounter{tocdepth}{1}
\hypersetup{bookmarksdepth=3}
\begingroup
\setstretch{1}
\setlength{\parskip}{0pt}
\tableofcontents
\endgroup
\clearpage
\section{Finite-window response and representation adequacy}
\label{si:response-theory}

\subsection{States, sources and readouts}
\label{si:scope-organization}

Forward response predicts how a source change affects an observable; inverse
comparison tests whether a candidate state preserves the distinctions needed
for that response. The definitions below separate readout adequacy from
autonomous dynamics. The physical specialization in
Section~\ref{si:formal-delayed-model} adds delay histories and well-posedness
conditions; stationary susceptibility also requires closure and stability.

Throughout, an admissible comparison fixes the instance class, interface,
loading coordinate, stopping rule and readouts. State coordinates and dynamics
are realization-specific. For a linear map, \(\operatorname{Im}\) is its
set of possible outputs and \(\ker\) its set of inputs mapped to zero.
In response formulas, \(\Delta\) denotes finite contrasts and \(\delta\)
tangent perturbations; hats mark measured or finite approximations. The derivative \(\mathrm D_xF\)
maps a perturbation of \(x\) to the first-order change in \(F\): a Jacobian
in finite dimensions, or a Fr\'echet derivative on the specified function space.

\begin{definition}[Distributed-system apparatus and experimental instance]
A distributed-system apparatus supplies external memory, tools, evaluation
and execution control around local operators. An instance \(a\) specifies one
task or experiment under fixed interfaces, permissions and stopping rules. It
labels the preparation, not a dynamical coordinate.
\end{definition}

\begin{definition}[Retained state]
For family \(f\), let \(X_f\) be the full state space, including the
variables, required histories, observation records and directly consumed inputs
of that realization, with \(s_t\in X_f\). Fields and local memories are
physical examples. Let \(Z_f\) be a candidate
coordinate space. A declared map
\(G_f:X_f\to Z_f\) gives \(z_t=G_f(s_t)\), the
retained state. This coarse-graining selects information from the full state.
Workflow coordinates may encode unresolved tasks, conflicting
or delayed evidence and pending corrections. Adequacy for a readout is tested
separately; autonomous closure is an additional condition.
\end{definition}

\begin{definition}[Preparation, coarse-graining, and endpoint readout maps]
Preparation \(P_f^{\rm prep}(a)=s_0\) sets the initial full state;
dynamics and inputs give \(s_T\in X_f\). The candidate is
\(z_T=G_f(s_T)\in Z_f\). The endpoint readout
\(O_T:X_f\to\mathcal Y_f\) takes values in a declared observation space;
numerical observables may be scalar or vector valued. A readout depending only
on the retained state would be a map \(\bar O_T:Z_f\to\mathcal Y_f\) with
\(O_T(s)=\bar O_T(G_f(s))\). The bar distinguishes its domain; it is not an
average. This composition, also written \(O_T=\bar O_T\circ G_f\), is a
condition to test. The nested case
uses outer workability readout \(P_f\) and contained endpoint readout
\(Q_f\), with \(w=P_f\), \(r=Q_f/P_f\) for \(P_f>0\)
(main-text Section~\mainref{si:nested-observables}); these differ from preparation.
\end{definition}

\subsection{Window observations as terminal readouts}
\label{si:observation-state}
Trajectory measurements become terminal readouts by retaining their records
in \(s_T\), together with any directly consumed input. For a measured
field component \(c_t\) of \(s_t\), an integral can be recorded by appending \(\dot{\mathbf a}_t=\psi(c_t,t)\), \(\mathbf a_0=0\), so
\(\mathbf a_T=\int_0^T\psi(c_t,t)\,\mathrm dt\). The weight \(\psi\)
selects the window and real/imaginary lock-in components. Amplitude and phase
are functions of these components; differentiable phase requires a local
continuous branch at nonzero amplitude. Other path readouts retain their
corresponding records. These passive observations neither feed back into the
dynamics nor establish candidate adequacy or autonomous stability.

\subsection{A falsifiable retained-state criterion}
\label{si:falsifiable-retained-state}

A candidate state misses a response-relevant distinction if a source change
leaves that state unchanged but changes the endpoint. The first-order test
compares the source-to-state and source-to-endpoint maps on the same measured
perturbations; it does not assume a state-only endpoint law in advance.
At a declared operating point, let \(\mathcal U_T\) be the finite-dimensional
measured perturbation space and \(\mathcal Z_T,\mathcal Y_T\) the vector
spaces of retained-state and endpoint variations. Define the first-order maps
\[
\mathcal K_{z,T}:\mathcal U_T\to \mathcal Z_T,
\qquad
\mathcal K_{O,T}:\mathcal U_T\to \mathcal Y_T
\]
from sources to states and endpoints, respectively.

On the declared time grid, let \(u_t\) denote the controlled source input
and let the full-state transition be
\(s_t=F_t(s_{t-1},u_t)\), with \(A_t=\mathrm D_sF_t\) and
\(B_t=\mathrm D_uF_t\) along the baseline trajectory. Define
\(\Phi(T,t)=A_T\cdots A_{t+1}\) and \(\Phi(T,T)=I\), the identity map.
Repeated substitution in its tangent recursion gives
\[
\delta s_T=\Phi(T,0)\delta s_0+
\sum_{t=1}^{T}\Phi(T,t)B_t\delta u_t.
\]
Applying \(\mathrm D_sG_f\) and \(\mathrm D_sO_T\) gives the
two response maps without assuming factorization. Allowed preparation changes
join source changes in \(\mathcal U_T\); fixed preparation sets
\(\delta s_0=0\), as in main-text Eq.~\mainEqref{eq:finite-window-endpoint-response}.

\begin{proposition}[First-order retained-state criterion]
There is a unique linear readout map, with superscript \((1)\) denoting first order,
\(\mathcal H_T^{(1)}:\operatorname{Im}\mathcal K_{z,T}\to \mathcal Y_T\) satisfying
\[
\mathcal K_{O,T}=\mathcal H_T^{(1)}\mathcal K_{z,T}
\]
on the measured perturbation space if and only if
\[
\ker\mathcal K_{z,T}\subseteq\ker\mathcal K_{O,T}.
\]
Uniqueness is only on the state displacements reached by the measured
perturbations; the result makes no claim about unprobed state directions.
\end{proposition}

\emph{Derivation.}
If the factorization holds, every perturbation that leaves the retained state
unchanged also leaves the endpoint unchanged. Conversely, under the kernel
inclusion define
\(\mathcal H_T^{(1)}(\mathcal K_{z,T}\delta u):=\mathcal K_{O,T}\delta u\).
Two perturbations producing the same state displacement differ by a vector in
\(\ker\mathcal K_{z,T}\); the inclusion makes their endpoint displacements equal.
The resulting map is therefore well-defined and linear, and fixes \(\mathcal H_T^{(1)}\) uniquely on
\(\operatorname{Im}\mathcal K_{z,T}\).

The delayed-history example in main-text Section~\mainref{iclr:physics}
gives a present-state null direction with a nonzero later readout response.
Equal current values do not mean equal initial histories, so solution
uniqueness is preserved. The violation requires a revised description to
separate those histories without prescribing a unique encoding.

The criterion remains local and first order, rather than an exact nonlinear
factorization theorem. Let \(\mathsf z_T(u)\) and \(\mathsf o_T(u)\) denote the
nonlinear retained-state and endpoint maps from a source history \(u\). Exact
sufficiency on a finite neighborhood instead requires that any two source
histories producing the same retained state also produce the same endpoint:
\[
\mathsf z_T(u_1)=\mathsf z_T(u_2)
\quad\Longrightarrow\quad
\mathsf o_T(u_1)=\mathsf o_T(u_2).
\]
For continuously differentiable (\(C^1\)) maps between finite-dimensional coordinate spaces,
constant rank of \(\mathsf z_T\) and kernel inclusion
throughout a neighborhood imply factorization on a sufficiently small local
neighborhood. To see why, use constant-rank coordinates \((x,y)\) in which
\(\mathsf z_T\) depends only on \(x\) and distinguishes nearby \(x\) values.
The \(y\) directions lie in \(\ker D\mathsf z_T\), so kernel inclusion gives
\(D_y\mathsf o_T=0\). On a sufficiently small product neighborhood,
\(\mathsf o_T\) is independent of \(y\) and is a function of \(\mathsf z_T\).
Without the constant-rank condition,
pointwise kernel inclusion alone is insufficient: \(\mathsf z(u)=u^2\) and
\(\mathsf o(u)=u^3\) satisfy the derivative-kernel condition at every scalar
\(u\), yet the arbitrarily close inputs \(u=\epsilon\) and \(u=-\epsilon\)
have the same retained state and different endpoints.

\paragraph{Interpreting measured contrasts.}
A resolved same-state violation can reject adequacy without a separate
positive control. Informative empirical local support covers only tested
comparisons and requires nontrivial null directions or matched histories, a
detectable endpoint positive control, adequate comparison coverage, specified
equivalence bounds and controlled matching error. These support conditions are
not necessary for mathematical sufficiency: a constant readout factors through
a constant state.

Fix source intervention, units, window and endpoint rule; estimate state and
endpoint contrasts and their resolution separately. Joint changes establish
association under intervention, not mediation; state changes without endpoint
changes can reflect cancellation or robustness. Unresolved responses do not
localize effects.

For an approximate match with a deterministic readout, suppose the true state
difference is bounded by \(\varepsilon_z\), the error in the measured endpoint
contrast \(\widehat{\Delta O}\) by \(\varepsilon_O\), and candidate readouts
\(O=h(z)\) have a Lipschitz bound \(L_h\) on the compared states justified
independently of the tested endpoint contrasts.
Factorization then requires
\(\|\widehat{\Delta O}\|\le L_h\varepsilon_z+\varepsilon_O\).
Exceeding the bound rejects this Lipschitz readout class. Exact matching sets
\(\varepsilon_z=0\); approximate matching cannot reject arbitrary readouts
without independent sensitivity control. Stochastic laws need a specified
comparison and uncertainty bound.
For example, \(z=u\), \(O=2u\) preserves the input but rejects \(L_h\le1\)
on every distinct pair when \(\varepsilon_z=|\Delta u|\),
\(\varepsilon_O=0\). This rejects the sensitivity constraint, not the state;
check the readout assumptions and matching before refining the representation.
The violation cannot itself justify a replacement bound.

\paragraph{Inverse use: distinctions a representation must preserve.}
Response equivalence groups histories whose specified responses agree under
every common continuation, fixing sources, readouts, windows and noise laws.
Compare conditional laws for distributional claims, means for mean claims,
and baseline levels when levels rather than contrasts are targeted. The
coarsest sufficient
partition constrains refinement without supplying a reconstruction algorithm
or autonomous realization. Related constructions include causal states and
predictive-state representations~\citep{shalizi2001computational,littman2001psr}
(main-text Section~\mainref{si:related-work}). Finite tests constrain only
their comparisons;
tolerance-based matching need not be transitive.

In the workflow examples, the downstream component is the program or language
model that reads the records and produces the output. Main-text Section~\mainref{si:consumer-family} gives the finite-history
shared-representation criterion and its measurability argument;
main-text Eq.~\mainEqref{eq:public-representation-transfer} gives the transfer bound.
Both concern declared systems and continuations. Transfer additionally
requires an independently justified uniform bound on the change of readout
law, which finite measurements alone do not supply.

\paragraph{State measurements and selective interventions.}
This deterministic example asks which measurement or selective intervention
distinguishes altered propagation from altered readout. In fixed physical
units, let \(g_p\) be the propagation gain and \(g_r\) the readout gain.
Assume
\[
z=g_p u,\qquad O=g_r z,
\]
with zero baseline, no direct source--endpoint path, and calibrated healthy
gains $g_p=g_r=1$. The two changes $(g_p,g_r)=(1/2,1)$ and $(1,1/2)$ have the same
complete source--endpoint map $O=u/2$. Before state-clamp or repair feedback,
source interventions observed only at the endpoint cannot distinguish them.
State measurements do distinguish $z=u/2$ from $z=u$. Alternatively, a known
selective clamp, written $\operatorname{do}(z=v)$, sets the state to $v$
while holding $u$ fixed. It gives $O=g_r v$ and identifies the readout
gain without separately measuring the natural state.

Estimate \(g_p\) from source--state contrasts and \(g_r\) from independent
clamps, then compare predicted and measured endpoints and calibrated gains.
With one reset allowed, reset the sole changed gain, leave healthy systems
alone, and abstain when both change. At \(u=1\), the correct reset takes
\(O=1/2\) to \(1\), while the other leaves \(O=1/2\). These are equation-based
deductions; ordinary state-space identification with the same information gives
the same decisions.

The assumptions matter: $z=u$, $O=2z$ and $z=u$, $O=2u$ give identical
source/state/endpoint records although only the first endpoint depends on the
state. A selective state clamp separates them. A fit to source interventions
alone therefore cannot certify mediation. Outside the declared class, a
mismatch calls for competing mechanisms rather than a forced component label.

Decision value also depends on the readout. For $\delta z=\operatorname{diag}(1,4)
\delta u$ and $\delta O=(1/4,1/100)\delta z$, positive unit source steps have
state norms $1,4$ but endpoint changes $1/4,1/25$. For a fixed upper-boundary
gap $1/4$, equal step costs and allowed steps of at most one, only the first
axis reaches the boundary. Standard constrained control gives the same choice.
Exact-arithmetic implementation checks verify these deductions,
including zero-gain, healthy, multiple-change and direct-path controls.
These are analytic checks, not measured AI fault or repair results.

\subsection{Source encoding and response equivalence}
\label{si:source-encoding}

When a workflow encodes evidence as numerical sources, an explicit map
connects visible records to source locations and amplitudes. Its specification
identifies the visible inputs, event fields, preparation and admission rules,
and assignment of numerical sources. The definition below separates these
operations; main Section~\mainref{si:worked-graph} gives the implemented graph
construction, state update and readout together, and
Section~\ref{si:live-paired-source-perturbation} gives its sampling procedure.
Encoding records as a graph is one realization: a language model may instead
read records directly.

\begin{definition}[Evidence-to-field source map]
Let \(E_f^{\le t}\) be the set of evidence histories up to time \(t\).
For a history \(\eta\in E_f^{\le t}\), proposed, prepared and
admitted records \(e_f^{\rm prop},e_f^{\rm prep},e_f^{\rm adm}\) form the
source chain
\[
\eta
\xrightarrow{\xi_f}
e_f^{\rm prop}
\xrightarrow{A_f^{\rm prep}}
e_f^{\rm prep}
\xrightarrow{A_f^{\rm adm}}
e_f^{\rm adm}
\xrightarrow{J_f}
j_f(\eta)\in Y_f^{\rm src} .
\]
Here \(Y_f^{\rm src}\) is the source-injection space. Extraction
\(\xi_f\) proposes events; preparation \(A_f^{\rm prep}\) preserves evidence visible before
verification, applies prespecified support-amplitude bounds based on that evidence,
and excludes endpoint labels. Admission \(A_f^{\rm adm}\) selects events
allowed to act. Finally \(J_f\) assigns numerical amplitudes, sites, channels
and operator tags. The composed source map is
\(j_f=J_f\circ A_f^{\rm adm}\circ A_f^{\rm prep}\circ\xi_f\),
with the stages applied in the order shown above.
The evidence process, action set, event alphabet, relation types and verifier
are family-specific; the source-chain structure applies to simulations,
recorded data and model interfaces.
\end{definition}

\subsubsection{When the source-event representation is sufficient}
\label{si:representation-sufficient}

A continuation fixes later inputs or external response and observation rules,
and the readout time, while preserving interface, dynamics and stopping rules.
For an adaptive system it fixes the policy rather than its sampled actions
(main-text Section~\mainref{sec:adaptive-response}). Source sufficiency asks whether the event representation preserves the
observable effects of evidence histories under every allowed continuation.

\begin{definition}[Source-representation sufficiency]
\label{prop:representation-sufficiency}
Fix an admissible neighborhood, a common set of admissible continuations,
a horizon \(T\), and the scalar or vector readouts to be compared. Include
verifier outputs or operating criteria among these readouts when required.
Let
\(\rho_f:E_f^{\le t}\to R_f^{\rm rep}\) map evidence histories to a
candidate representation space \(R_f^{\rm rep}\).
Define \(\eta_1\equiv_f\eta_2\) when replacing
\(j_f(\eta_1)\) by \(j_f(\eta_2)\) in every common admissible continuation
leaves the declared readouts unchanged. For deterministic readouts, this means
equality after the same fixed calibration or quantization map. For stochastic
readouts, compare the declared conditional statistic or readout distribution
given each evidence history under the same continuation and noise law.
This defines an equivalence relation,
rather than a pairwise tolerance test. In the nested specialization, the
readouts include \(P_f,Q_f\), from which \(w=P_f\) and \(r=Q_f/P_f\)
are derived where \(P_f>0\).
Then \(\rho_f\) is sufficient for the local
source representation if and only if no two inequivalent histories are assigned
the same representation,
\[
\rho_f(\eta_1)=\rho_f(\eta_2)
\quad\Longrightarrow\quad
\eta_1\equiv_f\eta_2 .
\]
The coarsest sufficient representation is obtained when each representation
class is exactly one equivalence class of \(\equiv_f\).
This is an abstract quotient; a finite-valued representation exists only when
there are finitely many classes. The quotient argument alone supplies no
finite-dimensional regular or autonomous realization.
\end{definition}

\emph{Derivation.}
If two inequivalent evidence histories are assigned the same representation,
then any source construction that factors through that representation
assigns them the same numerical source term although some admissible
continuation separates them through a declared readout. Such a representation
cannot reproduce all specified
continuation readouts. Conversely, if each representation class contains only
histories equivalent under \(\equiv_f\), then, for each fixed admissible
continuation, its calibrated deterministic values or declared conditional
statistics or laws are functions of the representation. This establishes
readout sufficiency on those tests;
autonomous evolution of the representation requires a separate closure condition.

Counting failed checks can discard what each failure concerns. A rejection
of a relation in use and a delayed warning about an unused candidate may have
the same count but different source locations, signs or delay channels in
\(Y_f^{\rm src}\). The count is insufficient only if an allowed continuation
makes their declared \(P_f,Q_f\) or \(O_T\) readouts differ. A revision must
retain the locations, channels or operator distinctions needed to separate
such responses. Differences may be discarded only when every declared
continuation leaves them invisible to the readout.

\subsection{Comparing realization-specific retained responses}
\label{si:comparing-retained-responses}

Comparing realizations requires a state correspondence and an endpoint map;
nested-readout agreement alone need not preserve other observables.

\begin{proposition}[Retained response and endpoint readout]
Let \(\mathcal Z_i^{\mathrm{reach}}\subset Z_{f_i}\), \(i=1,2\), be the retained states reached in two
realizations, and let \(J:\mathcal Z_1^{\mathrm{reach}}\to \mathcal Z_2^{\mathrm{reach}}\) be a specified comparison map.
Assume each endpoint factors through its retained coordinate, with maps
\(\bar O_{f_i,T}\) into a common declared readout space. Exact endpoint values agree on corresponding states \(z_2=J(z_1)\) if and only if
\(\bar O_{f_2,T}\circ J=\bar O_{f_1,T}\) there. If the reached sets are locally smooth
and these maps are differentiable, first-order endpoint changes under
\(\delta z_2=\mathrm D J(z_1)\delta z_1\) agree if and only if
\[
\mathrm D\bar O_{f_2,T}(J(z_1))\,\mathrm D J(z_1)\delta z_1
=\mathrm D\bar O_{f_1,T}(z_1)\delta z_1
\]
for every declared reached tangent direction \(\delta z_1\).
Parentheses specify where each derivative is evaluated; \(J\) need not
be linear.
For the optional nested pair, if
\((P_{f_2},Q_{f_2})\circ J=(P_{f_1},Q_{f_1})\), corresponding states have
the same reduced readouts \((w,r)\) for every
\(z_1\in\mathcal Z_1^{\mathrm{reach}}\) with \(P_{f_1}(z_1)>0\).
\end{proposition}

\emph{Derivation.}
Exact endpoint agreement is the composition identity for
\(\bar O_{f_i,T}\); the derivative condition follows by the chain rule
on corresponding displacements. Exact composition implies this condition;
equal derivatives alone do not imply equal endpoint values, since the maps
may differ by an additive constant. For scalar endpoint values, agreement
only of their declared ordering is weaker than equality of the values.
The nested-pair statement follows by composition with \(J\) and division
by the common positive outer readout.

\emph{Interpretation.}
Preparation, dynamics and readout projection remain realization-specific. No
comparison map \(J\) is assumed between the reported physics and AI systems.
A comparison requires specified state correspondence or matched readout
semantics; different projections can resolve different changes.

\subsection{Minimal representation of a terminal response}
\label{si:minimal-realization}
\label{sec:minimal-realization}

For fixed sources, terminal readout and finite horizon, linear response
identifies a minimal representation of the reached displacements visible to
that readout. This differs from independently declaring a physical retained
state. Consider a time-independent closed linear realization with
retained-state transition \(A\), source
injection \(B\), and differentiably factored terminal readout
\(O_T=\bar O_T\circ G_f\). For
\(\delta z_t=A\,\delta z_{t-1}+B\,\delta u_t\) with zero initial displacement,
write \(C_T=\mathrm D_z\bar O_T\). Let \(\mathcal R_T\) contain terminal
displacements reachable by the allowed sources, and \(\mathcal N_T\)
contain displacements invisible to the terminal readout:
\[
\mathcal R_T=\operatorname{Im}[B,AB,\ldots,A^{T-1}B],
\qquad \mathcal N_T=\ker C_T .
\]
For time-dependent transitions, \(\mathcal R_T\) is instead the image of the
finite-window source-to-state map. The quotient, which identifies
displacements differing by an invisible component,
\(\mathcal R_T/(\mathcal R_T\cap\mathcal N_T)\) is isomorphic to
\(C_T\mathcal R_T\): it retains exactly the terminal displacements that this
readout distinguishes. The discarded component may matter for another readout or later time.
This quotient need not evolve autonomously. For a finite-dimensional, time-independent linear
input--output system, a minimal dynamical realization is reachable and
observable. Constructing a dynamical quotient uses transition-invariant
reachable and unobservable subspaces; the kernel of one terminal readout
need not be invariant.

An effective dimension may instead summarize singular-value concentration.
Let \(\sigma_k\) be the singular values of a nonzero finite-window response
matrix at a stated
horizon. For a time-independent closed transition with spectral radius
\(\rho(A)<1\), the stationary choice is
\(\chi_\ast^G=(I-A)^{-1}B\), derived in
Section~\ref{si:reduced-state-equation}. For the modal energies
\(\sigma_k^2\), the participation ratio is
\[
r_{\rm PR}=\frac{\bigl(\textstyle\sum_k\sigma_k^2\bigr)^2}{\sum_k\sigma_k^4}
\]
Endpoint projection by \(\mathrm D_z\bar O_T\) can reduce rank; an
injective non-isometric readout can still change the singular-value spectrum.
Terminal minimality is relative to the declared source and readout.

The constructed code-repair graph in Section~\ref{si:empirical-protocols}
retains \(32\) coordinates. Identity observation under a constructed full
source basis does not establish empirical minimality for the endpoint pair
\(O=(P_f,Q_f)\).

\section{Field dynamics and conditional response}
\label{si:physical-field-law}

\subsection{Formal delayed field-coupled model}
\label{si:formal-delayed-model}

The field model realizes finite-window response through medium propagation,
local sources and prescribed delayed inputs. Their distinction determines
which histories enter the dynamical state.
Let \(H=L^2(X,\mathbb R^q)\), or another separable Hilbert field space,
with \(X\) a spatial or graph domain and \(q\) realization-specific channels.
Here \(L^2\) is the space of square-integrable fields.
The shared field is \(c_t\in H\). Local operator \(i\), among \(N\)
operators, has memory \(m_i(t)\in M_i\) in a finite-dimensional or Banach
space and contributes \(J_i(c_t,m_i(t))\in H\). Physical channels may encode
activity; workflow channels may encode candidate weights, conflicting evidence,
inhibition or tool availability. The field model is
\begin{equation}
\partial_t c_t
=
L c_t
+
\sum_{i=1}^{N} J_i(c_t,m_i(t))
+
\sum_{\ell=1}^{N_\tau} E_\ell y_\ell(t-\tau_\ell),
\label{eq:supp-universal-interface}
\end{equation}
Here \(L\) generates autonomous medium evolution. The \(N_\tau\) channels
supply prescribed histories \(y_\ell(t-\tau_\ell)\in Y_\ell\), whose
generators lie outside the modeled operators, through bounded couplings
\(E_\ell:Y_\ell\to H\). Internal delayed feedback belongs to \(J_i\),
including the consensus coupling. The displayed local-source arguments
suppress required internal histories; the augmented state below includes them
explicitly.

For instantaneous reads, a field-coupled local operator comprises
a read operator \(R_i:H\times M_i\to O_i\), a local update map
\(\pi_i:O_i\to U_i\), a write operator \(W_i:U_i\to Y_i^{\rm src}\), a
source-coupling operator \(G_i:Y_i^{\rm src}\to H\), and a memory-update rule
\(\Psi_i\). Its source contribution is
\[
J_i(c,m_i)=G_iW_i\!\left(\pi_i(R_i(c,m_i))\right).
\]
Here \(O_i\) is the local readout space, \(U_i\) the update-output space,
and \(Y_i^{\rm src}\) the source space exposed by operator \(i\).

Assume \(L\) generates a strongly continuous contraction semigroup on \(H\);
spatial and graph examples appear in Table~\ref{tab:operator-dictionary},
with diffusion coefficient or tensor \(D\), relaxation operator \(\Gamma\)
and graph Laplacian \(L_G\). For a PDE, the local write can enter through an injection kernel,
\[
(G_iw)(x)=\int_X g_i(x,\xi)w(\xi)\,d\xi,
\]
where \(w\) denotes the written source density in this formula. On a finite
graph, \(G_i\) selects the sites and channels receiving the write.

The injection kernel \(g_i\) places a write into the field; the Green
kernel \(K_L\) of \(\partial_t-L\) propagates it. When the propagator admits
a kernel, interpreted weakly when necessary, a specified source contribution is
\[
\delta c(t,x)
=\int_{0}^{t}\!\int_X
K_L(t-s;x,x')\,(G_i\delta w_s)(x')\,dx'\,ds,
\]
with the full coupled response additionally including derivatives of local sources,
memory dynamics and internal feedback, incorporated in the augmented state
below.

An external delayed coupling has finite gain \(\beta_\ell\):
\[
\|E_\ell y\|_{H}\le \beta_\ell\|y\|
\]
on the relevant signal space. For \(\|y_\ell\|\le M_\ell\) on the
sampled history, forcing is bounded
by \(\beta_\ell M_\ell\) and its integrated squared norm by
\(T\beta_\ell^2M_\ell^2\). A bounded coupling alone does not bound arbitrary
inputs or long-time state growth. Table~\ref{tab:operator-dictionary}
summarizes the operators.

\begin{table}[tbp]
\centering
\caption{\textbf{Field operators.} Sources resolve as
\(J_i=G_iW_i(\pi_i(R_i))\). Here \(D\), \(\Gamma\) and \(L_G\)
denote diffusion, relaxation and the graph Laplacian. Consensus delay is
internal; \(E_\ell\) couples externally generated records, even when
an operator action caused them.}
\label{tab:operator-dictionary}
\label{tab:rpw-specialization}
\begin{tabularx}{\linewidth}{@{}l>{\raggedright\arraybackslash}X>{\raggedright\arraybackslash}X@{}}
\toprule
\textbf{Symbol} & \textbf{Physical meaning} & \textbf{Typical realization} \\
\midrule
\(L\) & autonomous propagation, relaxation, and dissipation of the shared medium & diffusion--decay operator \(D\Delta-\Gamma\), graph Laplacian generator \(-D L_G-\Gamma\), or Markov/contraction generator \\
\addlinespace[0.25em]
\(J_i\) & composite source generated by modeled local operator \(i\) & feedback force, actuator drive, graph-node source, or typed evidence increment \\
\addlinespace[0.25em]
\(R_i\) & local readout available to operator \(i\) from the field and memory & projection, local averaging, attention window, tool-state readout \\
\addlinespace[0.25em]
\(\pi_i\) & local update or feedback law used by operator \(i\) & controller law, consensus update, model inference, decoding, or tool choice \\
\addlinespace[0.25em]
\(W_i\) & local write produced by operator \(i\)'s update rule & signed source density, graph-node injection vector, channel-specific evidence update \\
\addlinespace[0.25em]
\(G_i\) & source-coupling operator placing the local write into the medium & spatial source kernel \(g_i(x,\xi)\) or finite source matrix on a graph \\
\addlinespace[0.25em]
\(E_\ell\) & finite-gain coupling from delayed input outside the modeled local operators into the medium & verifier, tool, environment, or measurement channel with delay \(\tau_\ell\) \\
\bottomrule
\end{tabularx}
\end{table}

Continuous memory dynamics are specified by
\begin{equation}
\dot m_i(t)=\Psi_i(c_t,m_i(t),\eta_i(t)),
\qquad i=1,\ldots,N,
\label{eq:supp-memory-law}
\end{equation}
where \(\eta_i(t)\) is bounded apparatus input. The well-posedness result
below applies under its continuous-time hypotheses. Discrete or hybrid memory
needs its own recurrence or jump law. Externally recorded memory is instead a
prescribed input, subject to source regularity.

\begin{proposition}[Equivalent source terms]
\label{prop:interface-equivalence}
Fix the same initial field, the same external inputs and all other
field-source histories. If two local mechanisms expose identical effective
source histories,
\[
J_i(c_t,m_i(t))=J_j(c_t,m_j(t))
\]
for almost every \(t\in[0,T]\), variation of constants gives the same field
trajectory on that horizon. Every declared readout depending only on this
field trajectory therefore agrees. For readouts that also use separately
retained memories or other coordinates, equal additional inputs are sufficient
for agreement; otherwise the endpoints must be compared through the full
readout map.
\end{proposition}

Resolved component maps permit selective interventions, but equal exposed
sources do not identify their generators. Component diagnosis needs additional
measurements or interventions; workflow retriever, verifier and tool outputs
supply source or input channels.

\subsection{History, projection and conditional autonomous response}
\label{si:reduced-state-equation}
\label{sec:slow-state-derivation}

Projection preserves selected information; autonomous evolution requires
closure. Delayed dynamics must first be described on a state containing their
internal histories.

\paragraph{Include internal history before projecting.}
For each of the \(N_{\rm int}\) internally generated delayed quantities
\(q_a\), retain
\(\xi_a^t(\theta)=q_a(t+\theta)\) on \([-\tau_a,0]\). Together with the
field and local memories, these histories form
\[
s(t)=(c_t,m_1(t),\ldots,m_N(t),\xi_1^t,\ldots,\xi_{N_{\rm int}}^t)
\in X_f.
\]
Each \(q_a\) must be a specified causal function of modeled variables;
any additional signal generator needs its own state equation. Internal
histories evolve by \(\partial_t\xi_a^t=\partial_\theta\xi_a^t\), with
boundary value \(\xi_a^t(0)=q_a(s(t))\). Here \(u\) denotes the control
parameters. Field evolution, memory laws
\(\dot m_i=\Psi_i(c,m_i,\xi;u)\), and these history shifts define
\begin{equation}
\dot s=F_{\rm full}(s;u,y^t).
\label{eq:supp-full-state-flow}
\end{equation}
External histories \(y^t\) remain prescribed inputs. The history space
for mild solutions is specified in Section~\ref{si:well-posedness-coherence};
these derivatives require classical solutions in the generator domain. Append
observation records when needed (Section~\ref{si:observation-state}). Declare
\(z=G_f(s)\in Z_f\) before testing factorization: the current
field can omit relevant history, and endpoint factorization alone does not
establish autonomous closure.

\paragraph{Closure on an invariant branch.}
Fix control \(u\) and stationary external input \(\bar y\). Suppose a
locally invariant branch admits a continuously differentiable (\(C^1\))
parametrization \(s=h_u(z)\)
with \(G_f\circ h_u=\operatorname{id}\), where \(G_f\)
is also \(C^1\). Branch trajectories must be classical solutions: fields lie
in the operator domain \(D(L)\), and histories have compatible derivatives and boundary traces.
Assume \(F_{\rm full}(h_u(z);u,\bar y)\) is tangent to this branch. Then
\(\mathrm D_sG_f\,\mathrm D_zh_u=I\), and the chain rule gives
\begin{equation}
\dot z=\mathrm D_sG_f(h_u(z))
 F_{\rm full}(h_u(z);u,\bar y)=:f_u(z).
\label{eq:supp-slow-state-law}
\end{equation}
Exact closure holds for trajectories initialized on the branch at fixed
\(u\). Changing control can move the branch away from the same full-state
preparation, so \(f(z,u)=f_u(z)\) alone does not establish controlled transients.
A sufficient extension is
\[
\mathrm D_sG_f(s)F_{\rm full}(s;u,\bar y)
=f(G_f(s),u),\qquad s\in\mathcal D,
\]
for every admitted control, on a common domain \(\mathcal D\subset X_f\)
containing the preparations and invariant under admitted controlled evolution,
with classical solutions. Without exact closure, a reduced transient approximation
requires initialization and tracking error bounds, or the missing variables must
remain in the response state. Attraction alone does not eliminate transients
from initialization away from the branch. Arbitrary external forcing requires
a controlled description or its generator in \(s\). Finite-window response
requires neither branch reduction nor a potential representation.

\paragraph{Stable equilibrium and discrete response.}
Equilibrium susceptibility follows from stationary balance; identifying it
with accumulated transient response requires additional stability and closure
conditions.
For a finite-dimensional branch family \(f(z,u)=f_u(z)\) that is \(C^1\)
jointly in state and control on a common coordinate neighborhood, let
\(z_\ast\) be a stationary point at \(u_\ast\). The linearized reduced
equation is
\begin{equation}
\delta\dot z=A_z\delta z+B_z\delta u,\qquad
A_z=\mathrm D_z f(z_\ast,u_\ast),\quad
B_z=\mathrm D_u f(z_\ast,u_\ast),
\label{eq:supp-slow-linearization}
\end{equation}
If \(A_z\) is invertible, the implicit-function theorem gives a local
stationary branch, and differentiating
\(f(z_\ast(u),u)=0\) yields
\begin{equation}
\delta z_\ast=\chi_\ast\delta u,\qquad
\chi_\ast=-A_z^{-1}B_z.
\label{eq:supp-linear-branch-response}
\end{equation}
Tangency makes \(h_u(z_\ast(u))\) a full-system equilibrium; this stationary
susceptibility needs no controlled transient closure.
If \(A_z\) is also Hurwitz (all eigenvalues have negative real parts), the
equilibrium is locally attracting within its fixed-control branch. For the
linearized reduced equation, a constant source and zero displacement give the
integral of \(e^{A_z t}B_z\), converging to \(-A_z^{-1}B_z\). Under the
controlled closure above, this also gives the full system's retained-state
linear response; an approximation instead requires an explicit response-error
bound. A moving equilibrium adds \(-\chi_\ast(u)\dot u\) to its deviation
equation and requires a separate tracking bound.

For a verified closed retained-state transition, with a differentiably
factored readout \(\bar O_T\), let \(A_t^{\rm ret}\) and \(B_t^{\rm ret}\)
be its Jacobians with respect to the retained state and source input along
the reference trajectory. Write
\begin{equation}
\begin{aligned}
\delta z_t&=A_t^{\rm ret}\delta z_{t-1}+B_t^{\rm ret}\delta u_t,\\
\Phi_{\rm ret}(T,t)&=A_T^{\rm ret}\cdots A_{t+1}^{\rm ret},\\
\delta O_T&=\mathrm D_z\bar O_T\left[
\Phi_{\rm ret}(T,0)\delta z_0+
\sum_{t=1}^{T}\Phi_{\rm ret}(T,t)B_t^{\rm ret}\delta u_t\right].
\end{aligned}
\label{eq:supp-retained-window-response}
\end{equation}
Here \(\Phi_{\rm ret}(T,T)=I\); changing initial history can make
\(\delta z_0\ne0\). For time-independent \(A,B\) with spectral radius
\(\rho(A)<1\) (the spectrum lies inside the unit circle), a persistent tangent
source has the time-summed response
\[
\chi_\ast^G=\sum_{k=0}^{\infty}A^kB=(I-A)^{-1}B.
\]
Under exact zero-order-hold discretization,
\(A=e^{A_z\Delta t}\) and
\(B=\int_0^{\Delta t}e^{A_zs}B_z\,\mathrm ds\), so
\((I-A)^{-1}B=-A_z^{-1}B_z\). This follows from
\(\int_0^{\Delta t}e^{A_zs}\,\mathrm ds=A_z^{-1}(A-I)\).
An approximate propagator requires a consistent source discretization.
The stable endpoint response is therefore
\begin{equation}
\delta O=(\mathrm D_z\bar O)\chi_\ast\delta u
\quad\text{or}\quad
\delta O=(\mathrm D_z\bar O)\chi_\ast^G\delta u.
\label{eq:supp-linear-endpoint-response}
\end{equation}
The graph checks in Section~\ref{si:empirical-protocols} test this composition
under prescribed dynamics. They do not identify an autonomous language-model
state.

\subsection{Conditional well-posedness and coherence}
\label{si:well-posedness-coherence}

The following hypotheses give unique finite-horizon evolution from complete
preparations, including internal histories and prescribed external inputs.
Uniqueness does not imply attraction.

For a finite maximum internal delay \(\tau_{\max}\), take the dynamical base
space \(\mathcal B_{\rm dyn}=H\times\prod_i M_i\), with Banach memory spaces,
and the phase space
\(\mathcal X_\tau=C([-\tau_{\max},0];\mathcal B_{\rm dyn})\) with the supremum
norm. Write \(x(t)=(c_t,m_1(t),\ldots,m_N(t))\) and
\(\zeta_t(\theta)=x(t+\theta)\). The prescribed continuous preparation
\(\zeta_0\) has endpoint \(x(0)\). Internal quantities and their histories are
specified functionals of this modeled history; any extra generator is included
in \(\mathcal B_{\rm dyn}\). Point evaluation is bounded in this phase space.
Passive observation records are subsequently defined by their integrals when
integrable, rather than assumed to belong to the dynamical phase space.

The later coherence estimate additionally requires regularity and aggregate
dissipativity. It is separate from finite-window response, readout boundaries
and the branch assumptions in Section~\ref{si:reduced-state-equation}.

\begin{samepage}
\begin{assumption}[Well-posed delayed field system]
\label{assump:wellposed-delayed-field}
For any finite horizon \([0,T]\):
\begin{enumerate}
\item \(L\) generates a strongly continuous contraction semigroup on
      \(H\);
\item the combined source and memory functionals are defined on all of
      \(\mathcal X_\tau\), are Lipschitz on each bounded history set and have
      at most linear growth in \(\|\zeta\|_\infty\). Write their values as
      \(J_i(\zeta)\) and \(\Psi_i(\zeta,\eta_i)\), retaining all internal
      delay arguments. Global componentwise Lipschitz bounds suffice for
      finite compositions; local bounds require the separate growth condition;
\item the apparatus inputs \(\eta_i\) are strongly measurable and essentially
      bounded on finite horizons. Explicit time/input dependence of the
      functionals is strongly measurable at each fixed history, with Lipschitz
      and growth bounds uniform on finite horizons and bounded input sets;
\item each externally prescribed signal \(y_\ell\) is supplied on
      \([-\tau_\ell,T]\), is strongly measurable there and essentially bounded on
      that interval; and
\item each \(E_\ell:Y_\ell\to H\) is a bounded linear operator.
\end{enumerate}
\end{assumption}
\end{samepage}

\begin{proposition}[Conditional mild well-posedness]
Under the ``Well-posed delayed field system'' assumptions above,
\eqref{eq:supp-universal-interface} together with
\eqref{eq:supp-memory-law} admits a unique mild solution on
\([0,T]\) for the compatible continuous preparation \(\zeta_0\), with external
delayed signals treated as prescribed inputs. If the functionals are defined
only on an open neighborhood, the conclusion holds only up to exit from that
neighborhood unless continuation is separately established.
\end{proposition}

\emph{Derivation.}
Variation of constants gives the field for \(0\le t\le T\), with the
negative-time history prescribed by \(\zeta_0\)
\citep{pazy1983,henry1981,hale1993,fridman2014}:
\[
c_t=e^{tL}c_0+
\int_0^t e^{(t-s)L}
\Big(
\sum_i J_i(\zeta_s)+\sum_\ell E_\ell y_\ell(s-\tau_\ell)
\Big)\,ds.
\]
The memory components satisfy the integral equations
\[
m_i(t)=m_i(0)+\int_0^t \Psi_i(\zeta_s,\eta_i(s))\,ds .
\]
Measurability and growth make the Bochner integrals well-defined. Contraction
of the free semigroup and bounded-set Lipschitz continuity make their integral
map contractive on a sufficiently small ball in
\(C([-\tau_{\max},T];H\times M_1\times\cdots\times M_N)\), giving a unique
local mild solution. For
\(r(t)=\sup_{-\tau_{\max}\le s\le t}\|x(s)\|\), linear growth gives
\(r(t)\le r(0)+a t+b\int_0^t r(s)\,ds\), hence
\(r(t)\le(r(0)+a t)e^{bt}\), with \(a,b\ge0\) on the fixed horizon.
Global definition and bounded-set Lipschitz continuity therefore permit
continuation to \([0,T]\), without implying long-time boundedness or classical
differentiability.

This deterministic result covers the stated bounded prescribed inputs.
White noise is not such an input: a white-noise-driven model requires a
stochastic delay equation and a well-defined stochastic convolution under
separate noise assumptions. An expected quadratic-energy estimate must include
the It\^o correction. For trajectories with the same additive noise coefficient,
coupling them to the same noise realization cancels that term in their difference.

The following functional measures field deviation, memory mismatch and recent
external-input energy; a decay bound requires aggregate dissipativity.

\begin{definition}[Coherence functional]
Take Hilbert memory spaces \(M_i\), and write \(c,m_i,y_\ell\) for
deviations from an exact reference solution, including a fixed stationary
operating point. A moving equilibrium adds a tracking residual as forcing and
is excluded here. In the deviation equation, \(J_i\) means the source
difference between the two full trajectories, with their histories. Define
\begin{equation}
C(t)
=
\|c_t\|_{H}^{2}
+
\sum_{i=1}^{N} a_i \|m_i(t)-\Gamma_i c_t\|^2
+
\sum_{\ell=1}^{N_\tau} b_\ell \int_{t-\tau_\ell}^{t} e^{-2\alpha(t-s)}\,\|y_\ell(s)\|^2\,\mathrm ds.
\label{eq:coherence-functional}
\end{equation}
Here \(a_i,b_\ell>0\), \(\alpha>0\) is the discount rate, and
\(\Gamma_i:H\to M_i\) is a bounded linear comparison map. The terms measure
present field deviation, memory--field mismatch and discounted energy of the
external \(y_\ell\) histories. There is no norm of the full internal delay
history \(\zeta_t\) in \(C\); in general, \(C\) does not control the norm
in \(\mathcal X_\tau\).
\end{definition}

\begin{remark}[Bounded-coherence inequality]
For sufficiently regular trajectories in a bounded neighborhood, let
\(e_i=m_i-\Gamma_i c\) and assume \(\lambda_0>0\), \(d_\ell\ge0\) satisfy
\[
\begin{aligned}
&2\Big\langle c,Lc+\sum_iJ_i+\sum_\ell E_\ell
y_\ell(t-\tau_\ell)\Big\rangle\\
&\quad+2\sum_i a_i\langle e_i,\dot m_i-\Gamma_i\dot c\rangle\\
&\le-\lambda_0\Big(\|c\|_H^2+\sum_i a_i\|e_i\|^2\Big)
+\sum_\ell d_\ell\|y_\ell(t-\tau_\ell)\|^2 .
\end{aligned}
\]
This aggregate inequality is an additional hypothesis on the full source
differences, including their internal delay dependence; neither well-posedness
nor stability alone establishes it. If, in addition,
\(b_\ell e^{-2\alpha\tau_\ell}\ge d_\ell\) for every delay channel, then
\begin{equation}
\frac{\mathrm d}{\mathrm dt}C(t)
\le
-\lambda C(t)
+
\sum_{\ell=1}^{N_\tau}b_\ell\|y_\ell(t)\|^2,
\qquad \lambda=\min\{\lambda_0,2\alpha\}>0.
\label{eq:coherence-bound}
\end{equation}
The estimate applies only while the stated regularity and aggregate
dissipativity hypotheses hold.
\end{remark}

\emph{Derivation.}
Leibniz differentiation gives, for each history term,
\[
\frac{\mathrm d}{\mathrm dt}\int_{t-\tau_\ell}^{t}
e^{-2\alpha(t-s)}\|y_\ell(s)\|^2\,\mathrm ds
=\|y_\ell(t)\|^2-e^{-2\alpha\tau_\ell}
\|y_\ell(t-\tau_\ell)\|^2
-2\alpha\int_{t-\tau_\ell}^{t}e^{-2\alpha(t-s)}
\|y_\ell(s)\|^2\,\mathrm ds .
\]
Adding these identities, weighted by \(b_\ell\), to the aggregate
dissipativity bound and using
\(b_\ell e^{-2\alpha\tau_\ell}\ge d_\ell\) absorbs every delayed boundary
term and yields \eqref{eq:coherence-bound}. The calculation uses the
weighted-history differentiation underlying Lyapunov--Krasovskii
estimates~\citep{hale1993,fridman2014}; the aggregate dissipativity premise
remains a separate assumption.

\emph{Interpretation.}
The estimate bounds the selected deviations under its hypotheses; it is not
a general delayed stability theorem. For example, the asymptotically stable
scalar equation \(\dot x(t)=-x(t)+0.4x(t-\tau)\), \(\tau>0\), with no
auxiliary memories or external channels has \(C=x^2\). A continuous initial
history with \(x(0)=\varepsilon\ne0\) and \(x(-\tau)=3\varepsilon\) gives
\[
\dot C(0^+)=2\varepsilon(-\varepsilon+1.2\varepsilon)
=0.4\varepsilon^2>0.
\]
Stability permits this initial growth; the aggregate dissipation hypothesis
fails for this preparation. A Lyapunov estimate for the full delayed state
requires a functional controlling its internal histories and a separate
dissipation bound.
The estimate establishes neither controllability nor repair policy or cost.
Empirical lag, deficit-area and recovery-fraction readouts summarize
perturbation outcomes and do not test this differential inequality.

\subsection{Readout-defined operating boundaries}
\label{si:boundary-susceptibility}

Criteria on scalar or vector readouts define regimes of the full state; under
factorization they also define regimes on retained coordinates. For a nested
pair, the branch and path assumptions below govern boundary encounters.
Stationary susceptibility adds the closure and stability hypotheses of
Section~\ref{si:reduced-state-equation}.

\subsubsection{Local regimes and boundary crossings}
\label{si:local-readout-regimes}

For nested workability and exactness readouts \(P_f,Q_f\), the question is
when a loading path has separated regime crossings. Here \(w=P_f\) and
\(r=Q_f/P_f\) where \(P_f>0\); nesting alone does not order the crossings.

\begin{definition}[Admissible loading path]
Fix one admissible family \(f\) and let \(U_{\mathrm{adm}}\) be
a local control manifold on which the regular operating branch
\(u\mapsto z_\ast(u)\) exists smoothly. Call a \(C^1\) path
\(\gamma:[0,T]\to U_{\mathrm{adm}}\) an \emph{admissible loading path}
when it keeps the same instance class and the same outer/inner observables fixed, and
when its readout sequence has no re-entry: after leaving \(r=1\) it never
returns to \(r=1\), after reaching \(r=0\) it never returns to \(r>0\), and
after reaching \(w=0\) it remains failed. These conditions define monotone
loading but do not impose the relative locations of the first-hit thresholds.
The path must follow the specified intervention and be fixed before crossing
data are read; selection after observing the transition order is inadmissible.
\end{definition}

On the operating branch, these coordinates are
\(w(u)=P_f(z_\ast(u))\) and
\(r(u)=Q_f(z_\ast(u))/P_f(z_\ast(u))\) for \(P_f(z_\ast(u))>0\).
The four regimes and their boundaries are defined together below.

\begin{proposition}[Local boundary surfaces]
Define the four readout strata
\[
S_{\mathrm E}:=\{u\in U_{\mathrm{adm}}:\ w(u)>0,\ r(u)=1\},\qquad
S_{\mathrm{PE}}:=\{u\in U_{\mathrm{adm}}:\ w(u)>0,\ 0<r(u)<1\},
\]
\[
S_{\mathrm W}:=\{u\in U_{\mathrm{adm}}:\ w(u)>0,\ r(u)=0\},
\qquad
S_{\mathrm F}:=\{u\in U_{\mathrm{adm}}:\ w(u)=0\},
\]
the exact, partially exact, workable-but-inexact, and failed strata.
An overline on a set denotes its closure, including its limit points.
Define each transition locus as the common relative boundary of the two adjacent strata,
\[
\Sigma_{EW}:=\overline{S_{\mathrm E}}\cap\overline{S_{\mathrm{PE}}},\qquad
\Sigma_{WR}:=\overline{S_{\mathrm{PE}}}\cap\overline{S_{\mathrm W}},\qquad
\Sigma_{RO}:=\overline{S_{\mathrm E}\cup S_{\mathrm{PE}}\cup S_{\mathrm W}}\cap\overline{S_{\mathrm F}}.
\]
The closures for \(\Sigma_{EW}\) and \(\Sigma_{WR}\) are taken relative to
\(U_+=\{w>0\}\), and the closure for \(\Sigma_{RO}\) relative to
\(U_{\mathrm{adm}}\). Assume each locus is locally a \(C^1\) codimension-one
hypersurface in this relative topology and the loading path meets it transversally at nonzero rate.
At each encounter, additionally require the path to pass between the two
defining strata on nonempty adjoining intervals. This is an actual change of
regime, not merely an intersection with their closures; for \(\Sigma_{RO}\),
the two sides are workable and failed. These assumptions can fail at plateau
edges, intersections or corners. The set identities alone establish neither
smoothness, actual regime changes nor encounter order.
The parameters \(s_{EW}\), \(s_{WR}\) and \(s_{RO}\) are the first encounters
with the corresponding transition loci along the declared path; an
unencountered locus has first-hit parameter \(+\infty\).
The four-step statement requires all three encounters within the observation
interval. Along a
continuous admissible monotone path starting in the exact stratum, the
following order holds only when the first-hit parameters satisfy
\(s_{EW}<s_{WR}<s_{RO}\):
\[
\text{exact stratum}
\to
\text{partially exact region}
\to
\text{workable-but-inexact stratum}
\to
\text{failed stratum}.
\]
\end{proposition}

Nesting gives \(0\le r\le1\) on \(U_+\); failure is specified by
\(w=0\), where \(r\) is undefined. For discontinuous readouts, the declared
equality convention assigns threshold points. Smooth boundaries and transverse
encounters that change the adjacent strata are additional assumptions.

The no-re-entry condition, actual crossings and ordered finite encounters
\(s_{EW}<s_{WR}<s_{RO}\) give the displayed sequence, with nonempty
intervals in each adjacent regime. Coincident thresholds can skip a regime;
they are non-generic only if the threshold conditions vary independently.
Regularity and transversality isolate encounters but do not order them.
Thus the sequence is specific to the declared family and path
(main-text Figure~\mainref{fig:reduced-frontier}).

\begin{criterion}[Boundary-test selection principle]
For a fixed instance class, select controls whose projected responses
distinguish exactness loss from workability loss while
preserving the stated no-re-entry conditions along the same loading path.
\end{criterion}

Thresholds, response curves and instance locations require empirical calibration.

\subsection{Susceptibility and comparison of macroscopic readouts}
\label{si:susceptibility-comparison}

The question is when a finite-window field response admits a space--time
kernel or stationary limit and how it projects to an observable. The conditions
below govern these steps; \(h\) denotes the perturbing source history.

\begin{definition}[Susceptibility operator and kernel]
For a declared source space \(\mathcal U\) and field space \(\mathcal H\),
the primary response is the bounded derivative
\(\mathcal R_t=\mathrm D_h\langle c(t)\rangle:\mathcal U\to\mathcal H\),
when this Fr\'echet derivative exists. Angle brackets denote ensemble averages;
\(\alpha,\beta\) index field channels.
If this operator admits a space--time kernel, its channel components are
written
\[
\chi_{\alpha\beta}(\mathbf x,\mathbf y;t,t')
=
\frac{\delta\langle c_\alpha(\mathbf x,t)\rangle}
     {\delta h_\beta(\mathbf y,t')}.
\]
This notation denotes a weak or distributional kernel unless additional
spatial regularity gives pointwise values. In particular, point evaluation
is not a bounded readout on \(L^2\), and a bounded operator on that space
need not have an ordinary integral kernel. For \(\dot c=-c+h\), the impulse
response is \(e^{-(t-t')}I\) for \(t\ge t'\), and zero earlier; its spatial
kernel is a delta distribution.
Bounded linear readouts act directly on \(\mathcal R_t\); an \(L^2\) readout
weight gives the corresponding inner product. Pointwise readouts require a
stronger state space or a separately justified regularization.
\end{definition}

For a deterministic finite graph \(G\), or a pathwise derivative with the
noise realization held fixed, define the field and readout responses by
\[
\widehat\chi^G_{\alpha\beta}(x,y;t,t')
=
\frac{\partial c_\alpha(x,t)}{\partial h_\beta(y,t')},
\qquad
\widehat\chi^{\rm red}_{aj}
=
\frac{\partial O_a}{\partial\theta_j}.
\]
Here \(x,y\in V_G\) are vertices of \(G=(V_G,E_G)\), \(t,t'\) are
discrete times, and \(\alpha,\beta\) index the declared finite channel basis;
\(c_\alpha(x,t)\) is the field-channel value. The amplitude \(h_\beta(y,t')\) multiplies a unit source direction at the
specified vertex, channel and time. \(O_a\) is a smooth readout component,
and \(\theta_j\) is the amplitude along a specified source-control direction. When the readout depends
only on the field coordinates represented in this operator, applying the source
directions and readout differential gives \(\widehat\chi^{\rm red}\).
Readouts that also use memory or trajectory records require the corresponding
augmented-state response. This finite estimate becomes a continuum kernel only under graph, time and
channel refinement, and a stationary susceptibility only under the stable
limit below. For mean responses, differentiate the ensemble mean; averaging
pathwise derivatives is valid only when derivative and expectation interchange.
Solution, projection and readout differentiability are additional to Lipschitz
well-posedness. Nonsmooth points require finite contrasts or specified one-sided
responses.

For fixed preparation, including field, memories and histories,
let \(s_T\) be the full terminal state and \(z_T=G_f(s_T)\) its retained
description. If the readout has a differentiable representation
\(O_T(s_T)=\bar O_T(z_T)\), the chain rule gives
\[
\mathrm D_sO_T(s_T)=\mathrm D_z\bar O_T(z_T)\,\mathrm D_sG_f(s_T).
\]
Here \(\mathrm D\) denotes the derivative: a Jacobian in finite dimensions
or a Fr\'echet derivative on the specified function space. The bar labels
the readout as a function of \(z_T\), not an ensemble average. Read from
right to left, the derivatives map a full-state perturbation to a retained-state
change and then to a readout change.
If the graph contains the full required augmented state and is itself
retained, \(G_f\) is the identity and \(\mathrm D_sG_f=I\); otherwise keep this derivative in the
response chain (Section~\ref{si:falsifiable-retained-state}).

For a graph used as the retained state or a verified closed retained
transition, let \(A,B\) be the time-independent retained transition and
source maps. If \(\rho(A)<1\), a persistent tangent source gives the
resolvent \(\chi_\ast^G\) derived in Section~\ref{si:reduced-state-equation}.
A one-time source at \(t'\) instead gives the retarded Green function
\(A^{t-t'}B\) for \(t\ge t'\) and zero response earlier. The discrete
resolvent acts on retained coordinates; a full-state response must first be
mapped to retained coordinates by \(\mathrm D_sG_f\). Equality with the continuous-time branch
susceptibility in Eq.~\eqref{eq:supp-linear-branch-response} requires exact
zero-order-hold discretization (Section~\ref{si:reduced-state-equation})
or a demonstrated graph-and-time refinement limit.

Section~\ref{si:empirical-protocols} gives response checks for code-repair
constructions using endpoint information, deterministic replay and readouts
from fresh model calls.

The participation ratio \(r_{\rm PR}\) defined in
Section~\ref{si:minimal-realization} measures concentration in the response
matrix's squared singular values. It equals matrix rank when the nonzero
singular values are equal, and approaches one when a single squared singular
value dominates their sum. This spectral diagnostic does not determine the
minimal state needed for an endpoint readout or a common dimensionality across
realizations.

The continuum comparison below preserves source units and controls accumulated
response. Let \(h_n=\Delta t_n\to0\), \(\rho(A_n)<1\),
and let \(B_n\) be a per-unit-time source operator, not an integrated source
step. Interpolation \(I_n\) maps graph to continuum states; \(P_n^U\) maps
the declared source space to graph inputs. The complete linearized continuum
generator \(\mathcal L\) includes source, memory and feedback derivatives;
it differs from the medium generator \(L\) in
Eq.~\eqref{eq:supp-universal-interface} unless those contributions vanish.
Assume it generates a strongly continuous semigroup, \(0\) is in its
resolvent set and the corresponding continuum source operator \(\mathcal B\),
distinct from the graph source map \(B\), is bounded.
On the same declared spaces, sufficient joint consistency and stability
conditions, in source-space operator norm, are convergence for each \(t\ge0\)
and a bound uniform in all \(n,k\):
\[
I_nA_n^{\lfloor t/h_n\rfloor}B_nP_n^U\to e^{t\mathcal L}\mathcal B,
\qquad
\|I_nA_n^kB_nP_n^U\|\le Ce^{-\omega kh_n},
\qquad C,\omega>0.
\]
The identity
\[
h_n\sum_{k\ge0}I_nA_n^kB_nP_n^U
=\int_0^\infty I_nA_n^{\lfloor t/h_n\rfloor}B_nP_n^U\,\mathrm dt
\]
and dominated convergence then give
\[
I_n(I-A_n)^{-1}h_nB_nP_n^U\to(-\mathcal L)^{-1}\mathcal B.
\]
Joint consistency controls sources propagated through hidden coordinates;
separate projected consistency of \(A_n,B_n\) is insufficient. These are
refinement hypotheses, not a continuum-limit conclusion for a reported graph.

\begin{proposition}[Matched susceptibility and readout response]
For matched perturbation channels and the same fixed linear readout, equal
mean-field susceptibilities give equal first-order mean endpoint responses.
The same statement holds for smooth readouts linearized about a deterministic
reference with the same readout differential. Equality of
the susceptibility alone does not imply agreement under different readout
maps.
\end{proposition}

\emph{Derivation.}
Let \(K_\alpha(\mathbf x,t)\) be the fixed readout weighting kernel, with
summation over repeated channel indices. For the linear coarse observable
\(O[c](t)=\int K_\alpha(\mathbf x,t)c_\alpha(\mathbf x,t)\,d\mathbf x\),
\[
\delta\langle O(t)\rangle
=
\int K_\alpha(\mathbf x,t)
\chi_{\alpha\beta}(\mathbf x,\mathbf y;t,t')
\delta h_\beta(\mathbf y,t')\,d\mathbf x\,d\mathbf y\,dt'.
\]
The conclusion follows by contraction with the common readout kernel.
A smooth readout linearized about a deterministic reference has the same
first-order form with \(K=\mathrm DO|_{c_\ast}\). For nonlinear observables
of a fluctuating field, the ensemble response requires a separately justified
closure or the joint state and response statistics. The
statement is conditional on matched channels, coarse-graining and readout
semantics; none of the empirical realizations is claimed to share a measured
susceptibility, response spectrum, dimension or boundary geometry.

\section{Physical measurement methods and additional results}
\label{si:physical-realizations}

For the models and observables in main-text Section~\mainref{iclr:physics},
the methods below compare simulated delayed consensus with analytic response
predictions, then examine distinct measures of flock order. They specify
estimators, uncertainty and the interpretation of each comparison.

\subsection{Driven delayed consensus network}
\label{si:consensus-network}

Simulated network trajectories, one per delay and seed, provide the collective
mean signal and spatial variance. The mean's amplitude and phase are compared with
the harmonic response of the delayed mean equation; separate drive/noise
sweeps test the workability and exactness criteria.

\paragraph{Numerical trajectory and readout.}
The delayed network and harmonic target are defined in main-text Eqs.~\mainEqref{eq:supp-driven-consensus}--\mainEqref{eq:consensus-harmonic-targets}.
The independent unit-intensity white noises satisfy
$\langle\xi_i(t)\rangle=0$ and
$\langle\xi_i(t)\xi_j(t')\rangle=\delta_{ij}\delta(t-t')$, with Kronecker
and Dirac deltas, respectively; angle brackets denote ensemble averages.
The documented Euler--Maruyama scheme evaluates the drift at the beginning
of each step and adds independent Gaussian increments of standard deviation
\(\sigma\sqrt{\Delta t}\) at each site. Linear interpolation of stored
states approximates \(c(t-\tau)\). The documented initialization is
\(c_i(t)=0.01g_i\) for \(-\tau\le t\le0\), with independent standard-normal
\(g_i\). A Mersenne Twister generator is reset for each trajectory; its stream
supplies the initial values and subsequent noise.

The calibration tests local harmonic-response agreement in a stable weak-delay
regime. The \(13\)-delay sweep spans \(\tau\in[0.43,0.46]\), using window
\([300,400]\), \(\Omega=0.35\) and \(\Delta t=0.02\). The window length
\(T_{\rm win}=100\) gives the estimator
\((2\Delta t/T_{\rm win})\sum_{n=15000}^{20000}m(n\Delta t)e^{-i\Omega n\Delta t}\),
with \(5{,}001\) samples. Amplitude and phase lag are averaged over seeds
\(2026043201\)--\(2026043300\), reused at each delay, then compared with the
harmonic target. The records specify linear interpolation, whereas their cited
simulator revision rounds delay to a grid step. A matching executed-source record and
runtime environment are absent: the stated scheme is a documented reconstruction
with incomplete execution provenance.

Coverage is limited to \(\Omega\tau\in[0.1505,0.1610]\), with \(a>|K|\)
throughout. Workability remains saturated; the drive/noise sweeps below test
nested criteria. The calibration includes neither matched preparations with
equal current fields and different histories nor comparative history-compression
tests. Observation accumulators give terminal readouts of an enlarged state
(Section~\ref{si:observation-state}); they do not establish autonomous
three-dimensional dynamics. The calculation retains delay history and assumes
neither a slow manifold nor a time-summed resolvent.

Uniform forcing and mean readout select the consensus Fourier mode.
Circulant unit-row-sum coupling preserves that mode and its zero-sum
complement, whose \(N-1\) transverse modes have zero uniform-source response
and mean-readout sensitivity to numerical precision. The deterministic spatial
response is therefore rank one, while its evolution still requires delay history.
The following noise calculation explains what the additional variance readout
measures.

\paragraph{Projection of the noise.}
Normal modes separate collective noise in the lock-in estimator from
transverse noise measured by \(V\). The variance scale
$\mathcal O(\sigma^2|\chi|^2/(N T_{\rm win}))$ below concerns complex
lock-in fluctuations, not amplitude or phase error.

\emph{(i) Normal-mode decomposition.} Write $c_i(t)=m(t)+\delta c_i(t)$ with
$\sum_i\delta c_i=0$. Here $\delta c_i$ denotes the spatial fluctuation about
the instantaneous mean, rather than a source-response perturbation. The symmetric ring averaging matrix preserves the
consensus and zero-sum transverse subspaces. Main-text Eq.~\mainEqref{eq:supp-consensus-mean-eq} gives the exact mean projection,
including its noise $\eta_m$.

\emph{(ii) Collective-noise intensity $\sigma^2/N$.} Independence of the
$\xi_i$ gives
\begin{equation}
\langle\eta_m(t)\rangle=0,\qquad
\langle\eta_m(t)\,\eta_m(t')\rangle
=\frac{\sigma^2}{N^2}\sum_{i,j}\langle\xi_i(t)\xi_j(t')\rangle
=\frac{\sigma^2}{N}\,\delta(t-t'),
\label{eq:supp-consensus-etam}
\end{equation}
Thus spatial averaging reduces the noise amplitude from $\sigma$ to
$\sigma/\sqrt N$.

\emph{(iii) Lock-in as a finite-window matched-filter estimator.}
Main-text Eq.~\mainEqref{eq:supp-consensus-mean-eq} is linear, so
$m=m_{\rm det}+m_{\rm noise}$. After transients, the periodic component is
\(m_{\rm det}(t)=\operatorname{Re}\{F\chi e^{i\Omega t}\}\), with
\(\chi=\chi(\Omega,\tau)\); \(\overline\chi\) denotes its complex conjugate.
On a rectangular window
\([t_0,t_0+T_{\rm win}]\), its lock-in is
\[
\widehat Z_{\rm det}
=F\chi+F\overline\chi\,e^{-2i\Omega(t_0+T_{\rm win}/2)}
\frac{\sin(\Omega T_{\rm win})}{\Omega T_{\rm win}}.
\]
The second term is the contribution from the conjugate frequency of the real
signal. It vanishes when \(\Omega T_{\rm win}\in\pi\mathbb Z\), and decays
as the window grows. Thus \(F\chi\), with amplitude
\(\mathcal A_{\rm th}=F|\chi|\) and phase lag \(\phi_{\rm th}=|\arg\chi|\),
is the harmonic-response target rather than the exact value of every finite
window estimator. The recorded experiment uses the discrete rectangular sum
\[
\widehat Z_{\rm grid}(\tau)=\frac{2\Delta t}{T_{\rm win}}
\sum_{j=0}^{5000}\operatorname{Re}\{F\chi(\Omega,\tau)e^{i\Omega t_j}\}
e^{-i\Omega t_j},\qquad t_j=300+j\Delta t,\quad\Delta t=0.02.
\]
Set \(\mathcal A_{\rm grid}=|\widehat Z_{\rm grid}|\) and
\(\phi_{\rm grid}=|\arg\widehat Z_{\rm grid}|\). The finite window accounts
for most of the phase discrepancy across the 13 delays:
\(\phi_{\rm grid}-\phi_{\rm th}=0.674400\)--\(0.676273^\circ\), whereas
the recorded seed-mean phase minus \(\phi_{\rm grid}\) is
\(0.058675\)--\(0.058919^\circ\). These signed terms add.
The steady periodic solution of the documented Euler--Maruyama scheme adds
\(0.054115\)--\(0.054358^\circ\) to \(\phi_{\rm grid}\) on the same window;
the residual seed-mean difference is approximately \(0.00456^\circ\).
This analytic discretization correction assumes that scheme; it does not test
time-step convergence.

The amplitude corrections partially cancel: the
grid-minus-harmonic and seed-mean-minus-grid differences, each divided by \(\mathcal A_{\rm th}\),
are \(-0.003809\) to \(-0.003710\) and \(+0.001292\) to \(+0.001299\).
These are analytic reanalyses of the stated estimator
and stored means. The amplitude remainder combines discretization, residual
transients and stochastic averaging, whose contributions are unresolved.

Stochastic uncertainty is a separate contribution. The stochastic part $m_{\rm noise}$ is the output of the stable filter
$H(\omega)=(a-Ke^{-i\omega\tau}+i\omega)^{-1}$ (the continuation of $\chi$ to
general $\omega$) driven by white noise of intensity $\sigma^2/N$; by
Wiener--Khinchin its power spectral density is the smooth bounded
$S_m(\omega)=(\sigma^2/N)|H(\omega)|^2$. The lock-in integral is a Fourier
estimator with bandwidth $\sim1/T_{\mathrm{win}}$.

For stationary noise with integrable autocovariance
\(R_m(v)=(2\pi)^{-1}\int S_m(\omega)e^{i\omega v}\,d\omega\), the complex
variance means \(\operatorname{Var}_{\mathbb C}Z=\mathbb E|Z-\mathbb EZ|^2\).
The rectangular-window integral has variance
\(4T_{\mathrm{win}}^{-2}\int_{-T_{\mathrm{win}}}^{T_{\mathrm{win}}}
(T_{\mathrm{win}}-|v|)R_m(v)e^{-i\Omega v}\,dv\), hence
\begin{equation}
\operatorname{Var}_{\mathbb C}(\hat Z_{\rm noise})
=\frac{4\sigma^2|\chi(\Omega,\tau)|^2}{N T_{\mathrm{win}}}
 +o(T_{\mathrm{win}}^{-1}).
\label{eq:supp-consensus-Zvar}
\end{equation}
The remainder \(o(T_{\rm win}^{-1})\) decays faster than \(1/T_{\rm win}\).
This asymptotic expression concerns the continuous rectangular-window estimator;
the recorded discrete sum has its corresponding discrete covariance sum.
Spatial and temporal averaging suppress stochastic variance. The total
discrepancies reported in main-text Table~\mainref{tab:phys-anm-compact}
also include demodulation leakage, transients and discretization.

\emph{(iv) Transverse modes as the retained variance $V$.} Subtracting
main-text Eq.~\mainEqref{eq:supp-consensus-mean-eq} from
main-text Eq.~\mainEqref{eq:supp-driven-consensus}
removes the uniform drive and consensus-mode dynamics exactly, leaving the
transverse fluctuations forced by the transverse projection of the noise,
\begin{equation}
\dot{\delta c}_i(t)=-a\,\delta c_i(t)
+\frac{K}{|N_i|}\sum_{j\in N_i}\delta c_j(t-\tau)+\sigma\,\delta\xi_i(t),
\qquad
\delta\xi_i\equiv\xi_i-\frac1N\sum_k\xi_k.
\label{eq:supp-consensus-fluct-eq}
\end{equation}
On the transverse coupling eigenmodes $\{e_q\}_{q\ne0}$ (eigenvalues
$\kappa_q\in[-1,1)$; the consensus mode $\kappa_0=1$ is excluded), each amplitude
$\delta c_q$ is a stable scalar delay Ornstein--Uhlenbeck process whose
stationary variance is the spectral integral
\begin{equation}
\langle\delta c_q^2\rangle
=\frac{\sigma^2}{2\pi}\int_{-\infty}^{\infty}
\bigl|a-K\kappa_q e^{-i\omega\tau}+i\omega\bigr|^{-2}\,\mathrm d\omega
=:\sigma^2\,g_q(\tau),\qquad g_q(\tau)>0,
\label{eq:supp-consensus-modevar}
\end{equation}
where $g_q(0)=1/(2a_{\rm eff,q})$ is the memoryless Ornstein--Uhlenbeck
factor and $a_{\rm eff,q}=a-K\kappa_q>0$ for the stated parameters.
By Parseval, the stationary expectation of the spatial variance is
\begin{equation}
\langle V(t)\rangle=\frac1N\sum_{i=1}^N\langle(c_i-m)^2\rangle
=\frac1N\sum_{q\ne0}\langle\delta c_q^2\rangle
=\frac{\sigma^2}{N}\sum_{q\ne0}g_q(\tau)=\mathcal O(\sigma^2).
\label{eq:supp-consensus-V}
\end{equation}
The reported $V$ estimates this stationary expectation by averaging the
instantaneous spatial variance over the finite observation window.
Thus transverse noise gives $\langle V(t)\rangle=\mathcal O(\sigma^2)$,
while collective noise contributes at order
$\sigma^2|\chi|^2/(N T_{\rm win})$ to complex lock-in variance
(Eq.~\eqref{eq:supp-consensus-Zvar}), with subleading amplitude/phase effects.
The analytic amplitude is
$\mathcal A_{\rm th}=F|\chi(\Omega,\tau)|$. The stationary variance relative
to its square is
\begin{equation}
\frac{\langle V(t)\rangle}{\mathcal A_{\rm th}^2}
=\overline{g}(\tau)\,\frac{\sigma^2}{F^2|\chi(\Omega,\tau)|^2},
\qquad
\overline{g}(\tau)=\frac1N\sum_{q\ne0}g_q(\tau)=\mathcal O(1),
\end{equation}
the mode-averaged factor of Eq.~\eqref{eq:supp-consensus-V}. In the stated
simulation units, the noise-to-drive amplitude ratio $\sigma/F=0.06$ sets the
prefactor $\sigma^2/F^2\approx3.6\times10^{-3}$,
and the measured value across the delay window is $V/\mathcal A^2<10^{-3}$, i.e.\
$\mathcal A^2/(\mathcal A^2+V)\ge0.999$.

\paragraph{Meaning of the spatial summary.}
The concentration ratio above uses peak amplitude squared rather than its
time average. A sinusoid has mean square
$\mathcal A^2/2$, so this is not a time-averaged energy fraction.
A participation ratio follows from nonnegative weights
\[
E_0=\mathcal A^2,\qquad \sum_{k=1}^{N-1}E_k=V,\qquad
r_{\rm eff}=\frac{(\sum_{k\ge0}E_k)^2}{\sum_{k\ge0}E_k^2}.
\]
Writing \(C=\mathcal A^2/(\mathcal A^2+V)\), concentration of all transverse
weight in one mode and equal distribution over \(N-1\) modes give
\[
\frac{1}{C^2+(1-C)^2}\leq r_{\rm eff}\leq
\frac{1}{C^2+(1-C)^2/(N-1)}.
\]
At the least-concentrated delay, \(C\approx0.9990\) and \(N=64\). Using
the unrounded recorded concentration gives the outward-rounded bounds
\(1.00199\leq r_{\rm eff}\leq1.00200\), summarized as \(r_{\rm eff}\approx1.002\). These bounds are derived from recorded amplitude and variance,
not measured covariance spectra at all 13 delays. Their lower endpoint is
not a lower bound for every delay. Let \(N_{\rm samp}\) be the number of
observation times. To express these same weights through
eigenvalues \(\mu_k\) of the sampled second-moment matrix
\(N_{\rm samp}^{-1}\sum_j\delta c(t_j)\delta c(t_j)^\top\), use
\(E_k=\alpha\mu_k/N\), where
\(\alpha=N_{\rm samp}\Delta t/T_{\rm win}=5001/5000\).
This factor matches the rectangular-quadrature definition of \(V\);
for seed-averaged records, average these matrices over the same seeds.
Unscaled second-moment weights have a different normalization. The uniform drive
excites only the consensus spatial mode. Its dominance is a spatial-response
summary, not the minimal dimension of an autonomous three-dimensional state.

\paragraph{Drive and noise sweeps of the operating criteria.}
The sweeps compare an admissible signal with the stricter amplitude-and-phase
criterion as drive decreases or noise increases. At \(\tau=0.43\), main-text
Eq.~\mainEqref{eq:consensus-operating-criteria} defines \(P_f\) by finite-valued
integration through the horizon, a nonempty measurement window, amplitude
\(\mathcal A\ge0.16\) and variance \(V\le0.05\); \(Q_f\) additionally
requires \(\mathcal A\ge0.22\) and phase lag \(\phi\le35^\circ\).
Decreasing drive reduces the coherent signal at fixed noise \(\sigma=0.015\);
increasing noise raises transverse variance and perturbs phase at fixed drive
\(F=0.25\). Both retrospective sweeps reuse seeds
\(2026042801\)--\(2026042850\) at every control value with fixed thresholds;
their records bind the simulator source for the stated update and initialization.

\begin{figure}[!htbp]
\centering
\includegraphics[width=\linewidth]{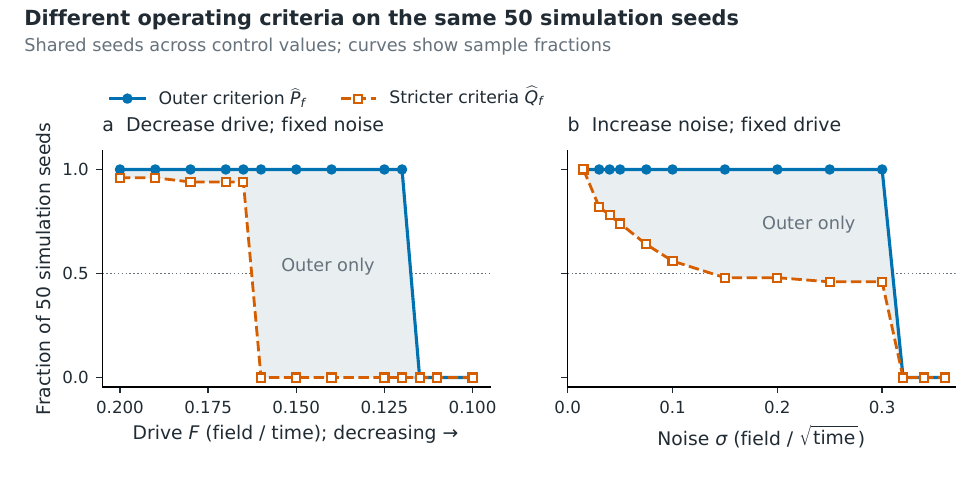}
\caption{\textbf{Sample support fractions for nested consensus criteria.}
Fractions \(\widehat P_f\) (circles) and \(\widehat Q_f\) (squares) of the same
50 simulation seeds meeting the outer and stricter nested criteria,
respectively, at all 13 recorded points on each axis.
\textbf{a}, Drive decreases at fixed noise \(\sigma=0.015\).
\textbf{b}, Noise increases at fixed drive \(F=0.25\). Both use delay
\(\tau=0.43\). Shading between curves gives the fraction passing only the
outer criterion at each sampled point. Lines interpolate points for display;
the dotted line marks half support. The outer rule requires finite-valued
integration through the horizon, a nonempty measurement window, lock-in
amplitude at least 0.16 and window-averaged spatial variance at most 0.05; the inner rule
also requires amplitude at least 0.22 and phase lag at most \(35^\circ\).
Nesting imposes \(Q_f\le P_f\); the measured fractions quantify the sample
gap. Reused noise realizations correlate the points. In panel b the early
half-support crossing is consistent with finite-sample phase fluctuations;
it does not identify a population phase-loss boundary. Axes use the simulation's field
and time units. Complete fractions, amplitude and variance means remain in
the authors' records.}
\label{fig:consensus-criteria}
\end{figure}

To quantify criterion separation, average the binary readouts over \(50\)
seeds to obtain support fractions \(\widehat P_f\) and \(\widehat Q_f\).
Linear interpolation gives half-support point estimates (fraction equal to
one half): under decreasing drive, \(F\approx0.163\) for \(Q_f\) and
\(F\approx0.118\) for \(P_f\); under increasing noise, the corresponding
estimates are \(\sigma\approx0.138\) and \(0.310\).
Figure~\ref{fig:consensus-criteria} shows both fractions on the complete
recorded grids. Across \(26\) control values and \(50\) seeds each, \(442\)
of the \(1{,}300\) trajectories are workable but inexact, \(Q_f=0<P_f\),
with no \(Q_f>P_f\) violations. Nesting guarantees weak order on both paths;
measurements quantify interpolated sample separation and workable-but-inexact
frequency. Drive crossings mainly reflect amplitude thresholds.

On the noise axis, phase failure accounts for the workable-but-inexact
trajectories: 179 of 650 satisfy \(P_f=1,Q_f=0\), all passing the extra
amplitude criterion. All outer failures exceed the variance threshold:
another 150 have \(P_f=0\), forcing \(Q_f=0\); 68 of these have phase at or
below \(35^\circ\). Phase failure alone therefore cannot explain all 329
cases with \(Q_f=0\).

Across \(20{,}000\) paired-seed bootstrap resamples, central 95\% ranges for
the interpolated \(Q_f\) half-support statistic are \([0.1625,0.1629]\)
on the drive axis and \([0.075,0.303]\) on the noise axis. These describe
variation for this grid, window and empirical seed distribution, without
established population-threshold coverage. The resampled crossing can reflect
variance failures even when phase support exceeds one half; the statistic
identifies neither a separate population phase boundary nor an outer threshold
between controls whose outcomes all change together.

\paragraph{Sampling dependence and population phase probability.}
Reusing the same 50 random-number streams makes the noise controls strongly
dependent. At
\(\sigma=0.10,0.15,0.20,0.25,0.30\), respectively, the phase-pass counts
are \(28,24,24,23,23\); all 50 runs pass the outer criterion and the
extra amplitude criterion. The corresponding pointwise 95\% Wilson intervals
at \(0.10\), \(0.15\) and \(0.30\) are
\([0.423,0.688]\), \([0.348,0.615]\) and \([0.330,0.596]\).
They are not simultaneous intervals. Across all pairs of noise controls,
the Pearson correlations of the 50 phase values exceed \(0.99969\);
the phase-pass vectors at \(0.15\) and \(0.20\) are identical.
Repeated sub-half fractions therefore do not supply independent replications.

Linearity explains this dependence and permits a population calculation.
After negligible initialization transients, write the numerical lock-in as
\(\widehat Z=Z_0+\sigma\zeta\), where \(Z_0\) is the deterministic
numerical response and \(\zeta\) is a zero-mean complex Gaussian variable
whose two real components have covariance fixed by the linear evolution and
window. A fixed seed reuses the same \(\zeta\) at every noise amplitude.
For the stated Euler step and linearly interpolated delay at \(\tau=0.43\),
the mean obeys
\[
m_{n+1}=0.98m_n+0.004(m_{n-21}+m_{n-22})
       +0.005\cos(0.007n)+\sigma\sqrt{0.02/64}\,\epsilon_n,
\]
where \(\epsilon_n\) are independent standard normal variables.
Applying the recorded lock-in sum gives
\(Z_0\approx0.280127897-0.195222004i\), with phase lag
\(34.872761^\circ\). Applying that sum to the continuous harmonic solution
instead gives \(34.818645^\circ\); these are different deterministic
references.

Let \(\theta=35^\circ\),
\(d=\operatorname{Re}Z_0\sin\theta+\operatorname{Im}Z_0\cos\theta\),
and let \(s^2\) be the variance of
\(\operatorname{Re}\zeta\sin\theta+\operatorname{Im}\zeta\cos\theta\).
The lower edge of the allowed phase sector is a linear Gaussian inequality.
The opposite edge has negligible tail probability over the recorded range,
so, with \(\Phi_{\rm N}\) the standard-normal cumulative distribution,
\[
\Pr\{\phi\le35^\circ\}\simeq
\Phi_{\rm N}\!\left(\frac{d}{\sigma s}\right),\qquad
d\approx0.000758257,\quad s\approx0.0241341.
\]
The covariance follows from the impulse response of the recurrence above,
without fitting the seed outcomes. This probability is approximately
\(0.583\) at \(\sigma=0.15\) and \(0.542\) at \(0.30\), remaining
above one half throughout the sampled range. The observed early crossing
at \(0.1375\) is consistent with finite-sample fluctuations about this
population response. This conclusion concerns the phase condition: the full
probabilities \(\Pr(Q_f=1)\) and \(\Pr(P_f=1)\) also include the
variance constraint. Their half-support points can remain ordered near the
variance-driven decline; the records do not establish the absence of any
population \(Q_f\)-before-\(P_f\) crossing.

\FloatBarrier
\subsection{Field-driven Vicsek flock}
\label{si:vicsek-flock}
\label{sec:vicsek-field-window}

Flock trajectories compare polarization \(P_f\), alignment among particles,
with \(Q_f\), the fraction within a specified angle of the imposed field.
Noise sweeps test whether these nonnested readouts decrease on the same scale.
Standard continuum descriptions retain density and vector
polarization~\citep{toner1995,marchetti2013,ramaswamy2010,chate2020}.

\paragraph{Sampling and angular-noise convention.}
Main-text Eqs.~\mainEqref{eq:vicsek-field}--\mainEqref{eq:flock-averaged-readouts}
specify the update, parameters and order of readout averaging. The independent
uniform angular noises have variance
\(\operatorname{Var}(\xi_i)=(2\pi)^2/12=\pi^2/3\).
Seeds \(21\)--\(24\) use \(500\) burn-in and \(300\) measurement steps.
The documented initialization draws positions uniformly in the square and
headings uniformly on \([-\pi,\pi)\). No
observation-time or fixed-density size limit is inferred. The loading
parameter \(b=\eta\) varies at fixed field, and the alignment tolerance
is compared over \(\{20^\circ,30^\circ,45^\circ\}\).

On \(\eta\in[0.15,0.70]\), measured \(Q_f\le P_f\);
\(r=Q_f/P_f\) for \(P_f>0\) compares alignment with polarization, not conditional
exactness. Order fails outside this interval: near disorder \(Q_f\) approaches
random-orientation baseline \(1/6\) while polarization has a finite-size
residual; at low nonzero noise \(Q_f\) can also exceed \(P_f\).
These nonnested readouts are excluded from workability--exactness regimes.
The wrapped-normal comparison below is a descriptive approximation to the
steady-state readouts, not a measured dynamical branch.

\paragraph{Separated sampled readout scales.}
Across \(\eta\in[0,0.85]\), polarization decreases \(1.000\to0.075\)
and alignment \(1.000\to0.178\). For seeds \(21\)--\(24\), first sampled
half-values occur at \(\eta=0.35\) (alignment) and \(0.60\) (polarization)
(main-text Fig.~\mainref{fig:physical-realizations}e). On \([0.15,0.70]\),
\(Q_f\le P_f\) but \(r\) is nonmonotone, excluding a global loading law
or four-regime trajectory.

At tolerances \(20^\circ,30^\circ,45^\circ\), all four seeds give alignment
half-values \(0.25,0.35,0.50\), each below \(0.60\). The \(45^\circ\)
gap is \(0.10\), two grid steps. The wrapped-normal summary places coincident
half-values at \(45.50^\circ\), close to reversal: separation is specific
to the tested thresholds. For \(\eta=0.35\) to \(0.55\), alignment within
\(30^\circ\) is below one half while polarization stays above it.
\paragraph{Wrapped-normal comparison.}
A wrapped-normal distribution gives a simple approximate relation between
polarization and the fraction aligned with the field. Let \(\sigma_\theta\)
be the standard deviation of the Gaussian angle before
wrapping. For a wrapped-normal heading distribution centered on the imposed field,
the population resultant is \(R=|\mathbb E e^{i\theta}|=e^{-\sigma_\theta^2/2}\).
The measured \(P_f\) instead averages the instantaneous finite-flock modulus;
it is not generally \(R\). Setting \(R=P_f\) defines an effective summary,
with inferred width (for \(0<P_f<1\))
\[
\sigma_\theta^{(P)}(\eta)=\sqrt{-2\log P_f(\eta)}.
\]
Let \(\Phi_{\mathrm N}\) be the standard-normal cumulative distribution function.
The corresponding probability for the principal wrapped angle to lie within
the tolerance is
\[
Q_f^{\rm WN}(\eta)
=
\sum_{k\in\mathbb Z}
\left[
\Phi_{\mathrm N}\!\left(\frac{\mathrm{tol}+2\pi k}{\sigma_\theta^{(P)}(\eta)}\right)
-\Phi_{\mathrm N}\!\left(\frac{-\mathrm{tol}+2\pi k}{\sigma_\theta^{(P)}(\eta)}\right)
\right],
\qquad \mathrm{tol}=30^\circ .
\]
Using the measured \(P_f\) values to infer \(\sigma_\theta^{(P)}\), with no
fit parameter, gives
\(\operatorname{RMSE}(Q_f,Q_f^{\rm WN})=0.0507\) across the 12 equally weighted
noise-point means on \(\eta\in[0.15,0.70]\). It overpredicts alignment at
11 of 12 points, with mean signed difference \(0.0443\) for \(Q_f^{\rm WN}-Q_f\)
and maximum absolute difference \(0.0717\). Finite-size effects,
collective-direction fluctuations and distributional mismatch can contribute
to these residuals; their separate contributions are unresolved. A perfectly
coherent flock can turn away from the field while its polarization stays one.
This is a descriptive comparison of the two curves, not a distributional
fit test, a measured dynamical state or a fluctuation--relaxation law.

\paragraph{Comparing normalized response profiles.}
An exploratory comparison tests whether flock and replayed AI response curves
agree beyond a shared normalization. It compares the flock's field-alignment
fraction with the correctly decoded field fraction from recorded AI events,
varying acceptance thresholds at fixed replayed graph scores. For each profile \(Q(b)\),
where \(b\) is noise or acceptance threshold, set \(S(b)=Q(b)/Q(0)\), with
\(Q(0)>0\). Let \(b_{1/2}\) be the first interpolated half-value and \(d<0\)
the secant slope of \(S\) through its bracketing samples. The dimensionless coordinate
\(\lambda=|d|(b_{1/2}-b)\) fixes an anchor and slope along
\(S=\tfrac12+\lambda\) by construction; common dynamics and a shared two-sided
tangent require separate evidence. Residuals also include sampling and
interpolation effects. Within windows \(|\lambda|\le W\) of half-width \(W=0.25\) or less,
the sole recorded AI loading level leaves its reference-line residual undefined
because reporting requires at least two samples in the window. At half-width 0.10, the two flock samples
define the secant, imposing zero residual. The control uses only recorded
profiles, with no new model calls or independent physical realizations, and
supplies no evidence of a universal response shape.

\FloatBarrier
\section{Methods for inverse tests under fixed inputs}
\label{si:inverse-experiments}
The inverse tests ask whether a candidate state preserves later responses and,
if rejected, which history distinctions it must retain. A known deterministic
program permits exact execution checks; a sampled model requires output-law
comparisons with uncertainty bounds. The methods for main-text
Section~\mainref{sec:worked-tool} therefore distinguish exact-program
reconstruction from bounded support on measured histories.
In both tests, two sensors \(S_1,S_2\) are assigned one-to-one to recipients
\(A,B\). The two association orientations exchange these assignments.
Time zero is the observation cut: requests precede it and returns follow it.
Pairs of request times or returned values are ordered by \((S_1,S_2)\),
independently of record order.
\subsection{Exact request--return matching}
\label{si:inverse-mwe-reconstruction}
The constructed task tests which pending-request information predicts a
recipient's later quantity. Histories associate sensors with recipients;
continuations supply values, arrival times and a query. The known program
matches returns to requests and reports a quantity and threshold indicator.
Reconstruction compares matched candidate states under common continuations
against the exact readout rule (main-text
Sections~\mainref{si:inverse-mwe}--\mainref{sec:routing-sufficiency}).
The selected representation and hashes of the protocol, implementation and
discovery records are saved before a separate invocation for new comparisons.
These bind local records, without independent registration or external chronology.

Discovery uses four histories combining two association orientations and two
notes, issue times \((-4,-2)\), query \(A\), values \((2,7)\), and
\(S_1\) returning first. Each snapshot candidate has four violating pairs
among six; the association state has two matched pairs without violations,
and full history has none. Minimizing state count after excluding violations
selects the association description, with ties resolved by fixed candidate order.

The subsequent execution design separates history variation from future-input
variation. Sixteen histories combine two association orientations, four new
notes and issue-time pairs \((-6,-1)\) and \((-5,-3)\). Each receives
all twenty-four continuations formed by value pairs
\((3,11),(11,3),(1,13),(13,1),(0,0),(8,8)\), both sensor arrival orders at
times \(1,2\), and both queries. Thus the complete numerical roster behind
main-text Table~\mainref{si:tab-inverse-mwe} gives 384 base executions.

A separate reconstruction reads the recorded protocol, discovery and subsequent
comparisons, requests, returns, predictions and selected representation without
importing the program, state extractor or predictor. It checks complete input
sets, associations, selection, matched pairs and controls; corruption tests
alter outputs, row coverage, representations and control baselines. Boundary
tests cover negative and fractional values, threshold equality and returns at
the horizon. These check the known implementation, not an unknown system.

\subsection{Language-model sampling and comparison}
\label{si:inverse-model-methods}
These comparisons test whether summaries of request histories preserve a
model's later quantity reports. Candidates retain a snapshot (zero current
quantity and two pending requests), that snapshot with issue times or
sensor--recipient associations, or full history. Matched candidate states must
give output laws within the declared tolerance; the descriptions never enter
the model's input.

Constructed requests associate sensors with recipients; returns carry the
matching request identifier, value and arrival time. The identifiers link
returns to requests; their spelling has no task-defined meaning. Inputs also
contain an unrelated text note, which the prompt states is not an instruction.
A continuation specifies the returns and query. Qwen2.5-14B-Instruct~\citep{qwen2024qwen25}
(MLX 4-bit) reads complete records and reports a single-field JSON integer;
other completed responses are INVALID. Request--return matching supplies
reference answers for scoring correctness separately from response agreement. Main-text Sections~\mainref{si:inverse-model}--\mainref{sec:routing-refinement} report the
broader development failure, the recipient-\(A\) restriction chosen afterward,
and the subsequent tests of identifiers, request order and revised states.

\paragraph{Sampling settings and assumptions.}
The restricted and crossed comparisons request temperature \(0.7\), top-p \(1\), top-k
\(0\) and a distinct fixed seed per scientific call. The model prompt,
record schema, model and parser are identical across the comparisons; the new
comparison in Section~\ref{si:refinement-methods} changes the joint seed schedule.

Sixteen instrument-check calls vary omitted, distinct and repeated seeds and
zero temperature. All return the same output and receive no correctness score,
leaving effective seed and temperature handling unverified.

Development queries \(A\) with sensor values \((2,7)\), \(S_1\) first,
or \(B\) with \((7,2)\), \(S_2\) first.
The broader development task queries both recipients. For a continuation
querying \(B\), two association-matched histories should both give 2:
one instead gives 7 in all four calls, the other gives 2 three times and 7 once.
The remaining histories give their reference 7 in all four calls each.
Thus 5/16 responses are incorrect and the matched pair has empirical total
variation (TV) 0.75, the largest event-probability difference computed from
the observed frequencies. This motivates the later recipient-\(A\) restriction.

\paragraph{Histories, comparisons and repeated responses.}
An input condition fixes one history and one continuation. A matched comparison
pairs distinct histories with the same candidate state under a common logical
continuation; several comparisons may reuse the same responses.
All 16 instrument-check calls and 32 development calls are retained;
the restricted selection reuses 16 of the latter, adding no new observations.
The restricted follow-up combines both association orientations with two
identifier/time/note preparations. Its continuations query \(A\) with values
\((3,11)\), \(S_2\) first, or \((13,1)\), \(S_1\) first.
For example, the two histories containing the literal note \texttt{silver-folder} have sensor-ordered
issue times \((-6,-1)\); values \((3,11)\) arrive at times \((2,1)\).
Exchanging associations changes the reference from 3 to 11 at a fixed snapshot.
Four histories under two continuation combinations give eight base input
conditions for comparing histories at the same candidate state. Four additional
control conditions increment queried or unqueried sensor values to test
response and invariance, respectively.
At 64 calls per condition, the follow-up comprises 512 base and 256 control
calls. Repeated calls do not add task cases or establish sampling independence.
Original requests, responses and parsing retain every completed output,
including the additional invalid category specified in the main text.

\paragraph{Reconstruction scope.}
Verification reads original requests and response bytes without new inference.
Local metadata report the model and requested sampling settings; they do not
attest model weights, server sampling or independent execution origin. Hashes
bind request and representation consistency, without authenticating external
registration or chronology.

\paragraph{Crossed identifiers, associations and request order.}
This comparison asks whether the sensor--recipient association state still
preserves responses when request identifiers and list order vary independently.
The model continues to read full records and to report recipient \(A\)'s
quantity. The design in main-text Section~\mainref{sec:crossed-routing}
uses two request-identifier schemes: \((\texttt{q17},\texttt{q42})\) and
\((\texttt{m83},\texttt{m29})\), indexed by sensors \((S_1,S_2)\).
The first association orientation assigns \(S_1\) to \(A\) and \(S_2\) to
\(B\); the second reverses these assignments. Each identifier scheme occurs
under both orientations and both request-list orders. Request issue times are fixed at \((-6,-2)\) by sensor, and the note
is fixed. These assignments remove the earlier correlation of identifier
family with association orientation; they do not assume token invariance.
Return records are ordered by their supplied arrival times. The first
continuation supplies sensor values \((5,17)\), with \(S_2\) returning first;
the second supplies \((19,4)\), with \(S_1\) returning first. Both query
\(A\) at horizon three. Renaming requests also renames the corresponding
return references. Values and arrival order remain two bundled
conditions, rather than a fully crossed pair of factors.

Each comparison uses two distinct histories with the same candidate state
under a common continuation. The snapshot and sensor-sorted issue-time state
each have 56 such comparisons; the sensor-sorted association state has 24, and
full history has none. Eight single-factor pairs test each of identifier
scheme, association orientation and request order. Queried and unqueried
increments use both orientations, the first identifier scheme, the original
request order and the first continuation.

The 20 conditions are interleaved
once per sampling round, with 64 rounds and distinct requested seeds.
Each reference quantity is calculated from the request--return associations.
Lower bounds on its output probability bound the overlap when the reference
quantities agree, or separation when they differ. All comparisons follow from 20
simultaneous one-sided reference-probability bounds, assuming independent draws from a stationary conditional output
distribution within each condition. With 64 responses equal to the reference
the TV upper and lower limits are approximately \(0.08937\) and \(0.82126\);
these are resolution calculations, not guaranteed experimental outcomes.

The combined support criterion requires TV upper bounds at most \(0.15\) for
every association-matched comparison and both unqueried-sensor increments;
TV lower bounds must exceed \(0.50\) for all eight comparisons that change only
association orientation and both queried-sensor increments. All invalid
responses enter the denominator. Missing records or transport failures stop
the experiment without replacement and preclude a confirmatory conclusion.

The crossed comparison leaves association-state agreement unresolved under
the fixed rule. Table~\ref{tab:crossed-counts} gives every condition's counts:
all 1,280 outputs are valid, with 981/1,024 base and 149/256 control responses
equal to their references. Each snapshot candidate has 28 rejected, 21
within-margin and seven unresolved pairs; the association state has 21
within-margin and three unresolved pairs. Full history supplies no matched pair.
Changing identifiers or request order alone gives seven within-margin and one
unresolved pair for each factor; changing association orientation gives seven
resolved differences above \(0.50\) and one unresolved pair. These comparisons
reuse responses rather than add independent tasks.

The controls explain the incomplete support. Both pass in the first orientation;
both remain unresolved in the second, whose base response often gives the other
sensor's value. Increasing that unqueried value produces empirical TV \(63/64\),
yet the reference-probability interval remains approximately \([0,0.99996]\).
The bounds can therefore miss changes among non-reference responses. The table
retains the underlying counts, and the fixed decisions are unchanged.
A separate parser reconstructs all conditions and bounds from saved requests
and responses without the experiment runner or new calls.

\begin{table}[tbp]
\centering\small
\caption{\textbf{Response sensitivity to request names, sensor assignments and list order.}
Request identifiers (IDs) match returns to their original requests; their
spelling has no prescribed effect on the answer. For example, \texttt{q17/q42}
names the requests from \(S_1\) and \(S_2\), respectively, even when their
list order is reversed. Both ID pairs and value pairs follow this sensor order.
The sensor assigned to recipient \(A\) determines the reference answer.
Each row records 64 model responses; changing names, assignments and list order
independently tests which distinctions affect them. For base values
\((5,17)\), \(S_2\) returns first; for \((19,4)\), \(S_1\) returns first.
The four controls use the first continuation, increasing one sensor value by
one. Entries in the final column are answer (count); every omitted integer
and INVALID has count zero. These counts reconstruct both the original
reference-probability bounds and the separate exploratory categorical analysis.}
\label{tab:crossed-counts}
\begin{tabular}{@{}llclrl@{}}
\toprule
Request IDs & Sensor for \(A\) & Request order & Values & Reference & Answers (count) \\
\midrule
\texttt{q17/q42} & \(S_1\) & \(S_1,S_2\) & \((5,17)\) & 5 & \(5\;(64)\) \\
\texttt{q17/q42} & \(S_1\) & \(S_1,S_2\) & \((19,4)\) & 19 & \(19\;(64)\) \\
\texttt{q17/q42} & \(S_1\) & \(S_2,S_1\) & \((5,17)\) & 5 & \(5\;(64)\) \\
\texttt{q17/q42} & \(S_1\) & \(S_2,S_1\) & \((19,4)\) & 19 & \(19\;(64)\) \\
\texttt{q17/q42} & \(S_2\) & \(S_1,S_2\) & \((5,17)\) & 17 & \(5\;(43),\; 17\;(21)\) \\
\texttt{q17/q42} & \(S_2\) & \(S_1,S_2\) & \((19,4)\) & 4 & \(4\;(64)\) \\
\texttt{q17/q42} & \(S_2\) & \(S_2,S_1\) & \((5,17)\) & 17 & \(17\;(64)\) \\
\texttt{q17/q42} & \(S_2\) & \(S_2,S_1\) & \((19,4)\) & 4 & \(4\;(64)\) \\
\texttt{m83/m29} & \(S_1\) & \(S_1,S_2\) & \((5,17)\) & 5 & \(5\;(64)\) \\
\texttt{m83/m29} & \(S_1\) & \(S_1,S_2\) & \((19,4)\) & 19 & \(19\;(64)\) \\
\texttt{m83/m29} & \(S_1\) & \(S_2,S_1\) & \((5,17)\) & 5 & \(5\;(64)\) \\
\texttt{m83/m29} & \(S_1\) & \(S_2,S_1\) & \((19,4)\) & 19 & \(19\;(64)\) \\
\texttt{m83/m29} & \(S_2\) & \(S_1,S_2\) & \((5,17)\) & 17 & \(17\;(64)\) \\
\texttt{m83/m29} & \(S_2\) & \(S_1,S_2\) & \((19,4)\) & 4 & \(4\;(64)\) \\
\texttt{m83/m29} & \(S_2\) & \(S_2,S_1\) & \((5,17)\) & 17 & \(17\;(64)\) \\
\texttt{m83/m29} & \(S_2\) & \(S_2,S_1\) & \((19,4)\) & 4 & \(4\;(64)\) \\
\midrule
\multicolumn{6}{@{}l}{\emph{Queried- and unqueried-sensor controls}} \\
\texttt{q17/q42} & \(S_1\) & \(S_1,S_2\) & \((6,17)\) & 6 & \(6\;(64)\) \\
\texttt{q17/q42} & \(S_1\) & \(S_1,S_2\) & \((5,18)\) & 5 & \(5\;(64)\) \\
\texttt{q17/q42} & \(S_2\) & \(S_1,S_2\) & \((5,18)\) & 18 & \(5\;(44),\; 18\;(20)\) \\
\texttt{q17/q42} & \(S_2\) & \(S_1,S_2\) & \((6,17)\) & 17 & \(6\;(63),\; 17\;(1)\) \\
\bottomrule
\end{tabular}
\end{table}

\paragraph{Exploratory categorical comparison.}
Reference-probability bounds can miss changes among incorrect responses. A
post-hoc categorical reanalysis resolves differences in the three association-matched
comparisons left unresolved, using the scientific-evidence comparison's simultaneous
component intervals with a different output alphabet.
The six quantities appearing in the recorded returns supplied to the model across
all 20 conditions are \(4,5,6,17,18,19\). Each is a category; other valid
integers form a seventh category and invalid outputs an eighth. This rule
depends on the supplied inputs, not the observed outputs or reference answers.
All categories, including zero-count categories, and all conditions enter
the uncertainty calculation. Each component receives a two-sided
Clopper--Pearson interval with per-tail error
\(\eta=0.05/(2\cdot20\cdot8)=0.00015625\). Bonferroni control gives nominal
simultaneous coverage at least 95\% for this specified family under stationary
independent sampling within each condition. The method was chosen after the
outcomes; this family is separate from the original reference-probability analysis and
does not give joint 95\% coverage across analyses.

Requiring category probabilities to lie in their intervals and sum to one
bounds every category-subset probability, and hence TV, as in
Section~\ref{si:scientific-evidence-methods}.
For those three association-matched pairs, the event ``quantity equals 5'' already attains
the resulting lower bound. Its counts are 43/64 in one condition and 0/64 in
the other. Writing \(B_q^{-1}(a,b)\) for the beta-distribution quantile,
the separation lower bound is
\[
 B_\eta^{-1}(43,22)-B_{1-\eta}^{-1}(1,64)
 \simeq 0.3121972.
\]
The other 21 association pairs have lower bound zero; this analysis makes no
positive adequacy classification. Category grouping cannot increase TV, so the
lower bound transfers to the original integer/invalid laws; a grouped upper
bound would not establish their agreement. A state-only predictor must use
one law for both inputs. The half-separation bound (main-text
Section~\mainref{sec:compression-error}) therefore gives worst-case behavioral
error at least \(0.1560\), not a task error rate. All 24 pairs and original decisions remain reported; the reanalysis adds no calls.

\subsubsection{Testing added identifiers and request order}
\label{si:refinement-methods}
The new comparison tests whether request identifiers and list order are needed
in addition to sensor--recipient associations on new recipient-\(A\) inputs.
The analyst compares the simpler and revised states; the model's task and
full-record inputs remain fixed. Main-text
Section~\mainref{sec:routing-refinement} reports the results.
All requests, candidates, margins, sample sizes and analysis rules were fixed
before its responses. No observations from the earlier studies enter its bounds.
The model, prompt, strict integer/invalid parser, temperature \(0.7\), top-p
\(1\), top-k \(0\), and 96-token output limit are unchanged.

\paragraph{Inputs and candidate states.}
The revision retains the snapshot and ordered
\((\text{request ID},\text{sensor},\text{recipient})\) triples. Two further
candidates retain order without IDs, or IDs without order.
Two arbitrary request-ID pairs, \((\texttt{v36},\texttt{v91})\) and
\((\texttt{r64},\texttt{r28})\), link sensors \((S_1,S_2)\) to their returns.
Each pair occurs under both association orientations and request orders.
To create distinct histories with the same revised state, two preparations
jointly vary request times and the unrelated input note, both omitted by that
state. They use issue times \((-9,-3)\) with the literal text
\texttt{grey-folder}, or \((-2,-7)\) with \texttt{orange-folder}; these are
note contents, not file references. Both requests precede checkpoint zero;
relative issue chronology reverses. Timing and note vary together, so their
separate effects are not identified.
The two continuations bundle values \((6,23)\) and \((27,8)\) with opposite
arrival orders, both querying \(A\) at \(T=3\).

Each comparison matches two distinct histories at the same candidate state
under a common continuation. Their counts are 240 for the snapshot, 112 for
associations, 48 for each partial extension, 16 for the revision and zero for
complete history. Forty-eight of the association comparisons have identical
raw future returns; the other 64 consistently relabel return references.
All primary revision pairs hold raw future returns identical. Four controls
increment the queried or unqueried sensor in each orientation, using the first
ID scheme, original request order, first timing/note setting and first
continuation. They cover these layouts only.

\paragraph{Reference-free simultaneous bounds.}
These bounds compare complete output laws using category frequencies for
separation and shared probability mass, and paired-output disagreement for an
additional agreement bound. The inputs above form 32 base conditions and four
controls, giving 36 conditions in total.
Each of 256 replicates requests one common seed for all 36 conditions;
replicates use distinct seeds and fixed interleaved condition orders.
Inference assumes independent stationary replicate vectors with the declared
marginals, allowing dependence within a replicate. Requested seeds do not
attest these assumptions or their implementation.

Let \(\mathcal S=\{6,7,8,23,24,27,\mathrm{INVALID}\}\) denote the seven
individual parsed outcomes fixed from the inputs and parser. Each is kept
separate; all other valid integers are grouped into one residual category
for computation only. Allocate error
\(0.025\) across both tails of all \(36\times8\) component intervals,
using per-tail error \(\eta=0.025/(2\cdot36\cdot8)\).
Requiring category probabilities to lie in these intervals and sum to one gives
lower and upper event-probability bounds \(\ell_i(E),u_i(E)\), as in
Section~\ref{si:scientific-evidence-methods}. Over subsets of the eight categories,
\[
 \underline d_{ij}=\max_E\{0,\ell_i(E)-u_j(E),\ell_j(E)-u_i(E)\}
\]
lower-bounds TV of the original integer/invalid laws. For agreement, the
unresolved integers in the residual category cannot be counted as shared
output. Overlap in the individual categories alone gives the conservative upper bound
\(\overline d^{\,\mathrm{marg}}_{ij}\), where \(\mathrm{marg}\) denotes the
marginal output distributions:
\[
 \overline d^{\,\mathrm{marg}}_{ij}
 =1-\min_{E\subseteq\mathcal S}
       \{\ell_i(E)+\ell_j(\mathcal S\setminus E)\}.
\]

Write \(Y_i\) for the full parsed output in condition \(i\), with probability
law \(\Pi_i\). Under any joint sampling with these marginal laws,
\(d_{\mathrm{TV}}(\Pi_i,\Pi_j)\le\Pr(Y_i\ne Y_j)\).
The reported pair family contains all snapshot-matched comparisons and the
four controls; other candidate comparisons are subsets of the base family.
Count unequal full parsed outputs in each paired replicate, including INVALID
and retaining distinct residual integers. One-sided exact binomial upper
bounds allocate the remaining \(0.025\) error across all 244 distinct reported
pairs. The minimum of this upper bound and
\(\overline d^{\,\mathrm{marg}}_{ij}\), together with
\(\underline d_{ij}\), has nominal simultaneous coverage at least 95\%
under the sampling assumptions. Pairwise independence is unnecessary.
Reference quantities enter only the separate correctness score.

\paragraph{Decisions and complete outcomes.}
The fixed margins are TV \(0.15\) for agreement and \(0.50\) for positive
controls. The primary refinement criterion requires both rejection of the
association-only description on the new inputs and agreement for all revised
pairs; the combined criterion also requires the four controls. Complete history
is uninformative. No retry, replacement, outcome-dependent extension or deletion
is permitted; all 9,216 calls complete with valid outputs.

Both revised and association-only states have bounded agreement, and the
controls pass. The primary refinement criterion fails because the simpler
state is not rejected. Agreement on these inputs leaves the earlier exploratory
separations unchanged.
The snapshot has 128 rejections and 112 within-margin comparisons; all other
nontrivial candidates are within margin on their full declared comparisons.
Of the 16 revised-state pairs, 15 have no paired output mismatch and one has
2/256 mismatches, giving upper bounds \(0.0353\) and \(0.0531\), respectively.
The 112 association pairs have these bounds in 105 and seven pairs;
both partial extensions have 45 and three such pairs. Snapshot rejection
lower bounds exceed \(0.9048\). The two positive-control lower bounds exceed
\(0.9230\); both invariance-control upper bounds are at most \(0.0353\).

The only two wrong responses occur with the second ID scheme, \(S_2\) assigned
to \(A\), original request order, second timing/note setting and first
continuation: 254 responses give 23 and two give the other sensor's value 6.
All other cells have 256 correct responses. The authors retain requests and
raw responses. A separately implemented parser
and analysis reproduce every count, bound and classification without the
experiment runner or new inference. This checks the recorded results, not
the server's implementation of the requested sampling settings.

\section{Methods for inverse tests under tool feedback}
\label{si:closed-loop}

This instructed workflow tests whether an initial summary must retain software
version information to predict the next check and release decision. Release
requires both unit and integration checks to pass for the current version.
Each history has two focal PASS results: one current and one predecessor result,
for different check types. Exchanging their version assignments changes which
current check is missing while preserving both statuses. Confirmation adds a
redundant predecessor unit-check result.

Requests identify version and check type; returns identify their requests.
The experiment program joins them before Qwen2.5-14B-Instruct (MLX 4-bit)
reads the completed results and outstanding requests. Model decisions use
fresh calls; histories, checks, release and hold are constructed or simulated,
without generated-code execution or external deployment. Main-text
Section~\mainref{sec:closed-loop} reports the comparisons.

The instructions below specify the complete decision rule. The model reports
unit status, integration status, then action, in that order. The schema allows every combination of PASS, FAIL, PENDING or
ABSENT with the five actions; it supplies no assessed status or correct action.
Requested temperature is \(0.7\), top-p \(1\), top-k \(0\), with an 80-token
completion limit and distinct recorded seeds. Duplicate, missing, extra or
misordered fields, disallowed values, malformed JSON and unfinished completions
are invalid. Recorded server metadata identify the loaded model, without
cryptographic verification of its weights.

The verbatim instruction prescribes the status assessment and identifier/order
invariance; the current-version candidate retains this explicitly required
information. Underscores identify transmitted JSON fields.
\begin{quote}
Schedule validation of the current software artifact. Both unit and integration checks must
PASS on this exact artifact before release. completed\_results lists finished checks with
their artifact, check type and status. pending\_requests lists checks requested but not
returned. These lists are complete. An old-artifact check is neither a completed nor pending
check for the current artifact. First assess unit\_status and integration\_status for the
CURRENT artifact: PASS or FAIL for its completed check; PENDING for its outstanding request;
ABSENT if neither exists. Then choose the next action. If either current check is FAIL, HOLD.
If both are PASS, RELEASE. If a needed current check is PENDING, WAIT. Otherwise RUN\_UNIT or
RUN\_INTEGRATION for an ABSENT check. Return JSON in this exact field order: unit\_status
first, integration\_status second, action last. Record order and identifier spelling do not
change which artifact was checked.
\end{quote}

The continuation fixes the instructed action rule, simulated check outcome
and whether the return is delivered or withheld (the observation branch). A requested check returns the externally fixed PASS or FAIL
for the current version, hidden from the first model call. Two branches reuse
that sampled first action: one receives the executed request and return;
the other receives the request with its return withheld. Fresh second calls use
the same instructions, in an order fixed by a shuffled schedule. A first action other than requesting a test
terminates the episode. The measured sequence therefore contains
at most two decisions. Its sixteen categories are each of the two test-request actions
followed by any of the five actions or INVALID, together with four sequences
ending at the first decision: WAIT, RELEASE, HOLD and INVALID. A second test request is recorded but does
not extend the window. Status assessments and arbitrary generated text are not
part of this readout.

Three predefined, analyst-only summaries retain respectively the two summarized
statuses and pending-request count; the latest status for each check without
version information; or unit and integration results for the current version,
including absence. They never replace the model's records. The first two are constant and induce
the same partition on the tested initial histories; their repeated rejection
does not represent independent discoveries. The third omits
identifiers, row order and redundant predecessor results while separating the
two missing-check situations. These initial preparation summaries omit
later pending flags, so agreement does not test autonomous state evolution.
Automatic discovery and uniqueness of the summary are also untested.

\paragraph{Interface selection and confirmation.}
Development screens interfaces for following the supplied decision rule. Two missing-check situations and two
return conditions receive eight starting episodes each. The fixed screen
requires at least seven correct first choices and seven correct feedback
sequences in every condition.

Action-only output gives RELEASE on all 32 calls. A separate presentation test
(32 calls) finds no format meeting both missing-check requirements. Joined
records with status assessment give two correct first choices in 32 episodes
(36 calls): alphabetic schema ordering requests action before assessment.
Preserving the intended field order instead gives 30 correct choices and
feedback sequences in 32 episodes (92 calls), with counts \(8,8,7,7\),
passing the screen. Confirmation retains this interface and its inference and
analysis code; correcting an empty-input test fixture changes neither.
Development responses fit forecasts but do not enter confirmation counts.

Confirmation crosses two missing-check situations, two identifier schemes and
two request-list orders, giving eight histories, each under PASS and FAIL:
16 conditions with 64 starting episodes each. New version names, request
identifiers and the redundant predecessor result distinguish these inputs from
development; the two version names are fixed within confirmation, so version-name
invariance is not tested. Within each sampling round every condition occurs
once, in a fixed shuffled order; branch order is also fixed before execution.
The local record binds the source revision, analysis files, development
forecasts, model configuration and sampling limit before the recorded calls.
This is a local execution record, not independent registration or external
attestation of chronology.
All 1,024 episodes and 3,072 calls completed. These repeats do not constitute
1,024 independently sampled tasks. Invalid outputs remain in every denominator;
transport failure stops execution without retry, incomplete confirmation permits
no acceptance, and outcomes never trigger additional sampling.

\paragraph{Comparing action-sequence distributions.}
Histories merged by a candidate summary are compared under the same imposed
tool outcome. Category-probability intervals bound how much their
sequence distributions can differ, including invalid outputs.
The 16 history/outcome conditions each have a delivered-return and a
withheld-return branch, giving \(M=32\) distributions. Let \(X_{ik}\) count
sequence category \(k\) in distribution \(i\), and \(p_{ik}\) denote its
probability, with \(n=64\) observations and \(K=16\) categories. Conditional on the fixed
development fit and interface, assume independent, stationary repetitions
within each distribution. The branches share their first action and are not
independent trials; independence across branches, conditions or categories is
unnecessary. Seeds and interleaving do not establish these assumptions.
For family error \(\alpha=0.05\), allocate \(t=\alpha/(2MK)\)
to every interval tail, including zero-count categories. With \(B_q^{-1}(a,b)\)
the beta-distribution quantile, the Clopper--Pearson bounds are\cite{ClopperPearson1934}
\begin{equation}
 L_{ik}=B_t^{-1}(X_{ik},n-X_{ik}+1),\qquad
 U_{ik}=B_{1-t}^{-1}(X_{ik}+1,n-X_{ik}),
\end{equation}
using \(L_{ik}=0\) at zero counts and \(U_{ik}=1\) at \(n\).
The union bound over all \(2MK\) tails gives joint nominal coverage at least
95\% under these assumptions. On this event each true law lies in
\(\mathcal P_i=\{p:\sum_kp_k=1,\ L_{ik}\le p_k\le U_{ik}\}\).

Total variation is \(d_{ij}=\tfrac12\sum_k|p_{ik}-p_{jk}|\).
Define the unavoidable coordinate gap
\(g_{ijk}=\max\{0,L_{ik}-U_{jk},L_{jk}-U_{ik}\}\). The fixed bounds are
\begin{equation}
 \underline d_{ij}=\max\!\left\{\max_k g_{ijk},\tfrac12\sum_k g_{ijk}\right\},
 \qquad
 \overline d_{ij}=1-\sum_k\min\{L_{ik},L_{jk}\}.
\end{equation}
The lower bound follows from coordinate differences and the upper bound from
guaranteed shared probability mass. Neither uses correctness labels.
Bounded agreement requires \(\overline d\le0.15\), rejection requires
\(\underline d>0.15\); otherwise the comparison is unresolved. Identical
observed point masses give \(\overline d=1-t^{1/64}\le0.1437\);
distinct point masses give \(\underline d=2t^{1/64}-1\ge0.7126\).
For these settings, any dispersed observed histogram prevents agreement at
\(0.15\): the smallest attainable upper bound for such a comparison exceeds
\(0.1791\). This limits resolution without invalidating the bounds or proving
that all-identical samples arose from deterministic laws.

Only histories with equal candidate summaries and the same return condition
are paired in the feedback branch: 56 pairs per simpler summary and 24 for
current-version results. Primary support requires a resolved violation for
each simpler summary under each return condition, agreement on all 24 revised
pairs, and lower bounds above \(0.50\) for all eight same-history PASS/FAIL
controls. Feedback participation separately requires that threshold for all
16 feedback/withheld-return contrasts. These comparisons reuse observations and
the same confidence event. For two merged histories separated by \(d\), any
common predicted law has worst-case TV error at least \(d/2\), by the triangle
inequality. The observed separation implies at least \(0.3563\); this is a
behavioral approximation bound, not a task-error rate or an exact optimal error.

\paragraph{Comparing fitted forecasts.}
The fitted forecasts compare action-sequence prediction from the status-only
and current-version summaries, conditional on the initial summary, known
PASS/FAIL forcing and observation branch. Saved before
confirmation, they use no sampled future action, assessment or acquired
confirmation result; they do not forecast an unknown test outcome.
Development sequences in Table~\ref{tab:closed-loop-development-counts}
are pooled by these inputs. If \(N_{gk}\) counts category \(k\) in group \(g\),
the fixed forecast is \(q_{gk}=(N_{gk}+1/16)/(N_g+1)\), where
\(N_g=\sum_kN_{gk}\). Both forecasts use a total uniform pseudocount of one:
\(N_g=16\) for the status-only summary and \(N_g=8\) for current-version
results. The instructed rule gives a point-mass forecast from the same
records and forcing without fitting.

\begin{table}[tbp]
\centering\small
\caption{\textbf{Development counts used to fit the closed-loop forecasts.}
Each row contains eight starting episodes. Unit and Integration denote the
actions requesting those tests; an arrow gives the second action. Entries are
sequence (count), with zero for all omitted categories among the sixteen
defined above. WAIT alone terminates at the first decision and is counted in
both branches without an additional call. The branches share their first
action; the table represents 32 episodes and 92 calls, not 64 independent
trials. Pool rows with the same return condition for the status-only forecast;
keep the missing-test distinction for the current-version forecast.}
\label{tab:closed-loop-development-counts}
\begin{tabularx}{\linewidth}{@{}llYY@{}}
\toprule
Missing test & Return & Feedback sequences & Return-withheld sequences \\
\midrule
Unit & PASS & Unit \(\to\) RELEASE (8) & Unit \(\to\) WAIT (8) \\
Unit & FAIL & Unit \(\to\) HOLD (8) & Unit \(\to\) WAIT (8) \\
Integration & PASS & Integration \(\to\) RELEASE (7); WAIT (1) & Integration \(\to\) WAIT (7); WAIT (1) \\
Integration & FAIL & Integration \(\to\) HOLD (7); WAIT (1) & Integration \(\to\) WAIT (7); WAIT (1) \\
\bottomrule
\end{tabularx}
\end{table}

Forecast accuracy is measured by the normalized Brier loss\cite{brier1950verification},
\(\ell(q,k)=\tfrac12\sum_j(q_j-\mathbf1_{j=k})^2
=\tfrac12(1+\|q\|_2^2)-q_k\).
Expected loss is linear in the unknown probabilities. Its extrema over
\(\mathcal P_i\) are obtained by starting at \(L_i\) and allocating remaining
mass up to \(U_i\), in increasing or decreasing loss order. Let
\(q_i^{\mathrm{s}},q_i^{\mathrm{v}}\) denote the status-only and current-version
forecasts and \(\mathcal F\) the 16 feedback conditions. Their equally weighted
expected loss difference \(\Delta\) has bounds
\begin{equation}
 \underline\Delta=\frac1{16}\sum_{i\in\mathcal F}
 \min_{p\in\mathcal P_i}\sum_k p_k
 [\ell(q_i^{\mathrm{s}},k)-\ell(q_i^{\mathrm{v}},k)],
 \qquad \overline\Delta\text{ replaces min by max}.
\end{equation}
Optimizing the difference retains the common outcome law for both forecasts.
The fixed prediction criterion is \(\underline\Delta>0\).
The fitted current-version forecast meets this criterion: the observed mean
loss difference is approximately \(0.2378\), and the conservatively rounded
lower bound on the expected difference is \(\Delta\ge0.1180\).
The corresponding outward-rounded expected-risk intervals
are \([0.2517,0.3194]\), \([0.0181,0.1379]\) and \([0,0.1437]\) for status-only,
current-version and instructed-rule forecasts, respectively; the rule has zero
observed loss. The loss-difference interval is \([0.1180,0.2379]\).
All these bounds share the original confidence event, conditional on the fitted
forecasts, and do not evaluate repeated refitting on new development samples.
Comparisons of these fitted forecasts establish neither each summary's optimal
loss nor superiority over the instructed rule. In confirmation every
observed first action is identical within its condition. The withholding
comparison therefore tests one subsequent decision for that action, not
robustness across alternative first actions or longer feedback sequences.

A summary can preserve the final decision while omitting how it was reached.
A supplementary analysis chosen after interim observations tests each sequence's
last action, including INVALID. Minimizing and
maximizing each
terminal-action indicator over the original \(\mathcal P_i\) supplies component
bounds for the projected laws; applying the same TV formulas yields conservative
comparisons. All 56 status-only pairs then have bounded agreement, whereas 32
reject for full sequences; PASS/FAIL controls remain resolved. This projection
adds neither observations nor an independent confidence guarantee, and leaves
the fixed primary classifications unchanged. A terminal action is not a terminal
state, and agreement at that readout does not validate the initial summary for
intermediate actions or autonomous dynamics.

Separate pilots test whether feedback improves task outcomes. In one, a
controller adjusts capacity over four decisions to regulate the difference
between demand and capacity. Its free-JSON attempt
stopped at request 11 after an unfinished response; a separate structured-output
run completed 104 calls. Returned observations changed trajectories, but all
four nonzero pulses failed the regulation criterion; two of five evaluation
amplitudes exceeded the fixed scalar-prediction error limit.

A second pilot tests feedback-guided acquisition of scientific evidence.
Its 216 calls reuse eight SciFact training claims in sixteen excerpt conditions,
once per arm. Feedback and frozen observations each give 11/16 final
source-label agreements, versus 12/16 for retrieval-order acquisition with
fewer calls. These are
descriptive source-label comparisons, not adjudicated accuracy for modified
inputs. Neither pilot supplies task-benefit evidence or enters this confidence
family. The two-check experiment excludes dependency-graph faults, cancellation,
retries and interacting agents; it does not validate the broader stopped study
in Section~\ref{si:workflow-development}.

\section{Methods for scientific-text comparisons}
\label{si:scientific-evidence-methods}

This experiment asks whether a document list or an unordered set of excerpts
preserves a reader's responses when the selected text or its document order
changes. The reader receives a scientific claim and excerpts, then judges
support, refutation or insufficient evidence. Repeated judgments estimate
output distributions; comparing these distributions tests the candidate
representations, separately from answer correctness.

\paragraph{Inputs and comparisons.}
SciFact associates scientific claims with research abstracts and annotated
evidence sentences\cite{wadden2020fact}. The design uses the first eight entries
of a fixed, ordered list of 96 claims from its official training split, previously used
in the action-selection development study (89 of 96 tasks completed;
Section~\ref{si:development-decisions}), without substitution or screening by model outputs.
Claim-only BM25 retrieval, a keyword-based ranking procedure, supplies six
abstracts per claim, with term-frequency saturation parameter \(k_1=1.2\)
and document-length normalization parameter \(b=0.75\). It uses Lucene
inverse-document-frequency weighting and document-identifier tie-breaking. Selecting the first
or last three sentences of each abstract, then retaining or reversing document
order, gives \(M=8\times2\times2=32\) input conditions (cells), each containing
18 sentences. Sentence order within each abstract is preserved.
Leading and trailing excerpts overlap in seven of the 48 retrieved abstract
occurrences; selection policies do not imply disjoint text.

Two analyst-defined states retain either the claim, document identities and
excerpt counts, or the claim and unordered exact excerpts with titles, document
identities and sentence indices. A matched pair consists of two input conditions
with the same candidate state. The document-level state matches all pairs of
the four conditions; the excerpt state matches the two document orders within
each excerpt policy. These give six and two matched pairs per claim,
respectively.

Two excerpt-policy contrasts at fixed document order provide positive controls
for detecting a response. The reader receives ordered text, without candidate states
or annotations. Annotations measure complete-rationale retention only; missing
annotated evidence neither defines an insufficient-evidence answer nor supplies
a correctness target for shortened inputs.

\paragraph{Sampling and simultaneous intervals.}
Input conditions, sampling settings and comparison rules were fixed before the
calls and checked against local hashes during execution. These records document
local preparation and integrity, without independent timestamp attestation.
The fixed Qwen2.5-14B-Instruct reader (MLX 4-bit) received 64 calls per cell,
interleaved across cells:
2,048 calls, not additional claims. Requested temperature is \(0.7\),
top-p \(1\), top-k \(0\), with a 16-token output limit.
After removing surrounding whitespace, exact A/B/C completions encode support,
refutation and insufficient evidence;
all other completions are invalid. Thus the alphabet is
\(\mathcal{Y}=\{\mathrm{SUPPORTS},\mathrm{REFUTES},\mathrm{NEI},\mathrm{INVALID}\}\),
with \(K=4\). Let \(X_{ik}\) count category \(k\) in cell \(i\),
\(n_i=\sum_kX_{ik}\), and \(p_{ik}\) its probability. Full counts, including
invalid responses, enter every comparison without using reference answers.
Transport or record-integrity failure halts sampling without retry; incomplete
runs yield no adequacy conclusion, and thresholds never trigger extra draws.

The intervals quantify uncertainty in each response-category probability before
those probabilities are compared across matched inputs.
Assume fixed sample sizes and independent stationary repeats within each cell;
requested seeds establish neither property. For family error \(\alpha=0.05\),
set the per-tail error budget to \(t=\alpha/(2MK)\).
Writing \(B_q^{-1}(a,b)\) for the \(q\)-quantile of the beta distribution
with shape parameters \(a,b\), the Clopper--Pearson
intervals \([L_{ik},U_{ik}]\)\cite{ClopperPearson1934} are
\begin{equation}
L_{ik}=B_t^{-1}(X_{ik},n_i-X_{ik}+1),\qquad
U_{ik}=B_{1-t}^{-1}(X_{ik}+1,n_i-X_{ik}).
\end{equation}
Use \(L_{ik}=0\) when \(X_{ik}=0\), \(U_{ik}=1\) when \(X_{ik}=n_i\),
and \([0,1]\) when \(n_i=0\). Each tail fails with probability at most \(t\);
the union bound over \(2MK\) tails gives simultaneous coverage at least
\(1-\alpha=0.95\), without independence across cells or categories.

\paragraph{Event and distance bounds.}
A response event is any subset of output categories, such as support or
refutation. To test whether matched inputs have similar output laws, bound the
largest probability difference over all such events. On the simultaneous
coverage event above, each category probability lies in its interval and
the probabilities sum to one. For every nonempty proper subset
\(E\subset\mathcal{Y}\), with complement \(E^c\), its probability extrema are
\begin{equation}
\ell_i(E)=\max\!\left\{\sum_{k\in E}L_{ik},1-\sum_{k\in E^c}U_{ik}\right\},\qquad
u_i(E)=\min\!\left\{\sum_{k\in E}U_{ik},1-\sum_{k\in E^c}L_{ik}\right\}.
\end{equation}
Indeed, the attainable sum intervals on \(E\) and its complement intersect at
exactly these endpoints after normalization. Total variation,
\(d_{ij}=\tfrac12\sum_k|p_{ik}-p_{jk}|=\max_E[p_i(E)-p_j(E)]\), has
simultaneous lower and upper bounds \(\underline d_{ij}\) and
\(\overline d_{ij}\):
\begin{equation}
\begin{aligned}
\underline d_{ij}&=\max\!\left\{0,\max_E[\ell_i(E)-u_j(E)],
                                      \max_E[\ell_j(E)-u_i(E)]\right\},\\
\overline d_{ij}&=\min\!\left\{1,\max_E[u_i(E)-\ell_j(E)]\right\},
\qquad \underline d_{ij}\le d_{ij}\le\overline d_{ij}.
\end{aligned}
\end{equation}
Each event difference is bounded separately; all 14 subsets include both
directions through complements. All pair bounds share the same coverage event,
requiring no further multiplicity correction. An unsampled cell gives a
conservative pair interval \([0,1]\).
For two samples concentrated on the same category, 64 draws per cell give
\(\overline d=1-t^{1/64}\le0.1250\), below the \(0.15\) agreement margin.
Concentration on distinct categories gives
\(\underline d=2t^{1/64}-1\simeq0.7501\). These precision cases motivate the
fixed repeat count without guaranteeing a resolved comparison.
For this sample size, alphabet and simultaneous error allocation, agreement
can be certified only for two observed samples concentrated on the same
category. Any dispersed observed histogram makes the upper bound exceed
\(0.15\); this restriction concerns the samples and comparison rule, not
determinism of the underlying laws. An unresolved interval can also
accompany a substantial empirical difference. Its overlap with the margin
therefore supports neither agreement nor rejection; increasing sample size
would improve precision, not predetermine either conclusion.

\paragraph{Reporting and scope.}
An upper bound at most \(0.15\) gives bounded agreement; a lower bound above
\(0.15\) rejects that tolerance; otherwise the comparison is unresolved.
Informative adequacy for an excerpt-set comparison additionally requires
\(U_{i,\mathrm{INVALID}}\le0.15\) at both matched cells and at both cells of
at least one same-claim policy control with \(\underline d>0.5\).
These reporting conditions reuse the simultaneous intervals; they do not
filter responses or stop calls. Failed validity bounds withhold positive adequacy support. Resolved violations
of the four-category tolerance remain reportable, including differences in
format failure; they do not by themselves isolate a semantic cause. Identical invalid outputs cannot supply
adequacy support; identical wrong valid labels can supply bounded agreement.
Approximate agreement is not transitive and does not define equivalence
classes (main-text Section~\mainref{sec:information-requirements}). Conclusions concern
output distributions for these claims, not correctness, held-out performance,
automatic state discovery or general sufficiency.

\paragraph{Complete observations.}
Both state descriptions have resolved violations on the full tested claim set
under the stated sampling assumptions. All planned calls completed; a separate
calculation reproduced every count, comparison and resource total.
Table~\ref{tab:scientific-evidence-counts} retains
all responses: 2,048 repeated measurements of eight claims.

\begin{table}[tbp]
\centering
\small
\setlength{\tabcolsep}{3pt}
\caption{\textbf{All scientific-evidence response counts.}
The reader judges a claim from six retrieved abstracts after excerpt selection
and document ordering. Columns give these input conditions; tuple entries count
model judgments, not correct answers.
Each tuple is ordered (SUPPORTS, REFUTES, NEI, INVALID) and sums to \(n=64\).
NEI denotes insufficient evidence; INVALID retains all other completions.
Leading and trailing select the first and last three sentences per abstract;
retrieval and reversed denote document order. Claim wording is reproduced as
input, with its SciFact source identifier in parentheses, rather than as a
scientific conclusion. In the quoted inputs, Ab denotes antibody and eRNAs
denotes enhancer RNAs. The simultaneous invalid-probability upper bound is
at most \(0.1250\) in every cell.}
\label{tab:scientific-evidence-counts}
\begin{tabularx}{\linewidth}{@{}Y*{4}{>{\centering\arraybackslash}p{0.12\linewidth}}@{}}
\toprule
& \multicolumn{2}{c}{Leading excerpts} & \multicolumn{2}{c}{Trailing excerpts}\\
Source claim (identifier) & Retrieval & Reversed & Retrieval & Reversed\\
\midrule
Depletion of T-helper 17 (Th17) cells during simian immunodeficiency virus (SIV) infection decreases dissemination of Salmonella Typhimurium from the gut. (332)
& \(64,0,0,0\) & \(64,0,0,0\) & \(64,0,0,0\) & \(64,0,0,0\)\\
\addlinespace
RAD52 is involved in break-induced DNA replication (BIR). (1003)
& \(64,0,0,0\) & \(64,0,0,0\) & \(62,0,2,0\) & \(32,2,30,0\)\\
\addlinespace
Exhaustion of B cells contributes to poor Ab response in HIV-infected individuals. (398)
& \(0,0,64,0\) & \(0,0,64,0\) & \(0,1,63,0\) & \(0,0,64,0\)\\
\addlinespace
40mg/day dosage of folic acid and 2mg/day dosage of vitamin B12 does not affect chronic kidney disease (CKD) progression. (12)
& \(0,0,64,0\) & \(0,0,64,0\) & \(64,0,0,0\) & \(0,12,52,0\)\\
\addlinespace
Binding of p53 to enhancer regions (p53BERs) regulates the activity of p53 transcription and cell cycle arrest through the production of p53-dependent eRNAs. (177)
& \(62,0,2,0\) & \(58,2,4,0\) & \(64,0,0,0\) & \(64,0,0,0\)\\
\addlinespace
Medications to treat obesity have unwanted side effects. (766)
& \(63,1,0,0\) & \(64,0,0,0\) & \(64,0,0,0\) & \(64,0,0,0\)\\
\addlinespace
Acute ablation of Snail in the embryonic cortex affects the proliferation and number of embryonic cortical precursors. (81)
& \(10,3,51,0\) & \(0,0,64,0\) & \(63,1,0,0\) & \(64,0,0,0\)\\
\addlinespace
taa1 tar1 tar2 triple mutants in Arabidopsis do not make root meristem during embryogenesis. (1405)
& \(0,0,64,0\) & \(0,0,64,0\) & \(0,0,64,0\) & \(0,0,64,0\)\\
\bottomrule
\end{tabularx}
\end{table}

Across the 48 pairs matched by claim,
document identities and excerpt counts, 21 have bounded agreement, nine
reject the \(0.15\) tolerance and 18 remain unresolved. The 16 pairs also
matched by exact unordered excerpts are a subset: nine have bounded agreement,
one rejects and six remain unresolved. Of the remaining pairs, 16 change
only excerpt policy and 16 change both policy and order; each group has six
bounded, four rejected and six unresolved comparisons. The two rejections
for claim 1003 have lower bound \(0.1578\), slightly above the fixed
\(0.15\) tolerance. Only the leading-excerpt pair for claim 12 additionally
satisfies the validity and positive-control conditions. Two of the 16
fixed-order policy controls exceed the \(0.5\) lower-bound threshold:
claim 12 in retrieval order and claim 81 in reversed order. The trailing-excerpt
pair for claim 12 has \(\underline d\simeq0.75014\).

Every packet for claim 332 retains a complete annotated contradicting rationale,
yet all 256 responses are SUPPORTS. Their agreement lacks a detectable positive
control; the annotation disagreement is not an adjudicated accuracy score
(main-text Section~\mainref{sec:scientific-evidence}). The comparisons concern
these eight reused claims.

\FloatBarrier
\section{Graph response methods and additional workflow studies}
\label{si:empirical-realizations}

These studies ask three distinct questions: whether a prescribed graph response
is calculated correctly, whether response measurements diagnose an output error,
and whether an intervention improves the task outcome. The sections separate
deterministic checks, fresh model measurements and exploratory interventions.
Constructed cases do not estimate deployment failure rates; repeated calls do
not add tasks. Section~\ref{si:implementation-guide} summarizes reconstruction
coverage.

\subsection{Prescribed graph responses and model readouts}
\label{si:empirical-protocols}

\paragraph{Response operators and numerical checks.}
\label{iclr:additional-methods}

The numerical checks ask whether the specified dynamics transmit a source
change into the predicted state and readout changes.
Table~\ref{tab:rpw-specialization} defines the local source operators;
Section~\ref{si:susceptibility-comparison} defines the graph response and
derivative conventions used in these checks.
The transition \(A\) and source map \(B\) follow from declared dynamics or
finite source perturbations, not endpoint fitting. Consensus includes delay
histories. The code-repair graph uses endpoint information; its readout consumes
scalar values and type labels, and sources are separated by site and channel.
Minimality, sufficiency and internal model-state dimension require separate tests.

For linearized graph dynamics closed in the retained coordinates, with time-independent maps
\(A,B\) and \(\rho(A)<1\), Section~\ref{si:reduced-state-equation}
gives the resolvent. Its residual checks the solve for assembled matrices,
not matrix construction. Sections~\ref{si:minimal-realization}
and~\ref{si:consensus-network} define the spectral and consensus participation
ratios, respectively, including the latter's modal-weight bounds and
quadrature normalization.

\paragraph{Checking response composition in a prescribed graph.}
The deterministic check asks whether direct input perturbations and composed
source-to-state/readout derivatives give the same action-score response. Two
small programming tasks---summing even numbers and testing strict sequence
increase---each have four candidate repair scores. Expected repairs set the
labels, source destinations and some weights, so agreement tests a prescribed
calculation, not independent repair performance or model dynamics.

The linear graph has \(\rho(A)=0.941<1\), allowing the resolvent
\((I-A)^{-1}\). Its solve residual is
\(\|(I-A)\chi_\ast^G-B\|_\infty=1.55\times10^{-15}\);
finite-difference responses and finite-window transition products agree to
maximum absolute error \(1.65\times10^{-13}\). These check different
calculations of the specified graph response.

For four action scores \(s(z)\), distinct from augmented state \(s(t)\), and
expected repair \(a_\ast\), differentiable endpoint response uses
\[
\boldsymbol q_{\rm soft}(z)=
\operatorname{softmax}\!\bigl(s(z)\bigr),\qquad
q_{\rm soft}(z)=\bigl[\boldsymbol q_{\rm soft}(z)\bigr]_{a_\ast},
\]
The binary readout has no differential at its jump. An action is recommended
only when its maximum score reaches threshold \(\theta\); otherwise none is
emitted. Thus \(P_f\) indicates an allowed recommendation and
\(Q_f=\mathbf1[a_{\rm rec}=a_\ast]\le P_f\). Raw-score margins locate
these boundaries:
\[
m_P(z)=\max_a s_a(z)-\theta,\qquad
m_Q(z)=s_{a_\ast}(z)-\max_{a\ne a_\ast}s_a(z),
\]
Away from ties, \(P_f=1\) iff \(m_P\ge0\), and \(Q_f=1\) iff also
\(m_Q>0\); ties use recorded action order. The softmax differential measures
response, whereas margins classify binary outcomes.

Four event amplitudes per task supply source directions through embedding
\(E_{\rm src}\) into an \(8\)-site, \(4\)-channel graph. Comparing
their four output-probability responses with
\(\mathrm D_z\boldsymbol q_{\rm soft}A^{t-t'}BE_{\rm src}\)
gives maximum absolute error \(1.13\times10^{-8}\) at amplitude step
\(0.003\). Increasing the step to \(0.3\) changes estimates by at most
\(1.10\times10^{-3}\) relative to the baseline. Raw-score differences
also agree with the composed derivatives.

Zeroing propagation and cyclically permuting endpoints give relative errors
1.00 and 1.63, respectively. These deliberate controls check use of the maps;
both tasks reuse one topology and are not independent graph replications.

\paragraph{Replay of events from recorded model calls.}
Recorded model calls test whether visible records can define source events
without consulting hidden answers. Seven file-grounded tasks from a live
Qwen2.5-14B-Instruct-MLX run each supply a plan, tool result and answer candidate;
six candidates are correct. Missing verification events are not synthesized.
The readout compares each visible answer candidate with a no-answer option.
Recorded statuses and a fixed candidate-event rule set source amplitudes. Hidden answers enter only endpoint scoring;
the incorrect task's true answer is absent from its actions and sources.

The three event classes have amplitudes \((0.7,1.0,1.0)\) in every task,
so all 21 events instantiate one numerical response template. Its stored
source-to-state matrix has rank three; normalization limits the nonzero
two-action probability response to rank one. The response vanishes before
source time, and difference-amplitude estimates vary by at most
\(9.71\times10^{-14}\). These deterministic replay checks establish source
encoding and propagation, not fresh model response or internal dynamics.

\subsubsection{Fresh model readouts after imposed propagation}
\label{si:live-paired-source-perturbation}
This experiment tests how changes in evidence amplitude affect a model's
support for candidate answers after prescribed graph propagation. Each of seven
constructed tasks requires choosing one of three values for each of three
categorical output variables. Main Section~\mainref{si:worked-graph} specifies
the task records, event extraction and checks, graph construction, state
update and candidate readout in execution order. Section~\ref{si:source-encoding}
defines the general source-encoding map. Here the sampling procedure separates
development choices, fixed-graph perturbations and readout controls; hidden
endpoint verifiers score the model's responses separately.

Each extraction block consists of one instance evaluated with one requested seed. Three
instances/two requested seeds form development; four other instances/one seed
form the confirmatory measurement, excluded from protocol calibration.
Development selects load \(\lambda=0.9375\), rather than calibrating a natural
boundary. Candidate input is \(\lambda\) times within-field mean support plus
\(1-\lambda\) times candidate support, attenuating contrast by \(16\).
The pilot fixes state-distance threshold \(0.05\) and absolute candidate-slope
threshold \(0.01\) before confirmation. Externally constructed graph
coordinates are not internal model states. The measured graph follows
main Eq.~\mainEqref{eq:ai-graph-update} with \(A=\alpha I+\beta W\),
where \(\alpha\) and \(\beta\) are dimensionless uniform retention and
transport gains and \(W\) specifies transport between graph coordinates.
The recorded settings are \(\alpha=0.72\) and \(\beta=0.45\).
Four updates deposit each event at its assigned time, apply \(A\),
rescale if the maximum exceeds one, and clip to \([0,1]\).
Neither rescaling nor clipping fires on the reported \(48\) perturbed
trajectories (\(192\) updates): the largest coordinate before rescaling is
\(0.5651\) and none is negative, so the reported perturbations lie in the
linear regime of the propagator.

For this graph, \(W^2=0\): edges run only from event coordinates to terminal
candidate coordinates. The binomial expansion gives, for integer \(n\ge1\),
\[
 A^n=\alpha^n I+n\alpha^{n-1}\beta W.
\]
An amplitude contrast deposited at \(t_d<T\) therefore gives
\(\Delta z_d(T)=A^{T-t_d}e_d\,\Delta u_d\), provided range control acts as
the identity along both trajectories. This is an exact graph-state response;
the model-readout measurements concern the fixed coefficient pair above.

Baseline event amplitude is \(0.5\). Selected direction \(d\) changes only
one event by \(\pm\epsilon\), with \(\epsilon=0.1\) and refinement
\(\epsilon/2=0.05\). Scalar score \(s=O_T\) from
main Eq.~\mainEqref{eq:ai-candidate-readout}, distinct from full state \(s(t)\),
gives
\[
\Delta z_d(T)=z_{d,+}(T)-z_{d,-}(T),\qquad
\widehat g^s_d(\epsilon)=
\frac{s_{d,+}-s_{d,-}}{2\epsilon}.
\]
Distance \(\|\Delta z_d(T)\|_2\) uses every terminal graph coordinate.
Each candidate receives a separate Y/N query; its conditional Y likelihood
is normalized over allowed values per field. Score \(s\) averages support
for three verifier-selected values, used only after source hashes are saved.
Reader input is task text, complete candidate support/opposing-evidence table
(and their difference) after the fixed mixing transformation, and one candidate.
It excludes event-source coordinates and raw evidence; it asks for best support, not an independently generated
plan. Extraction uses temperature \(0\); scoring uses \(1.0\), top-\(p=1.0\).
Seeds derived from \(4101\) and candidate index match paired counterparts.
This is not a calibrated joint-choice probability. Post-call exactness \(Q\)
is not differentiated; workability is not varied.

\paragraph{Source isolation and measurement units.}
All extraction blocks execute first. Each records normalized events, the graph
and hashes of visible inputs, extractor requests/responses, events and graph,
durably synchronized before the verifier file is opened. Historical records
bind source version, served model identity and files, code, data,
request/response identifiers and the planned denominator. Privately retained
transformed copies replace identifying metadata and dependent hashes, checking internal
consistency rather than authenticating original execution or chronology.

All \(4\) confirmatory blocks complete without adaptive retry, repair or exclusion:
\(4\) extraction plus \(648\) scoring calls (\(652\) total). Endpoint calls
comprise \(432\) perturbation-arm and \(216\) repeated-input/routing controls.
Before verifier access, fixed ordering by event time, type, document, field,
value and identifier selects three events per block from \(15\) total:
\(12\) directions and \(3\) unprobed events. Calls are repeated candidate
measurements within four tasks.

\paragraph{Resolved contrasts and step refinement.}
A first-order interpretation requires a resolved readout slope that remains
consistent when the perturbation is halved. This criterion is met in only
some of the tested directions.
Every selected direction exceeds the state-distance threshold by construction,
with five distinct displacement magnitudes. Nine candidate slopes exceed threshold; the
reactor-inspection instance's three remain below it, not necessarily zero.
Four of nine above-threshold slopes reverse sign, leaving five above-threshold,
sign-stable directions. Only \(2/12\) meet fixed slope agreement:
\[
\frac{|\widehat g^s_d(\epsilon)-\widehat g^s_d(\epsilon/2)|}
{\max\{|\widehat g^s_d(\epsilon)|,|\widehat g^s_d(\epsilon/2)|,10^{-12}\}}
\leq 0.25.
\]
The floor normalizes near zero. Steps are \(20\%\) and \(10\%\) of baseline.
Failure does not distinguish oversized steps from nonlinear or discrete
behavior or establish zero population response.
Post-run saturation counts are \(547/648\) (\(84.4\%\)) scoring calls and
\(355/432\) (\(82.2\%\)) perturbation calls at maximum conditional Y/N
probability at least \(0.99\); these counts do not isolate saturation's
contribution to insensitivity.

Per instance, contrasts occupy distinct event coordinates in a \(21\)- or
\(22\)-component state; their three-column finite-contrast matrix has zero
kernel. This does not identify differential \(\mathcal K_{z,T}\) or test
sufficiency. Two late coupling events and one support event remain unprobed.
Repeated-input controls test aggregate-score repeatability. Each block's two
scores use nine calls each (72 of the 216 controls) and are identical; this
establishes neither sampling independence nor identical raw responses.
Restoration replaces both perturbed graph states with the baseline state;
the no-propagation control disables edge transport. Each operation makes the
paired candidate tables identical, and their readout differences are zero.
These checks verify the numerical inputs, not causal mediation.
Four variants of one template supply no task-family confidence interval,
live multi-agent dynamics, natural unloaded reference or workability crossing.
Within each fixed-task reader interface, access to original evidence is
restricted to the projected support table. This design does not independently
test the full graph's sufficiency or its transfer across tasks.

\subsection{Program-output diagnosis}
\label{si:program-diagnosis}
These checks use task-specific response relations to choose which computed
output to replace. The retrospective examples and prospective pilot differ
in when the diagnostic rule was selected relative to the observed failures.

\subsubsection{Response checks on two saved programs}
\label{si:offline-graph-diagnosis}
\label{si:worked-program}

Two saved model-written programs test whether a known input--output response
can identify a faulty calculation. Each returns an integral and an inverse:
time to a target volume for flow, or the coordinate at a target calibration
value. Correction replaces a selected output with a reference formula while
leaving the program unchanged. Programs and rules were selected after their
defects were known; this is a retrospective diagnostic check.

\paragraph{Task and diagnostic rule.}
At each of three positions, two sensor readings define a mean. The program
receives a vector of prepared means and a separate worksheet of position-indexed
values; both are represented in the graph. The separate input channels permit
tests of which records the program uses. Flow rates
\((2,4,3)\) litres/second over three one-second intervals give total volume
\(9\) litres and time \(13/8\) seconds to \(4.5\) litres. Linear interpolation
of calibration values \((10,20,40)\) at \((0,1,2)\) gives integral \(45\)
and target coordinate \(1/2\) for value \(15\); these coordinates and values
have no assigned physical units.

A uniform shift \(\epsilon\) of the readings must change an integral by
\(\epsilon\) times domain length. Each probe at \(\epsilon=\pm0.1\) adds
\(\epsilon\) to both sensor readings at every position, then recomputes their
means and rebuilds both inputs while preserving any imposed additive preparation
error. At tolerance
\(10^{-8}\), checks compare preparation with sensor means, the inverse with
the allowed domain, and the central-difference integral slope with domain
length (3 for flow, 2 for calibration). A unique violation selects preparation
recomputation, inverse replacement or integral replacement, respectively;
multiple violations cause abstention, and execution/domain errors are
unmeasurable. Reference endpoints and repair outcomes are excluded from
selection. An incorrect inverse inside its domain can escape this rule.

The flow program returns \((9,25/6)\), with target time outside the domain;
inverse replacement gives \((9,13/8)\). The calibration program returns
\((15,1/2)\); integral replacement gives \((45,1/2)\). Pairs denote
(integral, inverse), in the units above. Unchanged execution, deterministic retry, identity intervention and
replacement of the other output each leave both programs jointly incorrect; replacing both outputs corrects both.
Nine constructed controls contain two healthy cases, two repaired preparation
errors, two missed defects, one incomplete repair, one ambiguous case and one
unsupported-domain case. Eight are scorable: selection gives \(4/8\) jointly
correct outcomes versus \(6/8\) for both replacements. Healthy cases enter
these counts, so they are not repair rates. Ordinary diagnostics have access
to the same relations and formulas.

\paragraph{Graph encoding and representation tests.}
The graph transports inputs, not program-generation dynamics. Six events
write preparation/worksheet values as \(v/64\), \(v\in[0,64]\), into twelve
sites and two channels. Positive polarity marks a write, not correctness.
Forward/return weights \(1,0.15\), retention \(1/2\), propagation \(1/4\)
and zero channel coupling produce six terminal action/channel values
\(v/256\), decoded by multiplication by \(256\); both updates are stored.
Programs also receive raw sensor history, targets and durations/spacings.
Blocking preparation fails domain checking; blocking worksheets leaves
preparation readers unchanged but fails constructed worksheet readers.
There is no fallback to original inputs. Restoring the original graph
recovers the wrong answers. Substituting a perturbed graph while keeping the
other inputs fixed reproduces the corresponding source-perturbation output.
These controls establish the declared dependencies.

The representation tests ask which of the supplied records are needed to
account for program outputs. Identical worksheets with different preparation
yield different outputs, rejecting worksheet-only state for those programs. A constructed reader that
also uses raw sensors changes its total \(1.8\to2.0\) at fixed graph,
rejecting graph-only state for that wider input class. The six independent
encoded input directions have zero state-response kernel and cannot provide
positive sufficiency evidence. Readout choice is illustrated by rate direction
\((1,0,-1)\): after inverse replacement, \(V(\delta)=9\) but
\(t_{4.5}(\delta)=13/8-\delta/4\). A total-only readout misses this change.

The two programs come from 30 development calls (6 preparations and 24
continuations), with no passes among 18 eligible accuracy evaluations.
Reconstruction uses deterministic executions of these saved programs, with
response checks at \(\pm0.1,\pm0.05\), and no new model calls.
Separate code checks graph updates, decoding, reference arithmetic, decisions
and grading. The conclusions concern diagnostic logic and representation scope,
not unknown-task repair performance or model-internal mechanisms.

\subsubsection{Prospective checks on four new programs}
\label{si:prospective-diagnostic-pilot}

This pilot asks whether checks of preparation consistency, inverse-output range
and integral response can select a correction before a new program's defects
are known. It uses two flow tasks (total volume
and target time) and two calibration tasks (integral and target coordinate).
The diagnostic rule, settings, actions and analysis were fixed and committed
before Qwen2.5-14B generated one preparation and one program per case: eight
fresh calls without retries or replacement cases. Corrective actions recompute
preparation or replace selected outputs; generated code stays fixed.

\paragraph{Inputs and outcomes.}
Each pair of readings equals its prescribed mean \(\pm0.2\). Flow 01 uses
means \((6,2.5,6)\), durations \((3.5,2,3)\) and target \(13\); Flow 02 uses
\((5.5,7,3)\), \((2,3,2.5)\) and target \(37\). Units are litres/second,
seconds and litres. Calibration 01 uses means \((3,9.5,14)\) and target \(6\);
Calibration 02 uses \((5,12.5,17.5)\) and target \(15\). Both interpolate
linearly at \((0,1,2)\), with no physical units. Reference integration and
inversion use the preceding task definitions. Decisions precede outcomes;
64 deterministic graph executions compare all actions and controls.
Table~\ref{tab:prospective-diagnostic-pilot} shows which outputs each action corrects.

\begin{table}[!htb]
\centering
\caption{\textbf{A corrected component need not repair the task.}
Each of four newly generated programs returns two outputs. Flow outputs are
total volume and time to a target volume; calibration outputs are the integral
of an interpolated curve and the coordinate at a target value. Counts compare
correctness before and after the selected action; any correction replaces
numerical outputs without changing the program.
A task is correct only at 2/2. Both healthy programs remain correct, while
neither faulty program is fully repaired. Exact outputs and reference
calculations are retained in the authors' records.}
\label{tab:prospective-diagnostic-pilot}
\begin{tabularx}{\linewidth}{@{}l c X@{}}
\toprule
\textbf{Program} & \textbf{Correct outputs} & \textbf{Action and remaining error} \\
\midrule
Flow 01 & \(1/2\to1/2\) & No change; target time remains incorrect \\
Flow 02 & \(2/2\to2/2\) & No change; both outputs correct \\
Calibration 01 & \(2/2\to2/2\) & No change; both outputs correct \\
Calibration 02 & \(0/2\to1/2\) & Integral replaced; target coordinate remains incorrect \\
\bottomrule
\end{tabularx}
\end{table}

\paragraph{Interpretation.}
The selected integral correction leaves both faulty programs' inverse outputs
incorrect (Table~\ref{tab:prospective-diagnostic-pilot}). Calibration 02 uses
\((p(1)-p(0)+p(2)-p(1))/2\) for decoded preparation values \(p\):
a uniform shift cancels, exposing the wrong integral response. Replacement gives
the reference \(95/4\). Its inverse, \(4/3\) rather than \(3/2\),
nevertheless passes the domain check \([0,2]\); Flow 01's wrong time 6
rather than \(13/6\) likewise lies in \([0,8.5]\). Response and range
checks can identify one defective output while missing another. Joint
correctness stays \(2/4\), component correctness changes \(5/8\to6/8\),
and full repair is \(0/2\); neither healthy case changes.

Across all four cases, unchanged execution, deterministic repeat, identity
intervention, preparation/worksheet rebuilding and integral replacement each
give \(2/4\) joint correctness; inverse replacement gives \(3/4\), and both
reference replacements \(4/4\). The latter is an arithmetic control, not model
self-repair; no ordinary diagnostic comparator is implemented.
Graph transplants change at least one readout per case; preparation blocking
is inadmissible, worksheet blocking leaves Calibration 02 unchanged, and
restoration returns wrong outputs. No case meets the fixed criterion of full
repair together with a resolved pre-intervention response of the diagnosed
readout. A separate calculation checks recorded decisions, executions and
references; these four cases provide no comparative diagnostic advantage.

\FloatBarrier
\subsection{Presentation changes in a delayed-tool workflow}
\label{si:history-decision}

An inventory reader associates delayed quantity returns with requests, then
decides whether stock suffices for an order. The experiment tests whether
explicitly pairing existing input records restores a missing response. Four
constructed cases cover irrelevant-warehouse returns, reversed arrival/revision
order, cancellation and repeated delivery (cases 1--4). Delays are prescribed
event order, not measured latency.

\paragraph{Task and controlled contrast.}
Qwen2.5-14B-Instruct-MLX-4bit receives all requests (identifier, warehouse,
revision and active/cancelled status), identified returns, a snapshot and an
unrelated note. It should select the highest active revision for the target
warehouse up to the current revision, then allocate only if its quantity
suffices. A fixed warehouse--revision lookup supplies quantities. Exchanging
two requests' warehouses preserves their returns and arrival order.
Snapshot state \(z_P\) records target warehouse, order quantity and current
revision; \(z_H\) additionally records requests. The exchange preserves
\(z_P\) but changes \(z_H\). These analytical descriptions do not restrict
model access to the full input.

\begin{figure}[tbp]
\centering
\includegraphics[width=\linewidth]{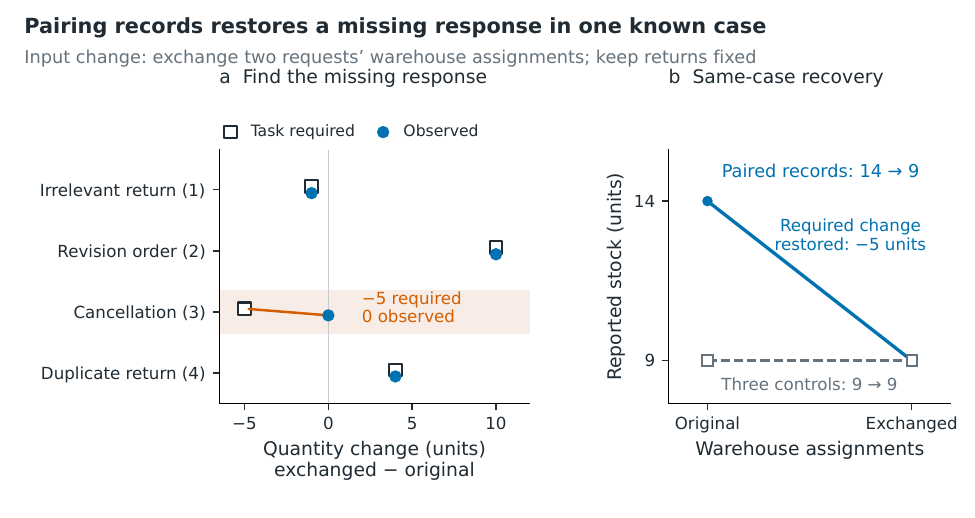}
\caption{\textbf{Pairing existing records restores a missing response on one known case.}
The task selects the highest active request revision for the target warehouse,
excluding revisions later than the current one, and allocates only when the
returned quantity meets the order. In case 3, the order requires ten units
and the current revision is seven.
The source contrast exchanges two requests' warehouse assignments while
keeping returned quantities fixed. \textbf{a}, Required changes (squares)
and observed changes (circles) in four cases. Only the cancellation case
(case 3) lacks the required response: zero rather than \(-5\) units.
Small vertical offsets separate coincident markers. \textbf{b}, A separate
24-call follow-up on case 3. Explicitly pairing every request with its return
by identifier gives \(14/\mathrm{allocate}\to9/\mathrm{hold}\): all
three pairs have correct endpoints and difference \(-5\). Unchanged input,
a checklist, and an unrelated note each give
\(9/\mathrm{hold}\to9/\mathrm{hold}\), or 0/3 correct pairs, and share
the dashed line. Observed values summarize three repeated calls per condition
or three matched response pairs, not independent tasks; squares in panel a
show task-rule reference contrasts. Lines in panel b connect two inputs, not times.
The pairing changes presentation, not available information; the test covers
the same selected case rather than held-out repair performance.}
\label{fig:history-decision}
\end{figure}

The cancellation case (case 3) shows why arrival order alone is insufficient.
It requests ten units at revision seven. Three requests, A, B and C,
return in order B, A, C: A is active at revision six for the target warehouse
and returns 14 units; B is active at revision five for the other warehouse
and returns 9; C is cancelled at revision seven for the target and returns 25.
Exchanging A's and B's warehouses changes the selected quantity
\(14\to9\), and the required action from allocate to withhold. This is the
\(-5\)-unit response in Figure~\ref{fig:history-decision}.

\paragraph{Calibration and exploratory interventions.}
Calibration reproduces the required quantity changes in cases 1, 2 and 4,
while case 3 stays at \(9\to9\). Four cases, two conditions and three repeats
give 24 calls: all exact-key JSON outputs parse, but only 21/24 give the correct quantity and
action, below the fixed 22/24 criterion. The planned 320-call
exploration/validation therefore remains unrun.

The task list, response rule and direct association diagnostic were fixed
before calls, without independent task custody. Settings are temperature 0.2,
top-p 1, 128-token output cap, 8192-token context and 90-second deadline,
with fixed case-derived condition shuffling and no repair or retry.
Both response-based and direct association checks select pairing for case 3
and no change elsewhere before intervention outcomes. Pairing appends every
request joined to its return by identifier; it does not select an answer.

A separately specified exploratory comparison runs six operations on all four
cases, three times each (72 fresh calls). Join, join plus checklist, and an
idealized refresh each give 12/12 jointly correct quantity/action outputs;
unchanged retry, an unrelated note, and checklist alone each give 9/12,
as do the twelve unperturbed calibration calls (excluding the twelve
source-contrast calls). All differences occur in case 3 (3/3 versus
0/3); the other three cases contribute 9/9 in every operation. The checklist
restates filtering, revision choice and allocation. Refresh supplies an
authoritative quantity and is answer-informed. No failed call is replaced.

A descriptive cost comparison also favors a fixed join. Utility assigns
\(+1\) to jointly correct output, \(-2\) to unsafe allocation, \(-1\)
to invalid output and zero otherwise, minus 0.01 per deployment call and 0.05
per refresh. Both diagnostic policies use six probes per case and one selected
continuation on case 3, reusing historical outputs elsewhere. Their mean utility
is 0.9375 versus 0.99 for fixed join or join plus checklist; unused-arm collection
is excluded. These illustrative weights are not measured costs, and fixed
strategies have no independent selection set. Ordinary diagnosis selects the
same action; the deterministic task rule could also be executed directly.

\paragraph{Same-case response restoration.}
A separate exploratory test repeats both conditions of the known failure case:
four operations, two inputs and three repeats give 24 fresh calls. The protocol,
randomized schedule and reconstruction were fixed before these calls; the case
and successful join were already known. No historical output enters a pair,
and pair indices do not imply shared seeds. All outputs parse. The criterion
requires both correct endpoints and the exact signed difference in all three
join pairs; Figure~\ref{fig:history-decision} reports every result.

Restoring the two-point response rules out an always-14 answer; the unchanged
\(9/\mathrm{hold}\) endpoint alone cannot do so. Both the full task rule and
choosing the first target-warehouse return remain compatible with the result.
Joining changes presentation, salience and length, rather than available
information. This same-case exploratory result identifies neither a unique
mechanism nor state sufficiency, held-out repair performance or diagnostic
advantage. The note control supplies no distributional-equivalence bound.
Exact requests and responses remain in the authors' records.

\FloatBarrier
\subsection{Development study of response-guided action selection}
\label{si:development-decisions}

This development study asks whether measuring changes in selected evidence
helps choose an intervention that improves a claim-verification answer.
It compares selection from state and endpoint responses with endpoint-only
features and direct reading of the retrieved evidence. SciFact claims~\citep{wadden2020fact} enter a two-stage pipeline: evidence
selection followed by claim classification.
BM25 keyword ranking retrieves six abstracts from the claim text alone.
Qwen2.5-14B-Instruct (MLX 4-bit) selects sentence identifiers; a source-checked
parser resolves these to ordered original text. A second call reads only the
claim and selected evidence, returning support, contradiction or insufficient
evidence. Reference labels score correctness. Format conversions preserve
source text; invalid selections, unlike valid empty ones, skip the reader and
leave its outcome unavailable. Serial calls use fresh caches, explicit seeds,
temperature 0.7, no probability truncation or repetition penalties, and
output limits of 512 tokens for evidence selection and 16 for claim classification.

Fixed hash order selects one training claim per source group linked by citations
or duplicates, excluding earlier design and pilot groups. Of 96 tasks, 89
complete, one is partial and six unrun. A post-interruption amendment, made
before scoring, includes the completed initial segment without imputation.
Each task has two recorded sets of action outcomes, sharing some calls:
89 groups, not 178 independent tasks. Outcomes average the two sets within
each task.

\paragraph{Actions and comparison methods.}
Five actions keep the original answer, reselect evidence, change the reader
prompt at fixed evidence, read all six abstracts, or aggregate upstream and
altered-reader answers. Aggregation returns their common valid label or uses
another call with visible evidence and answers, never reference labels.
Consistency voting uses up to eight readings of the original evidence.
All actions are evaluated on the original, undeleted task.

The feature comparison asks whether observing a response helps beyond ordinary
answer and evidence diagnostics. Two probes delete the first-ranked abstract
or first three, without replacement. Ordinary features \(B\) include validity,
label, evidence length and retrieval rank, retrieval scores, available confidence,
missingness and changes caused directly by deletion. Endpoint features \(E_i\)
add label/score contrasts and repeat controls. State features \(S_i\) also
measure whether the selector changes its choice among surviving sentences;
matched \(E/S\) sets share all other values and availability indicators.

The index \(i\) counts probes. One probe uses the first deletion and
fixed-evidence rereading; two add the second deletion and repeated selection
and classification, with maximum diagnostic costs of five and nine calls.
Consistency features \(K\) add repeated-answer statistics to \(B\).
The primary comparisons are \(S_2\) against \(E_2\) and direct reading;
\(S_1\) comparisons are secondary.

Actions maximize predicted repair minus harm relative to the original answer,
with fixed ties. Six ridge predictors use five fixed group folds, training-fold
median imputation/standardization and penalties \(\{0.1,1,10,100,1000\}\)
chosen by Brier loss (squared probability error). Six sensitivity
comparators use single-split probability rules, including a constant candidate,
with training-only features, thresholds and leaf frequencies. Tuning reuses
assessment folds, and accuracy-first fixed-action selection uses the same
development outcomes. These are descriptive comparisons, not independent
confirmation. Stored chronology does not independently authenticate label
withholding or original request provenance.

\begin{figure}[!htbp]
\centering
\includegraphics[width=\linewidth]{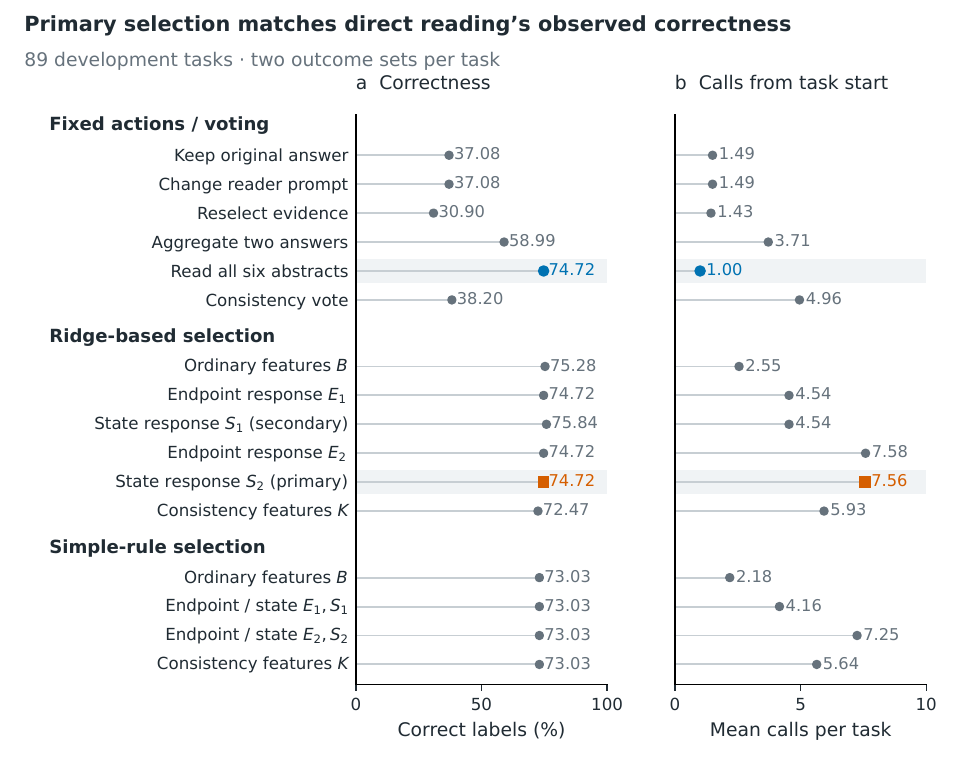}
\caption{\textbf{The primary response-guided strategy matches direct reading's observed correctness at higher call cost.}
\textbf{a}, Correct-label fractions for all declared methods on 89 completed
SciFact tasks. \textbf{b}, Mean calls from task start, including diagnostic
measurements and the selected action; shared calls count once. Both panels
average two sets of action outcomes within each task and then across tasks:
89 source groups, not 178 independent samples. Invalid or unavailable labels
remain incorrect. Highlighted rows are direct reading and the primary
state-response strategy \(S_2\). Both have 65/89 and 68/89 correct outcomes
in the two recorded sets, averaging 74.72\%, at 1.00 and 7.56 calls per task.
\(B\) uses answer and evidence diagnostics, including validity, labels,
retrieval information and changes caused directly by deletion. The index
\(i=1,2\) counts probes: remove the first-ranked abstract, then additionally
test removal of the first three. \(E_i\) adds changes in the reader's label
and score; \(S_i\) also measures changes in the selection of surviving sentences
beyond mechanical deletion. \(K\) adds repeated-answer statistics. Direct
reading supplies all six abstracts. Identical simple-rule pairs share a row.
These are development estimates: tuning and fixed-action selection reuse the
assessment data. Point differences do not establish superiority or equivalence.
Per-method counts and token costs remain in the authors' records.}
\label{fig:development-decisions}
\end{figure}

\paragraph{Results and measurement coverage.}
The primary response-guided strategy \(S_2\) matches direct reading and
\(E_2\) in observed correctness on all 178 recorded task outcomes
(Figure~\ref{fig:development-decisions}). It costs 7.56 calls per task,
compared with 1.00 for direct reading.
The ordinary-feature selector \(B\), without state or endpoint response
features, gives 66/89 and 68/89 correct answers in the two outcome sets
(75.28\% on average), at 2.55 calls per task. Secondary \(S_1\) exceeds \(E_1\) and direct reading by 1.12 percentage
points: one task is corrected in both outcome sets by changing the reader
prompt. This does not replace the primary comparison.
These development estimates establish neither population superiority nor
equivalence.

The response measurements require a usable evidence selection. Initial
selection calls produce 44 valid, 42 invalid and three capped/unavailable
outputs. The invalid outputs comprise 24 sentence-object schema errors,
eight failures to reproduce source text, six top-level key errors and four
JSON parsing errors; sentence objects occur despite requests for identifiers.
Reselection among surviving sentences is measurable on 23/89 tasks after
removing the first-ranked abstract and 28/89 after removing the first three:
16 allow both measurements, 35 at least one, and 54 neither. Seventeen calls reach the output cap,
including the three unavailable initial evidence selections. All 89 tasks stay in the
denominator. Missing measurements are not zeroes and do not establish the
cause of absent primary benefit. The accuracy contrast between selected-evidence
and direct-reading paths does not compare matched states;
it cannot reject a representation.

The stopped run has 1,511 attempts, one with unknown completion; independent
confirmation was not performed. Recorded outputs permit reanalysis of policy
choices and observed costs, without authenticating historical execution.
Costs of the unknown completion remain unavailable.

\FloatBarrier
\subsection{Build-and-test workflow: task and screening results}
\label{si:workflow-development}
An authored software workflow asked whether required packages could be released:
every prerequisite had to use the correct component and source revision, remain
uncancelled, and return successfully within the observation window. Build and
test operations were executed locally; delivery times were assigned.
Candidate summaries are tested by comparing release, hold and invalid-output
frequencies for histories they merge. Correctness is scored separately against
the prerequisite rules; agreement can include a shared incorrect judgment.
Qwen2.5-14B-Instruct (MLX 4-bit) read complete histories and tool returns;
candidate states and reference judgments were withheld from its inputs.
Unlike the closed-loop task (Section~\ref{si:closed-loop}), the model reports
only the release judgment, without first reporting current-version test statuses.
The task rules, record presentation and output requirements also differ; this
is not a matched interface ablation.

Each history contains build, test and package jobs for two components.
Four fault families alter a prerequisite's component, its source revision,
a required test's cancellation status, or the build revision used by a test.
Each family has healthy and faulty variants, each with two preparations that
differ in an unrelated job, note and dispatch time.
Five candidate representations retain:
(i) current revisions and the pending-job count;
(ii) this snapshot, required package jobs, and each job's component and operation;
(iii) the preceding description plus job revisions and cancellation status;
(iv) current revisions and the dependency graph of required package jobs and
their transitive prerequisites; or (v) the full history.
The graph consistently renames identifiers and omits jobs outside those
dependencies, while the model still sees the original complete histories.
This candidate library is not a nested sequence.

The readout distinguishes a release decision, a hold decision and an invalid
response. READY and HOLD denote valid JSON judgments with the single field \texttt{ready}
set to \texttt{true} or \texttt{false}, respectively; all other complete responses
are INVALID. Sixteen ten-job histories, eight healthy and eight faulty, received
four calls each: 64 discovery calls plus eight preliminary sampling checks.

Discovery requests temperature \(0.7\), top-p \(1\), top-k \(0\),
distinct fixed seeds and a 64-token output limit. Eight unscored preliminary
calls vary omitted, distinct or repeated seeds and zero temperature, while
allowing either Boolean answer. All return READY, leaving sampling independence
and effective seed/temperature handling unverified.

Faulty prerequisites often leave the release judgment unchanged: strict
correctness is 32/64, the count obtained by always answering READY.
All 32 calls on healthy histories returned READY, as did all 24 calls across
six faulty histories with wrong-component prerequisites, obsolete source
revisions or required tests using obsolete builds. The two cancellation
histories gave one READY and seven INVALID responses. The invalid responses
began with a false judgment but added prose and reached the output limit;
their classification is unchanged. These 32 calls on eight faulty histories
are repeated measurements, not independent tasks.

No candidate met both requirements for response agreement and comparison coverage.
Selection compared distinct histories assigned the same candidate state within
each fault family, requiring
at least one pair per fault family and empirical total variation---the largest
probability difference over output events---at most \(0.15\) on every pair.
The graph retains component associations, revisions, cancellation statuses and
dependency edges; it had eight matched pairs within the prescribed families.
One cancellation pair gave empirical distance \(0.25\), the other seven zero.
With four draws per history, \(0.25\) is the smallest nonzero histogram distance.
Snapshot and association-only candidates also exceeded the histogram threshold.
The revision/status description had matched pairs in only one family, while
full history had none; both lacked the required comparison coverage.
The fixed stopping rule left all 1,024 confirmation calls unrun. This preliminary
comparison is not a formal statistical rejection of the graph
representation. The observations do not isolate the causes of task errors or
format failures and provide no positive confirmation of representation adequacy.

\FloatBarrier
\subsection{Historical readout comparisons}
\label{si:cross-family-comparison}
\label{si:reduced-coordinate-comparison}
\label{si:reported-coordinates-controls}
\label{si:selected-path-measurements}

These historical records compare tool use with answer correctness as the
permitted number of tool steps decreases. Tool use defines \(P_f\); tool use
with a correct answer defines the nested readout \(Q_f\). Write \(w=P_f\)
and \(r=Q_f/P_f\) for \(P_f>0\). The comparisons illustrate these fractions,
with retrospectively selected paths rather than a common response law.

\paragraph{Spreadsheet paths and nested readouts.}
Eight historical paths vary tool-step budgets on author-constructed
spreadsheet tasks derived from GAIA~\citep{mialon2023gaia}. They are not official
benchmark evaluations. Seven paths, selected retrospectively
across three spreadsheet task groups, decrease from exact or partially
exact operation to failure. None visits all four regimes. The eighth path,
on another spreadsheet, has
\((w,r):(1,5/6)\to(1,1)\to(0,{-})\), contradicting monotonicity.

Three workflow implementations provide 12 records on four tasks and 36 on twelve tasks.
At tool budgets 2 and 1, respectively, all are workable; exact outcomes are
12 and 10 in the smaller set, 25 and 22 in the larger. Thus ratios such as
\(11/18=22/36\) describe pooled outcomes, not an 18-record sample. At zero
budget \(w=0\) and \(r\) is undefined. Separate constraint-planning,
recommendation and delayed-counting records illustrate workable but inexact
outcomes. Their different controls and readouts, incomplete counts and
retrospective selection do not support a common response law. The ordered
boundary result requires Section~\ref{si:local-readout-regimes}'s hypotheses.

Records contain task identifiers, answers, correctness, tool use, tool-step
budgets, run indices, elapsed time and model-call counts. They lack the
stopping rule, independent pre-scoring timestamp, temperature, seed, model
revision and per-run tool/verification counts. They therefore support
coordinate illustrations, not complete execution reproduction or
pre-scoring protocol attestation. These paths are distinct from the seven
live-origin records used for graph replay, although four official task
identifiers overlap.

\subsubsection{Exploratory persistence records}
\label{si:exploratory-persistence}
\label{si:intervention-scope}

A historical rule-based counting task tests information carried between steps
by removing, masking or restoring it while fixing task, model family, interface,
actions and stopping rule. It also records a persistent reference, a
mixture-of-agents baseline using aggregated summaries, and a single-agent
snapshot. No cross-agent propagation was measured.

Removal gives a wrong answer; the original baseline, masking and restoration
time out. A separately recorded persistent reference and the mixture baseline
answer correctly. Restoration recovers information-visibility proxies but not
the answer. Removal, masking, mixture and snapshot each fail the joint criterion
of no worse recovery and no greater resource use than the persistent reference,
with at least one strict improvement.

These proxies lack complete formulas and denominators; they are neither
probabilities, nested \((P_f,Q_f)\) readouts nor the coherence functional
of Section~\ref{si:well-posedness-coherence}. The removal, masking and
restoration reruns duplicate earlier records byte-for-byte. Missing execution
settings, an unreleased fuller report and no snapshot recovery trajectory
preclude source-level reconstruction, independent replication or a general
necessity claim. Derived comparisons remain in the authors' records.

\FloatBarrier
\section{Evidence status, data and code availability}
\label{si:data-code-availability}
\subsection{Evidence status and reconstruction limits}
\label{si:implementation-guide}\label{si:reconstruction-levels}
Analytic results, comparisons fixed before measurement, and developmental or
exploratory analyses support different inferences
(Table~\ref{tab:reconstruction-coverage}). Each study's methods specify its task,
comparison rule and sampling assumptions. Local hashes and timestamps bind files;
they do not independently authenticate registration or historical execution.
Without raw records and implementations, readers cannot numerically reanalyze
all reported results.

\begin{table}[!htbp]
\centering\small
\caption{\textbf{Scientific role and inferential scope of the evidence.}
New-history comparisons are distinct from independent tasks or external
replication. Sampling-based claims remain conditional on each study's stated
assumptions. Analyses of the same observations are not independent evidence.}
\label{tab:reconstruction-coverage}\label{tab:execution-roles}
\begin{tabularx}{\linewidth}{@{}>{\raggedright\arraybackslash}p{0.22\linewidth}>{\raggedright\arraybackslash}p{0.25\linewidth}Y@{}}
\toprule
Study & Evidence status & Supported interpretation and limit \\
\midrule
Physical measurements & Analytic response and numerical calibration & Local consensus calibration and distinct flock readouts; neither empirically tests inverse state adequacy. \\
\addlinespace
Request--return matching & Constructed program and proof & Association suffices for the specified continuations; executions check its implementation. \\
\addlinespace
Recipient-A queries & Readout selected after development; comparisons fixed before new calls & Bounded support on four histories, restricted to recipient A. \\
\addlinespace
Varied identifiers and order & Fixed comparisons; exploratory reanalysis & Three pairs remain unresolved under original bounds; post-hoc categorical bounds resolve differences. \\
\addlinespace
Added identifiers and order & Fixed new comparisons & Association and extended states both have bounded support; added benefit is unestablished and the earlier anomaly was not retested. \\
\addlinespace
Adaptive tests & Instructed task; fixed confirmation after interface development & Feedback tests prescribed version information; the instructed rule predicts every observed action. \\
\addlinespace
Excerpt comparison & Fixed comparisons on eight SciFact training claims & Categorical bounds reject both compressed descriptions; neither is adequate across the claim set. \\
\addlinespace
Build and test & Stopped development screen & No candidate selected or confirmation run; not an interface ablation of the adaptive task. \\
\addlinespace
Delayed tools and programs & Constructed cases and pilots & Specific presentation changes and output replacements; no independent decision benefit established. \\
\addlinespace
Graph readout & Imposed propagation; fresh model readouts & Finite differences test local linearity; each graph's three-direction state-response matrix has full column rank. \\
\addlinespace
Development decisions & Reused assessment folds & Primary selection matches direct reading at greater cost; missing measurements do not explain this result. \\
\addlinespace
Historical persistence & Exploratory records & Incomplete provenance and scoring limit authentication and interpretation. \\
\bottomrule
\end{tabularx}
\end{table}

\FloatBarrier

\subsection{Data availability}
The physical studies use numerical simulations of delayed consensus and polar
flocking. AI data comprise constructed event graphs and request--return
histories, saved model-generated programs, and recorded model responses.
The excerpt-comparison and action-selection studies use SciFact's
training set and abstracts~\citep{wadden2020fact}, available at
\url{https://github.com/allenai/scifact}. Historical comparisons use
author-constructed variants derived from GAIA~\citep{mialon2023gaia}, rather
than an official benchmark evaluation.

The manuscript and SI describe the task inputs, sampling and scoring rules,
controls, and aggregate results, including unsuccessful comparisons. Raw
experimental records and physical simulation arrays have not been publicly
deposited. Consequently, the publication alone does not permit reconstruction
of every reported figure or response comparison. The source dataset's public
availability does not supply the constructed prompts, generated responses or
simulation records.

\subsection{Code availability}
The manuscript and SI specify equations, task rules and statistical procedures
for independent implementation. LaTeX sources reproduce submitted documents
and figures but provide no executable experimental reproduction package.
Experiment, analysis and plotting implementations are not publicly released.

\section{Notation and terminology}
\label{si:notation-terminology}

A loading parameter is a scalar or vector control varied about a fixed
reference protocol. A finite-window response maps a source perturbation to
retained-state and endpoint-readout changes during a fixed window.

Each graph experiment specifies its own coordinates and event-to-coordinate
map. The model-readout graph separates event sources, candidate support and
opposing evidence; the program graph encodes preparation and worksheet values.
These externally constructed states are neither graph neural networks nor
model hidden states.

Prescribed delayed inputs originate outside the modeled local operators;
delay alone does not imply endogenous feedback. Internally generated delayed
quantities require their histories in the augmented state (retarded dynamics).
Retarded response specifies causality---zero response before the source
acts---independently of endogenous delayed feedback.

An operating branch is a local regular branch of retained-state equilibria
parametrized by control; stability is a separate condition. A slow manifold
is a locally invariant manifold on which eliminated variables are slaved to
retained coordinates, where such a reduction exists. A linear-system residual is
the back-substitution error for a linear solve using assembled \(A,B\);
it does not check matrix assembly.

{
\begingroup
\setlength{\LTcapwidth}{\linewidth}
\renewcommand{\arraystretch}{1.08}
\begin{longtable}{@{}>{\raggedright\arraybackslash}p{\dimexpr 0.360\linewidth-0.720\tabcolsep\relax}>{\raggedright\arraybackslash}p{\dimexpr 0.640\linewidth-1.280\tabcolsep\relax}@{}}
\caption{\textbf{Selected symbols.} Full-state, retained-state and readout roles are distinguished; realization-specific meanings are identified explicitly.}
\label{si:tab:symbols} \\
\toprule
\textbf{Symbol} & \textbf{Meaning and scope} \\
\midrule
\endfirsthead
\caption[]{Selected symbols (continued).}\\
\toprule \textbf{Symbol} & \textbf{Meaning and scope} \\
\midrule
\endhead
\bottomrule
\endfoot
\(c_t\in H\) & shared coordination field in Hilbert field space \(H\), the field component of the augmented state \\
\(s_t\in X_f\) & full augmented state containing the field, memories, required delay histories, direct readout inputs and observation records \\
\(z_T=G_f(s_T)\in Z_f\) & candidate retained state; may include the full field. Low dimension, endpoint sufficiency and autonomous closure each require separate justification \\
\(O_T:X_f\to\mathcal Y_f\) & endpoint readout on the full augmented state, with observation space \(\mathcal Y_f\); numerical observables take values in \(\mathbb R^d\) \\
\(\bar O_T:Z_f\to\mathcal Y_f\) & readout as a function of retained state, where \(O_T(s)=\bar O_T(G_f(s))\); the bar distinguishes the domain and does not denote an average \\
\(\mathrm D_xF(x)\) & derivative at \(x\), mapping an input perturbation to a first-order output change; a Jacobian in finite dimensions or a Fr\'echet derivative on a function space \\
\(\mathrm D_sO_T\) & readout sensitivity to the full state; under differentiable factorization it equals \(\mathrm D_z\bar O_T\,\mathrm D_sG_f\), evaluated at the corresponding state values \\
\(\bar O_{f_i,T}\), \(\mathcal Z_i^{\mathrm{reach}}\) & realization-specific projected readout and reached retained-state set in a two-realization comparison \\
\(d\ge1\), \(q\) & readout-component count and field-channel count; neither fixes augmented- or retained-state dimension \\
\(u_t\), \(\delta u_t\), \(\Delta u_t\) & controlled source coordinates, tangent perturbation and finite source contrast; \(u\) also parametrizes a fixed operating branch \\
\(\mathcal U_T\) & declared finite-dimensional source-perturbation space over the observation window \\
\(\mathcal K_{z,T},\mathcal K_{O,T}\) & first-order maps from \(\mathcal U_T\) to retained-state and endpoint displacements \\
\(\mathcal H_T^{(1)}\) & first-order readout map on reached state displacements; exists when every source direction with zero state response also has zero readout response \\
\(\boldsymbol y_{\rm cons}=(\mathcal A,\phi,V)\) & consensus window summary: amplitude, phase lag and transverse variance, evaluated on the observed field trajectory \\
\(\sigma_\theta^{(P)}\) & flock width inferred from \(P_f\); a steady-state consistency summary \\
\(r_{\rm PR}\) & participation ratio using squared singular-value weights of the constructed response matrix \\
\(r_{\rm eff}\) & consensus participation ratio using \(E_0=\mathcal A^2\) and transverse weights summing to \(V\) \\
\(h_t,a_t,r_{t+1}\) (adaptive workflow) & visible interaction history, model-selected action and subsequent delivered tool observation \\
\(\pi(a\mid h),q_u(r\mid h,a)\) & fixed decision policy and tool-observation law under external condition \(u\); neither implies autonomous evolution of a candidate summary \\
\(A_t,B_t,\Phi(T,t)\) & linearized full-state transition, source Jacobian and ordered transition product \\
\(A_t^{\rm ret},B_t^{\rm ret},\Phi_{\rm ret}(T,t)\) & corresponding maps for a closed retained-state linearization, where that closure holds \\
\(A,B\), \(\rho(A)\) & time-independent transition and source maps of the closed retained graph; spectral radius of \(A\) \\
\(\alpha,\beta\) (event graph) & dimensionless uniform retention and transport gains; fixed settings \(0.72,0.45\) \\
\(L\), \(J_i\) & intrinsic field operator and composite local source \\
\(R_i,\pi_i,W_i\) & local read, update and write maps \\
\(G_i\) & source-coupling operator for a modeled local operator \\
\(E_\ell\), \(\tau_\ell,y_\ell\) & coupling, delay and prescribed signal for a delayed input whose generator remains outside the modeled local operators \\
\(\xi_a^t\) & history of an internally generated delayed quantity, retained in \(s(t)\) \\
\(C(t)\) & coherence functional, Section~\ref{si:well-posedness-coherence} \\
\(F_{\rm full}(s;u,y^t)\) & augmented controlled evolution, including memory laws and internal history transport; \(y^t\) is prescribed \\
\(f_u(z)\) & reduced drift in \(\dot z=f_u(z)\), where the stated local reduction exists \\
\(P_f,Q_f\) (nested criteria) & outer workability and inner exactness; logical inclusion and normalization give \(0\le Q_f\le P_f\le1\). Numerical order alone does not establish nesting \\
\(\widehat P_f,\widehat Q_f\) (consensus) & sample support fractions obtained by averaging the binary readouts over the same \(50\) seeds \\
\(P_f,Q_f\) (flock) & time- and seed-averaged polarization and alignment fraction; these observables are not nested \\
\(w=P_f\), \(r=Q_f/P_f\) & workability and inner-to-outer ratio for nested
criteria; \(w\) remains defined at \(P_f=0\), where \(r\) is undefined \\
\(r=Q_f/P_f\) (flock) & alignment fraction divided by polarization for
\(P_f>0\); not conditional exactness; the nested-readout interpretation of \((w,r)\) does not apply \\
\(P_f^{\rm prep},G_f\) & initial-state preparation and full-to-retained-state coarse-graining maps; preparation differs from subsequent dynamics and scalar \(P_f\) \\
\(\widehat\chi^G(t,t')\) & finite-graph finite-window response \\
\(\chi_\ast^G=(I-A)^{-1}B\) & time-summed graph response for time-independent stable dynamics \\
\(\chi_\ast\) & continuous-time stationary branch susceptibility \(-A_z^{-1}B_z\), under the stated branch hypotheses \\
\(\Sigma_{EW},\Sigma_{WR},\Sigma_{RO}\) & regular readout boundaries; the stated sequence requires actual crossings, no re-entry and ordered, separated thresholds on a fixed path \\
\end{longtable}
\endgroup
}

\FloatBarrier
\bibliographystyle{naturemag}
\bibliography{refs}